\documentclass[10pt,twocolumn,letterpaper]{article}

\usepackage[pagenumbers]{wacv} 

\usepackage{algorithm}
\usepackage{listings}
\usepackage{graphicx}
\usepackage{amsmath}
\usepackage{amssymb}
\usepackage{booktabs}
\usepackage{tabularx}
\usepackage{makecell}
\usepackage{enumitem}
\usepackage{multirow}
\usepackage{indentfirst}

\usepackage[pagebackref,breaklinks,colorlinks]{hyperref}

\usepackage[capitalize]{cleveref}
\crefname{section}{Sec.}{Secs.}
\Crefname{section}{Section}{Sections}
\Crefname{table}{Table}{Tables}
\crefname{table}{Tab.}{Tabs.}

\def\wacvPaperID{*****} 
\def\confName{WACV}
\def\confYear{2024}

\begin{document}

\title{AnimateDiff-Lightning: Cross-Model Diffusion Distillation}

\author{Shanchuan Lin \quad Xiao Yang \\
ByteDance Inc.\\
{\tt\small \{peterlin, yangxiao.0\}@bytedance.com}
}

\newcommand\banner{%
    \vspace{30pt}
    \hspace*{-0.26\textwidth}
    \centering
    \footnotesize
    \setlength\tabcolsep{0pt}
    \renewcommand{\arraystretch}{0}
    \begin{tabular}{ c c c c c c }
        \includegraphics[width=0.25\textwidth]{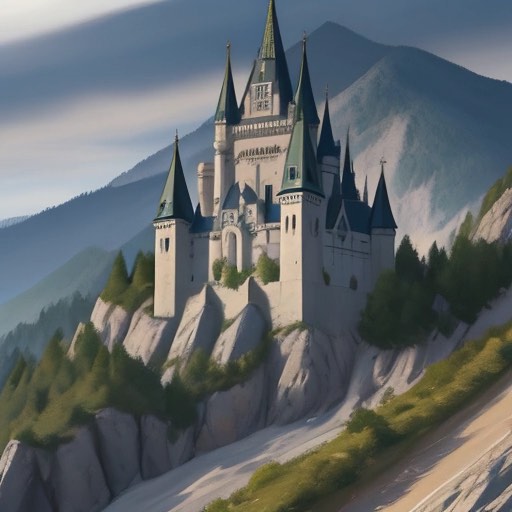} &
        \includegraphics[width=0.25\textwidth]{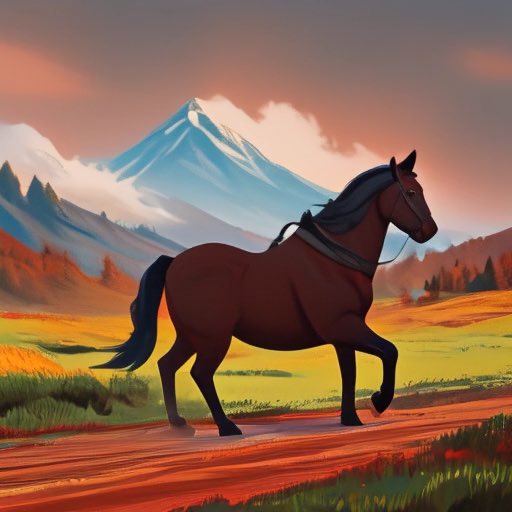} & 
        \includegraphics[width=0.25\textwidth]{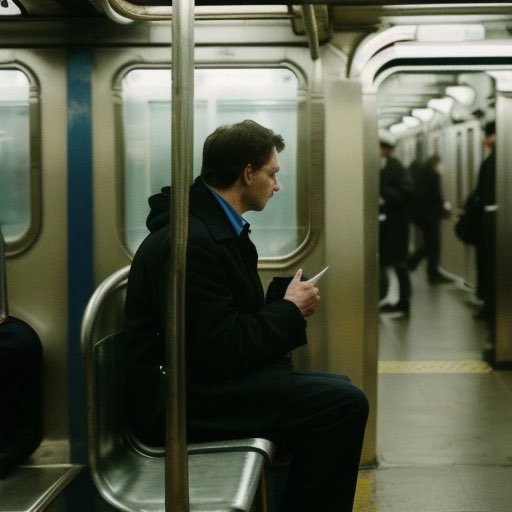} & 
        \includegraphics[width=0.25\textwidth]{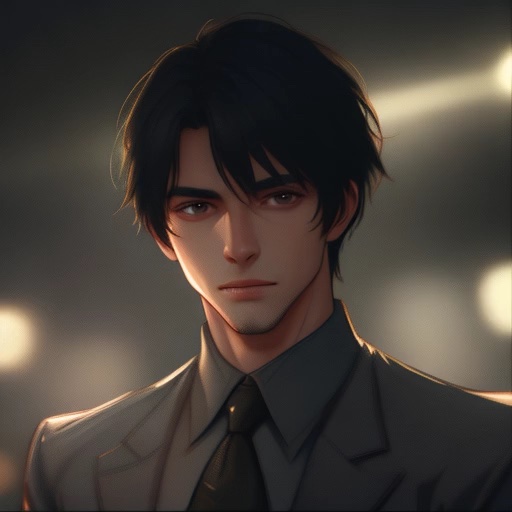} &
        \includegraphics[width=0.25\textwidth]{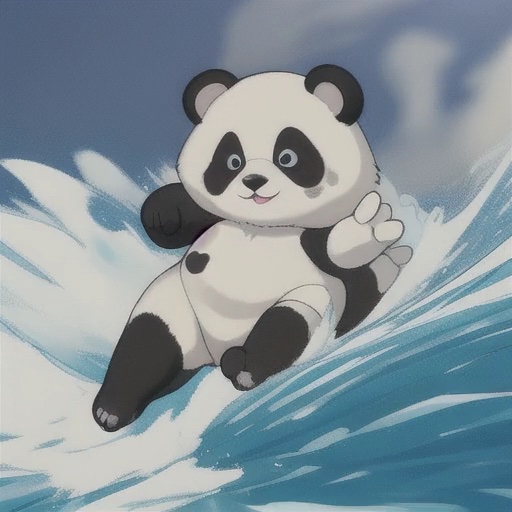} &
        \includegraphics[width=0.25\textwidth]{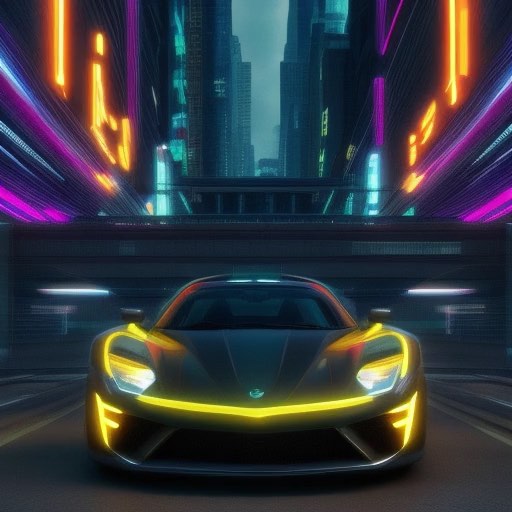} \\
        \includegraphics[width=0.25\textwidth]{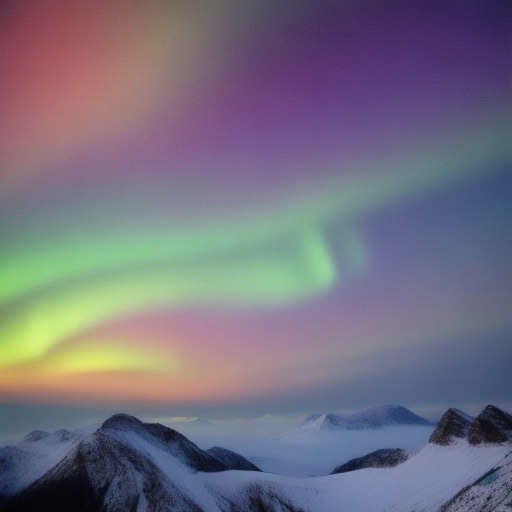} &
        \includegraphics[width=0.25\textwidth]{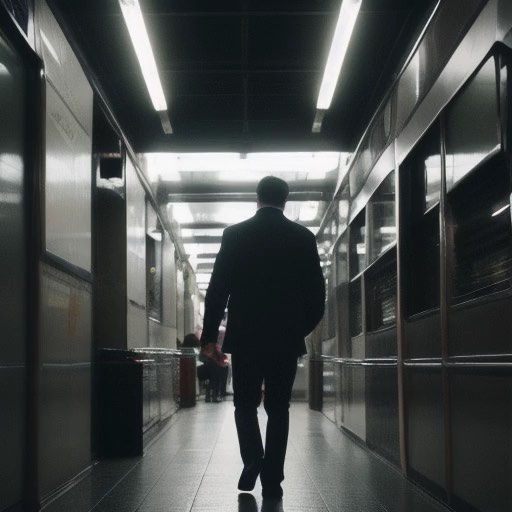} &
        \includegraphics[width=0.25\textwidth]{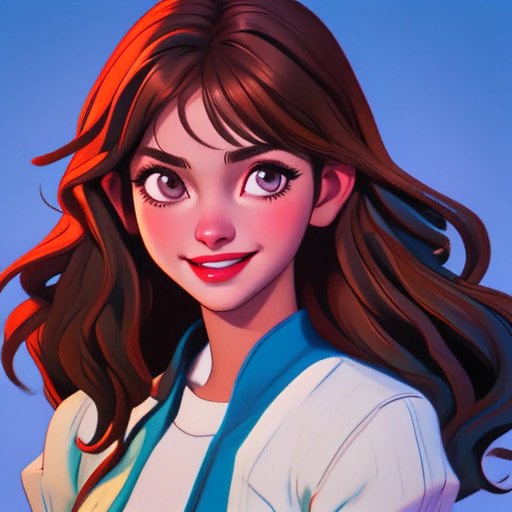} &
        \includegraphics[width=0.25\textwidth]{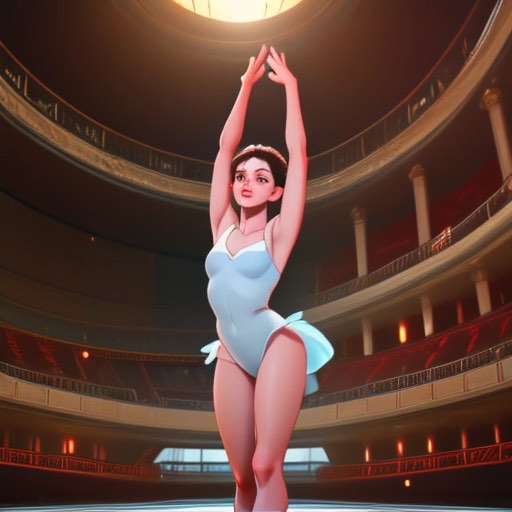} & 
        \includegraphics[width=0.25\textwidth]{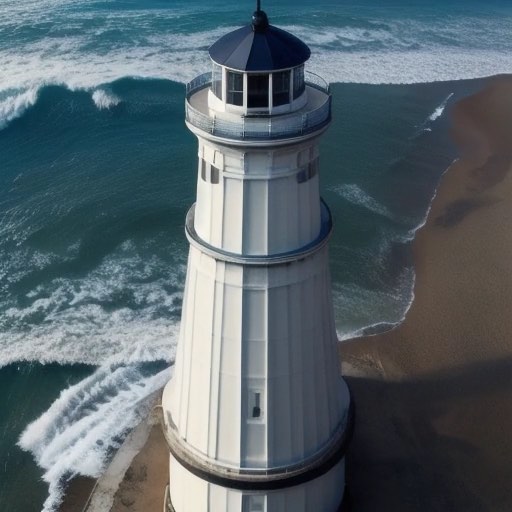} &
        \includegraphics[width=0.25\textwidth]{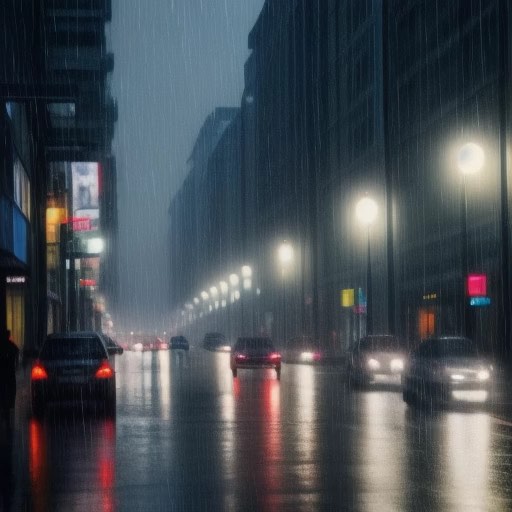}
    \end{tabular}
    \vspace{25pt}
}

\date{}
\maketitle


\begin{abstract}
We present AnimateDiff-Lightning for lightning-fast video generation. Our model uses progressive adversarial diffusion distillation to achieve new state-of-the-art in few-step video generation. We discuss our modifications to adapt it for the video modality. Furthermore, we propose to simultaneously distill the probability flow of multiple base diffusion models, resulting in a single distilled motion module with broader style compatibility. We are pleased to release our distilled AnimateDiff-Lightning model for the community's use.
\end{abstract}

\let\thefootnote\relax\footnotetext{Model: \href{https://huggingface.co/ByteDance/AnimateDiff-Lightning}{https://huggingface.co/ByteDance/AnimateDiff-Lightning}}

\section{Introduction}
Video generative models are gaining great attention lately. Text-to-video models \cite{blattmann2023align,zhou2023magicvideo,ho2022imagen,singer2022makeavideo,blattmann2023stable,esser2023structure,guo2023animatediff,wang2024magicvideov2} allow the creation of videos straight from ideation; image-to-video models \cite{blattmann2023stable,esser2023structure,guo2023animatediff,wang2024magicvideov2} enable more fine-grained control over content and composition; video-to-video models \cite{esser2023structure,guo2023animatediff} can convert existing videos to different styles, such as anime or cartoon. The advancement in video generation has enabled brand-new creative possibilities.

Among all methods, AnimateDiff \cite{guo2023animatediff} is one of the most popular video generation models. It takes a frozen image generation model and injects learnable temporal motion modules into the network. This allows the model to inherit the image priors and learn to produce temporally coherent frames from limited video datasets. Since the image model's architecture and weights are unchanged, it can be swapped with a wide range of stylized models post-training to create amazing anime and cartoon videos, \etc. Additionally, AnimateDiff is compatible with image control modules, such as ControlNet \cite{controlnet}, T2I-Adapter \cite{mou2023t2iadapter}, IP-Adapter \cite{ye2023ipadapter}, \etc, which further enhance its versatility.

However, speed is one of the main hurdles preventing video generation models from wider adoption. State-of-the-art generative models are slow and computationally expansive due to the iterative diffusion process. This issue is further worsened in video generation. For example, many video stylization pipelines using AnimateDiff with ControlNet and a stylized image model can take up to ten minutes to process a ten-second video. Making the generation faster while retaining its quality is the main focus of this work.

Diffusion distillation \cite{lin2024sdxllightning,salimans2022progressive,song2023consistency,kim2023consistency,sauer2023adversarial,luo2023latent,luo2023lcmlora,yin2023onestep,liu2022flow,liu2023instaflow,zheng2024trajectory,wang2024animatelcm,song2023improved} has been more widely researched in image generation. Recently, progressive adversarial diffusion distillation \cite{lin2024sdxllightning} has achieved state-of-the-art results in few-step image generation. In this paper, we apply it to video models for the first time, demonstrating the applicability and superiority of this method on the video modality. We will discuss our designs and changes made specifically for video model distillation.

In addition, we propose to simultaneously distill the probability flow of multiple base diffusion models. Specifically, we take special consideration into the fact that AnimateDiff is widely used with different stylized base models. However, all existing methods perform distillation only on the default base model, and can only hope that the distilled motion module will still work after swapping onto a new base. In practice, we find the quality degrades as the inference step reduces. Therefore, we propose to explicitly and simultaneously distill a shared motion module on different base models. We find this approach not only improves quality on the selected base models, but also on unseen base models.

Our proposed AnimateDiff-Lightning can generate better quality videos in fewer inference steps, out-competing the prior video distillation method AnimateLCM \cite{wang2024animatelcm}. We release our distilled AnimateDiff-Lightning model for the community's use.

\section{Background}

\subsection{Diffusion Model}

Diffusion models \cite{ho2020denoising,song2021scorebased} are behind most state-of-the-art video generation methods. The generation involves a probability flow \cite{song2021scorebased,liu2022flow,lipman2023flow} that gradually transports samples $x_t$ from the noise distribution $t=T$ to the data distribution $t=0$. A neural network $f$ is learned to predict the gradient at any location of this flow. Because the flow is curved and complex, the generation must only take a small step along the gradient at a time, repeatedly invoking expansive neural network evaluations. Diffusion distillation trains the neural network to directly predict the next flow location farther ahead, allowing traversing the flow with bigger strides and fewer steps.

\subsection{Progressive Adversarial Diffusion Distillation}

Progressive adversarial diffusion distillation \cite{lin2024sdxllightning} proposes to combine progressive distillation \cite{salimans2022progressive} and adversarial loss \cite{goodfellow2014generative}. Specifically, progressive distillation \cite{salimans2022progressive} trains a student network to directly predict the next flow location $x_{t-ns}$ from the current flow location $x_t$ as if the teacher network has stepped through $n$ steps of stride $s$. After the student converges, it is used as the teacher and the process repeats itself for further distillation:
\begin{align}
    x_{t-ns} &= \mathbf{EulerSolver}(f_{\mathrm{teacher}}, x_t, t, c, n, s) \\
    \hat{x}_{t-ns} &= \mathbf{EulerSolver}(f_{\mathrm{student}}, x_t, t, c, 1, ns) \\
    &\quad\quad\mathcal{L}_{\mathrm{mse}} = \| \hat{x}_{t-ns} - x_{t-ns} \|_2^2 \label{eq:mse}
\end{align}

However, theoretical analysis \cite{lin2024sdxllightning} has shown that exact matching with mean squared error (MSE) as in \Cref{eq:mse} is impossible due to reduced model capacity, so adversarial loss is introduced to trade-off between quality and mode coverage. The method proposes to first distill with discriminator $D$ conditioned on $x_t$ and caption $c$ to enforce flow trajectory preservation:
\begin{align}
    p &= D(x_t, x_{t-ns}, t, t-ns, c) \\
    \hat{p} &= D(x_t, \hat{x}_{t-ns}, t, t-ns, c)
\end{align}

Then, distill with discriminator $D'$ without the condition on $x_t$ to relax the trajectory requirement to improve quality:
\begin{align}
    p &= D'(x_{t-ns}, t-ns, c) \\
    \hat{p} &= D'(\hat{x}_{t-ns}, t-ns, c)
\end{align}

The distillation trains the diffusion model and the discriminator with non-saturated adversarial loss \cite{goodfellow2014generative} in alternating iterations:
\begin{align}
    \mathcal{L}_{D} &= -\log(p) - \log(1 - \hat{p}) \\
    \mathcal{L}_{G} &= -\log(\hat{p})
\end{align}

SDXL-Lightning \cite{lin2024sdxllightning} achieves new state-of-the-art in one-step/few-step text-to-image generation with this distillation method. Our work is the first to apply this method in video diffusion distillation, demonstrating the applicability and superiority of the method in other modalities.

\subsection{Other Diffusion Distillation Methods}

Diffusion distillation is mostly studied in image generation. Most notably, Latent Consistency Model (LCM) \cite{luo2023latent,luo2023lcmlora} applies consistency distillation \cite{song2023consistency} for latent image diffusion models; InstaFlow \cite{liu2023instaflow} uses a technique called rectified flow (RF) \cite{liu2022flow} to gradually make the flow straighter as a way to reduce sampling steps; SDXL-Turbo \cite{sauer2023adversarial} uses adversarial loss with score distillation sampling (SDS) \cite{poole2022dreamfusion} to push generation down to one step. SDXL-Lightning \cite{lin2024sdxllightning} is the latest research in distillation and achieves even better quality compared to previous methods with progressive adversarial distillation.

Research on video diffusion distillation is very scarce. AnimateLCM \cite{wang2024animatelcm} is the only work on video diffusion distillation so far to the best of our knowledge. It follows LCM \cite{luo2023latent,luo2023lcmlora} to apply consistency distillation \cite{song2023consistency} on AnimateDiff. AnimateLCM can generate great quality videos with eight inference steps but starts to show artifacts with four inference steps, and the results are blurry under four inference steps.

\subsection{Distillation as Pluggable Modules}

LCM \cite{luo2023lcmlora}, AnimateLCM \cite{wang2024animatelcm}, and SDXL-Lightning \cite{lin2024sdxllightning} have explored training the distillation as a pluggable module. The module contains additional parameters on top of the frozen base model, allowing the module to be transplanted onto other stylized base models post-training.

However, the distillation module is only trained on the default base model and the whole approach depends on the assumption that other stylized base models have similar weights. Empirically, we find the quality degrades as the inference step reduces on unseen base models.

In this paper, we explore explicitly and simultaneously distilling the distillation module on multiple base models for the first time. This provides a quality guarantee on the selected base models. We also find it improves compatibility on unseen base models.

\section{Method}

We propose to train a shared distilled motion module on multiple base models simultaneously for AnimateDiff \cite{guo2023animatediff}. The resulting motion module has better few-step inference compatibility with different base models.

\subsection{Model and Data Preparation}

Besides the default Stable Diffusion (SD) v1.5 base model \cite{rombach2022highresolution}, we select multiple additional target base models based on their popularity. For realistic style, we select RealisticVision v5.1 \cite{realisticvision} and epiCRealism \cite{epicrealism}. For anime style, we select ToonYou Beta 6 \cite{toonyou}, IMP v1.0 \cite{imp}, and Counterfeit v3.0 \cite{counterfeit}.

The existing video dataset WebVid-10M \cite{Bain2021FrozenIT} only contains realistic stock video footage. The samples are especially out-of-distribution when distilling the anime models. Therefore, we apply AnimateDiff on all the selected base models to mass-generate data samples. Specifically, we generate video clips using the prompts from WebVid-10M \cite{Bain2021FrozenIT}. We use DPM-Solver++ \cite{lu2023dpmsolver} with 32 steps and a classifier-free guidance (CFG) scale of 7.5 without negative prompts. All the clips are 16 frames and 512$\times$512 resolution. In total, we have generated 1.75 million clips.

\subsection{Cross-Model Distillation}

The AnimateDiff model $F_i$ is composed of the frozen image base model $f_i$ and the shared motion module $m$, where $i$ denotes the index of the specific base model.
\begin{equation}
    F_i := f_i \circ m
\end{equation}

At distillation, we only update the weights of the motion module and keep the weights of the image base model unchanged. We load different image base model $f_i$ on different GPU ranks and initialize the motion module $m$ with the same AnimateDiff v2 checkpoint \cite{guo2023animatediff}. The specific assignments are shown in \Cref{tab:cross-model-asignment}.

\begin{table}[b]
    \centering
    \setlength\tabcolsep{3pt}
    \begin{tabularx}{\linewidth}{lXl}
        \toprule
        Rank & Base Model & Dataset \\
        \midrule
        0 & Stable Diffusion v1.5 \cite{rombach2022highresolution} & \multirow{2}{*}{WebVid-10M \cite{Bain2021FrozenIT}} \\
        1 & Stable Diffusion v1.5 \cite{rombach2022highresolution} & \\
        \midrule
        2 & RealisticVision v5.1 \cite{realisticvision} & \multirow{2}{*}{Generated Realistic} \\
        3 & epiCRealism \cite{epicrealism} & \\
        \midrule
        4 & ToonYou Beta 6 \cite{toonyou} & \multirow{4}{*}{Generated Anime} \\
        5 & ToonYou Beta 6 \cite{toonyou} & \\
        6 & IMP v1.0 \cite{imp} & \\
        7 & Counterfeit v3.0 \cite{counterfeit} & \\
        \bottomrule
    \end{tabularx}
    \caption{Model and dataset assignments across 8 GPU ranks in a single machine. The same configuration is replicated to additional machines.}
    \label{tab:cross-model-asignment}
\end{table}

This design allows the motion module to be simultaneously distilled on multiple base models. Spreading different base models across GPUs eliminates the need for constant swapping of the base models on each GPU. We modify the PyTorch Distributed Data Parallel (DDP) framework \cite{paszke2019pytorch} to prevent synchronization of the frozen image base model from erasing our model assignments. After the modification, the gradients are automatically accumulated using the existing distributed training mechanism to ensure optimization toward accurate distillation on all base models.

We also assign different distillation datasets according to the image base model. For distilling the Stable Diffusion base model, we use the WebVid-10M dataset \cite{Bain2021FrozenIT}. For distilling each realistic or anime model, we pool together all the generated data of its kind to improve diversity. We also employ random horizontal flips to double the sample count.

\subsection{Flow-Conditional Video Discriminator}

Progressive adversarial diffusion distillation \cite{lin2024sdxllightning} proposes to use discriminator $D$ to ensure that the student prediction of $x_{t-ns}$ from $x_t$ given caption $c$ is sharp and flow-preserving. Since our distillation now involves multiple flows of different base models, we must extend the discriminator to be flow-conditional. Specifically, we provide the corresponding base model index $i$ to the discriminator. This way the discriminator can learn and critique separate flow trajectories for each base model:
\begin{equation}
\begin{aligned}
    &D(x_t, x_{t-ns}, t, t-ns, c, i) \\
    & \quad := \sigma\bigg(\mathrm{head}\Big( d(x_{t-ns}, t-ns, c, i), d(x_t, t, c, i)\Big)\bigg)
\end{aligned}
\end{equation}

We follow prior works \cite{lin2024diffusion,lin2024sdxllightning} to take the diffusion UNet \cite{ronneberger2015unet} encoder and midblock as the discriminator backbone $d$. In our case, we use the AnimateDiff architecture \cite{guo2023animatediff}, which consists of the image base model initialized with SD v1.5 weights \cite{rombach2022highresolution} and the motion module initialized with AnimateDiff v2 weights \cite{guo2023animatediff}. We include flow condition $i$ as a new learnable embedding and add it to the time embedding. The shared backbone processes $d(x_{t-ns}, t-ns, c, i)$ and $d(x_t, t, c, i)$ independently. The resulting midblock features are concatenated along the channel dimension before passing to a prediction head. The prediction head consists of blocks of 3D convolution with a kernel size of 4 and a stride of 2, group normalization \cite{wu2018group}, and SiLU activation \cite{hendrycks2023gaussian,ramachandran2017searching} to further reduce the dimension to a single value. Finally, the sigmoid function $\sigma(\cdot)$ clamps the value to $[0, 1]$ range, denoting the probability of the input $x_{t-ns}$ being generated from the teacher as opposed to the student. The entire discriminator, including the backbone, is trained.

Progressive adversarial diffusion distillation \cite{lin2024sdxllightning} also proposes to further finetune the model without condition on $x_t$ at each stage to relax the flow trajectory preservation requirement and further improve the quality. But note that despite the flow trajectory preservation is relaxed, we still must enforce the student prediction to be within the distribution of the target flow. Therefore, we also modify this discriminator $D'$ to be conditional on flow $i$:
\begin{equation}
\begin{aligned}    
    & D'(x_{t-ns}, t-ns, c, i) \\
    & \quad := \sigma\bigg(\mathrm{head}\Big( d(x_{t-ns}, t-ns, c, i)\Big)\bigg)
\end{aligned}
\end{equation}

\subsection{Distillation Procedure}

We progressively distill the model in the following step count order: $128\rightarrow32\rightarrow8\rightarrow4\rightarrow2$. We use mean squared error (MSE) and apply classifier-free guidance (CFG) on $128\rightarrow32$ distillation. The CFG scale is set to 7.5 and no negative prompts. We use adversarial loss for the rest of the stages. Note that our data generation uses DPM-Solver++ \cite{lu2023dpmsolver} for 32 steps. Since DPM-Solver++ produces better quality than Euler, we still decide to start the distillation from 128 steps for extra quality.

The distillation is performed on 64 A100 GPUs. Each GPU can only process a batch size of 1 due to the memory constraint, so we apply a gradient accumulation of 4 to achieve a total batch size of 256. Other hyperparameters, such as learning rate, \etc, follow SDXL-Lightning \cite{lin2024sdxllightning} exactly. We adopt the linear schedule \cite{ho2020denoising} as used in the original AnimateDiff but use pure noise at the last timestep as model input during training following \cite{lin2024sdxllightning} to ensure zero terminal SNR \cite{lin2023common}.

Unlike SDXL-Lightning \cite{lin2024sdxllightning}, we cannot switch to $x_0$-prediction while keeping the base model frozen for one-step generation, so we train the model in $\epsilon$-prediction.

Compared to AnimateLCM \cite{wang2024animatelcm}, which first distills the image base model as a LoRA module \cite{hu2021lora} on image datasets and then distills the video motion module on limited video datasets to combat data scarcity, our method distills the whole AnimateDiff model as a whole. Furthermore, we find the distillation can be trained on the motion module alone for satisfactory quality and there is no need for an additional LoRA module on the image base model.

\section{Evaluation}

\subsection{Qualitative Evaluation}

\Cref{fig:qualitative} shows qualitative comparison of our model to the original AnimateDiff \cite{guo2023animatediff} and AnimateLCM \cite{wang2024animatelcm}. Our method achieves better quality with 1-step, 2-step, and 4-step inference compared to AnimateLCM. The difference is particularly pronounced when using 1-step and 2-step inference as AnimateLCM fails to generate sharp details. Additionally, our method using cross-model distillation can better retain the original style of the base model. AnimateLCM sometimes over-exposes and differs from the base model's style and tone even when using 8-step inference.

\Cref{fig:qualitative-mistoon} shows the results of our model when applied to an unseen base model: Mistoon Anime v1.0 \cite{mistoon}. The style gradually deviates from the original style as the inference step reduces, but note that our model still generates results closer to the original compared to AnimateLCM in terms of the overall anime style, clothing, and hair color of the characters. More analysis on the effect of cross-model distillation is provided in \Cref{sec:cross}. More analysis on unseen models is provided in \Cref{sec:unseen}

The 1-step model produces heavy noise artifacts. This is likely due to the numerical instability of the epsilon formulation, which is also encountered by SDXL-Lightning \cite{lin2024sdxllightning}. For the 2-step model, we notice that it produces more pronounced brightness flickers. Note that the flickers have existed since the original AnimateDiff model. We find the 4-step model strikes the balance between quality and speed.

\begin{figure*}
    \centering
    \captionsetup{justification=raggedright,singlelinecheck=false}
    \small
    \setlength\tabcolsep{2pt}
    \begin{tabularx}{\textwidth}{|X@{\hskip 6pt}|X|X|X|X@{\hskip 6pt}|X|X|X|X|}
        \normalsize{\textbf{Original}} \cite{guo2023animatediff} & \multicolumn{4}{l|}{\normalsize{\textbf{Ours}}} & \multicolumn{4}{l|}{\normalsize{\textbf{AnimateLCM}} \cite{wang2024animatelcm}} \\
        \footnotesize{CFG7.5} & \multicolumn{4}{l|}{\footnotesize{No CFG}} & \multicolumn{4}{l|}{\footnotesize{No CFG}} \\
        32 Steps & 8 Steps & 4 Steps & 2 Steps & 1 Step & 8 Steps & 4 Step & 2 Steps & 1 Step \\
    \end{tabularx}

    \begin{subfigure}[b]{\textwidth}
        \centering
        \setlength\tabcolsep{0pt}
        \renewcommand{\arraystretch}{0}
        \begin{tabularx}{\textwidth}{@{}X@{\hskip 4pt}XXXX@{\hskip 4pt}XXXX@{}}
            \includegraphics[width=\linewidth]{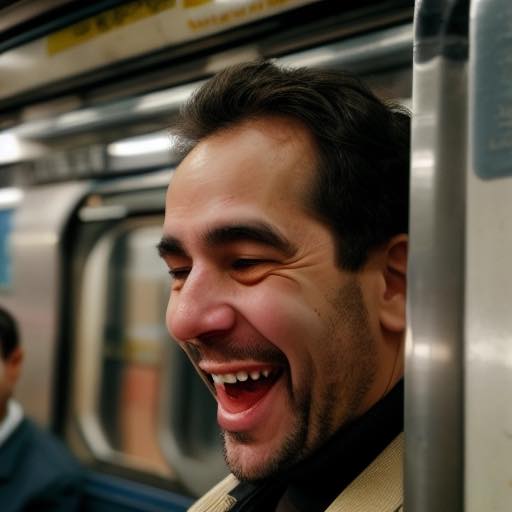} &
            \includegraphics[width=\linewidth]{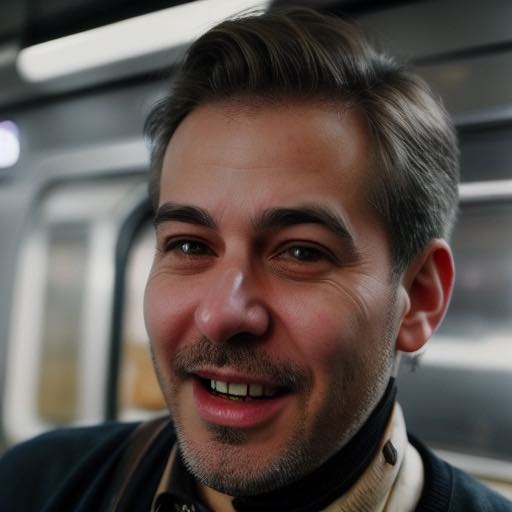} &
            \includegraphics[width=\linewidth]{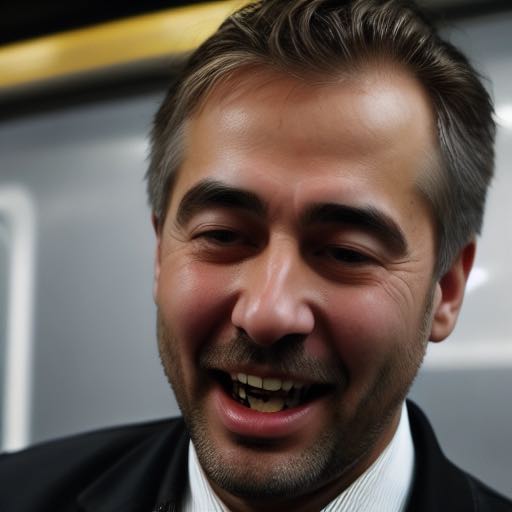} &
            \includegraphics[width=\linewidth]{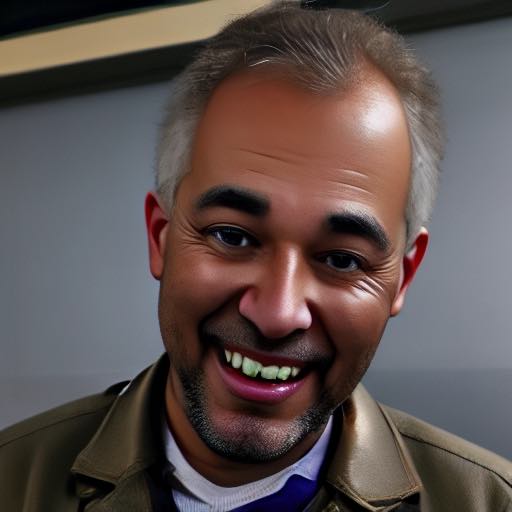} &
            \includegraphics[width=\linewidth]{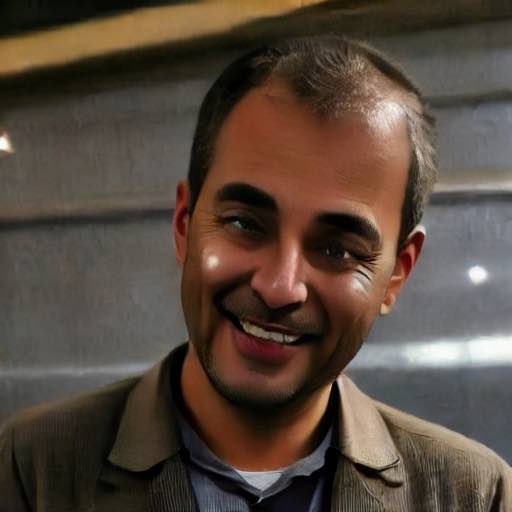} &
            \includegraphics[width=\linewidth]{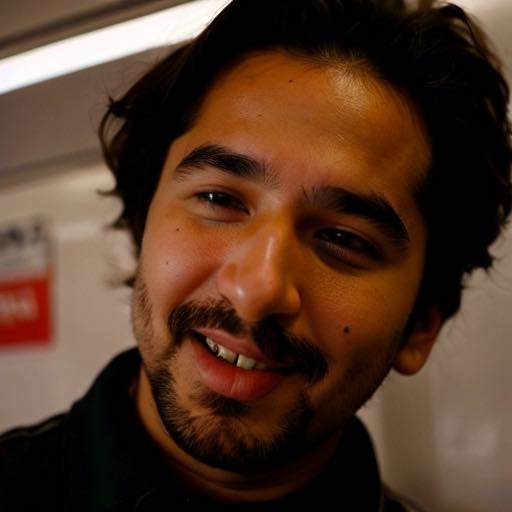} &
            \includegraphics[width=\linewidth]{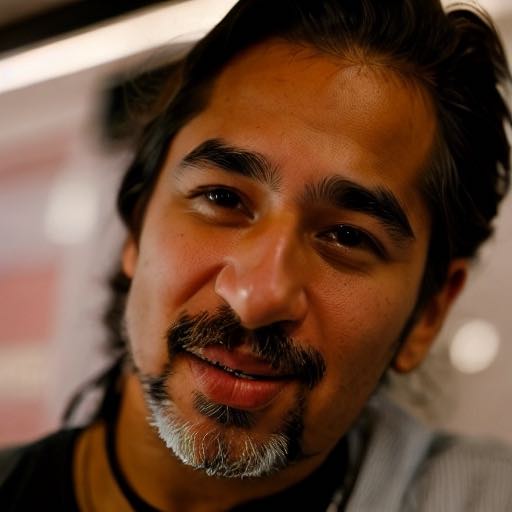} &
            \includegraphics[width=\linewidth]{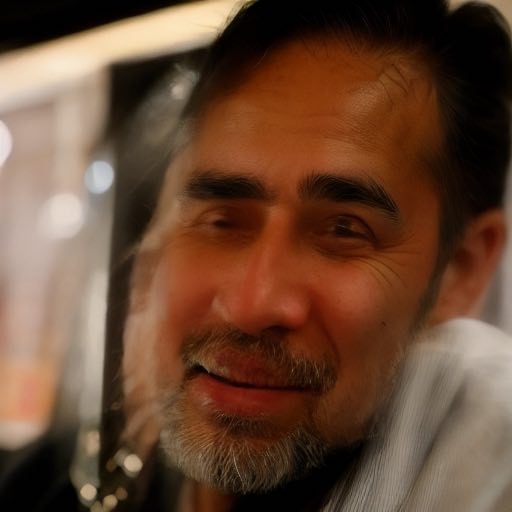} &
            \includegraphics[width=\linewidth]{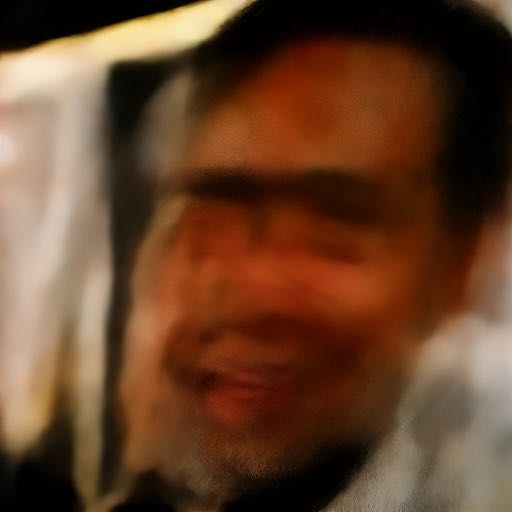} \\
            \includegraphics[width=\linewidth]{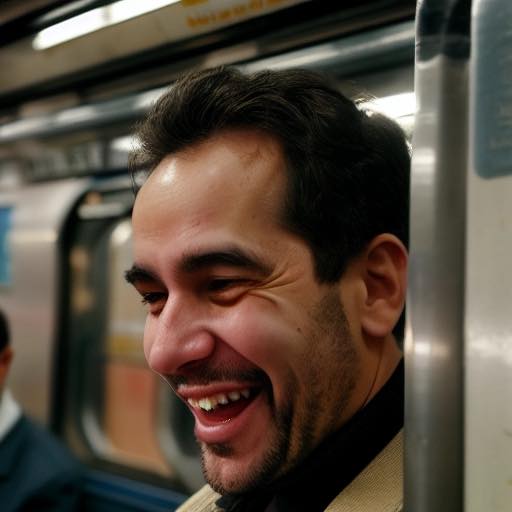} &
            \includegraphics[width=\linewidth]{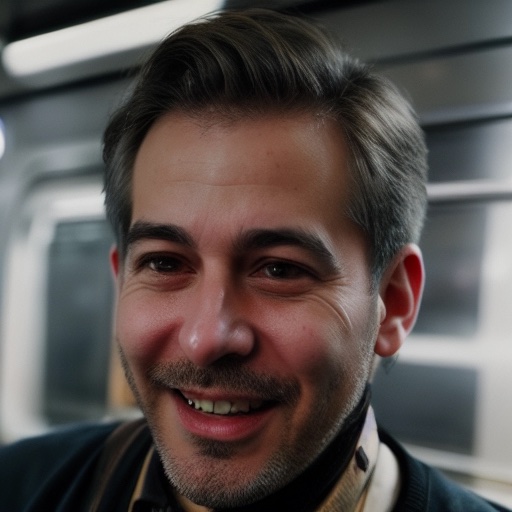} &
            \includegraphics[width=\linewidth]{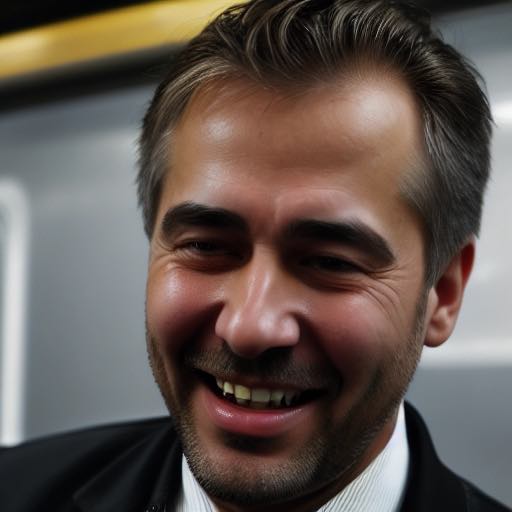} &
            \includegraphics[width=\linewidth]{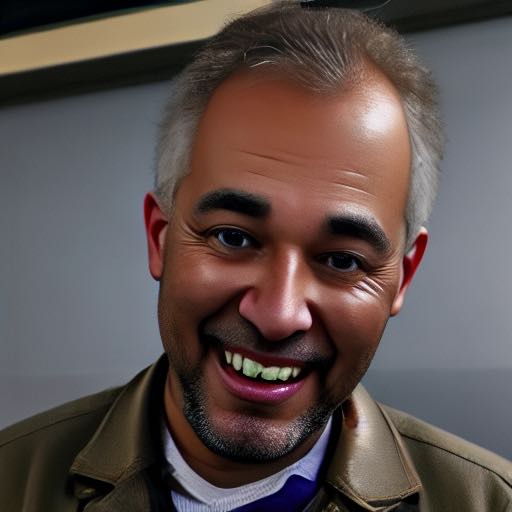} &
            \includegraphics[width=\linewidth]{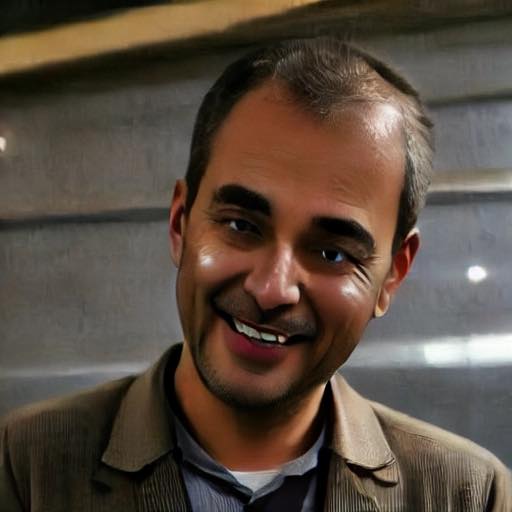} &
            \includegraphics[width=\linewidth]{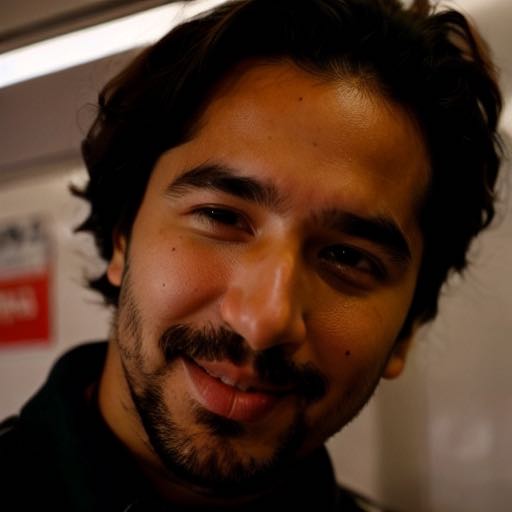} &
            \includegraphics[width=\linewidth]{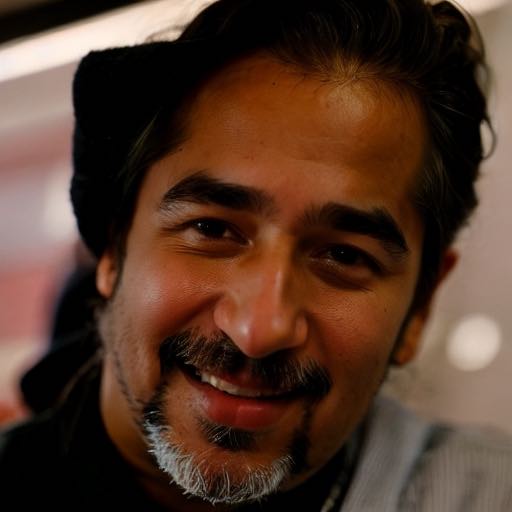} &
            \includegraphics[width=\linewidth]{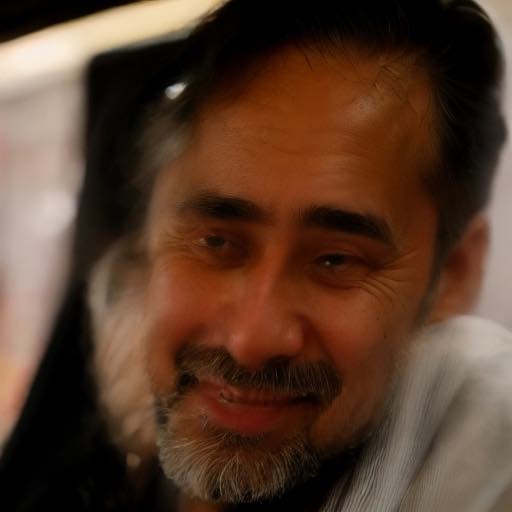} &
            \includegraphics[width=\linewidth]{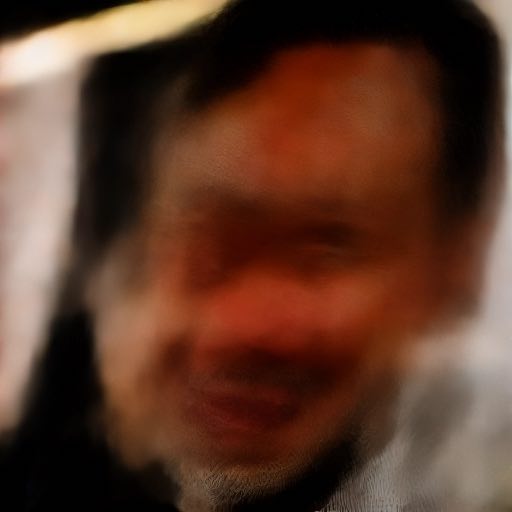} \\
            \includegraphics[width=\linewidth]{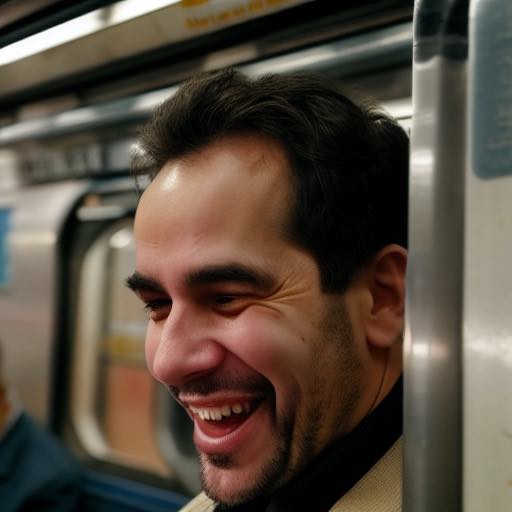} &
            \includegraphics[width=\linewidth]{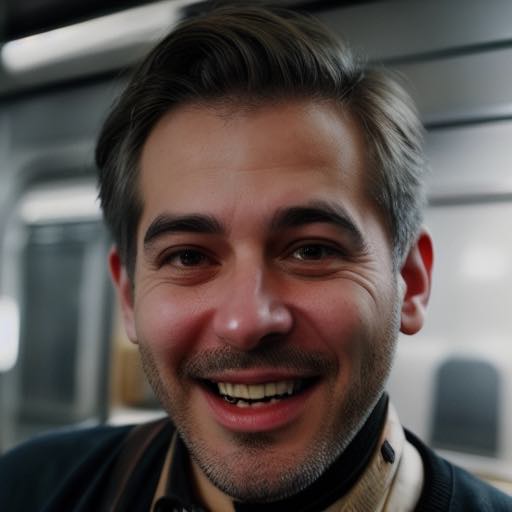} &
            \includegraphics[width=\linewidth]{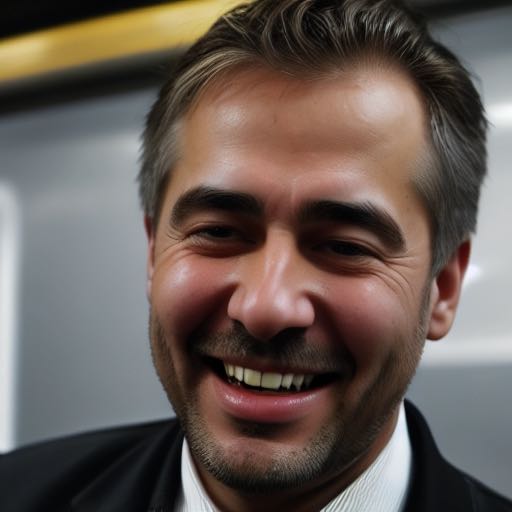} &
            \includegraphics[width=\linewidth]{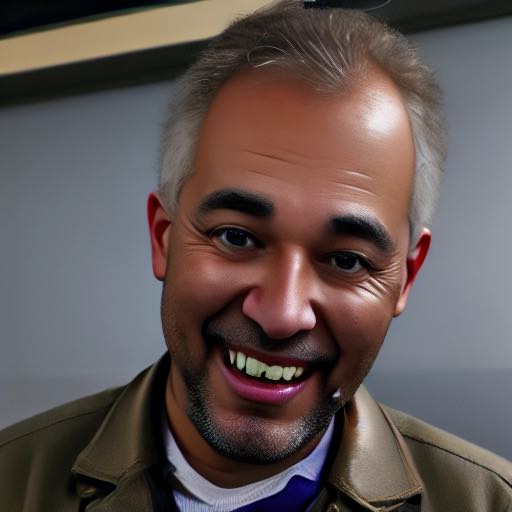} &
            \includegraphics[width=\linewidth]{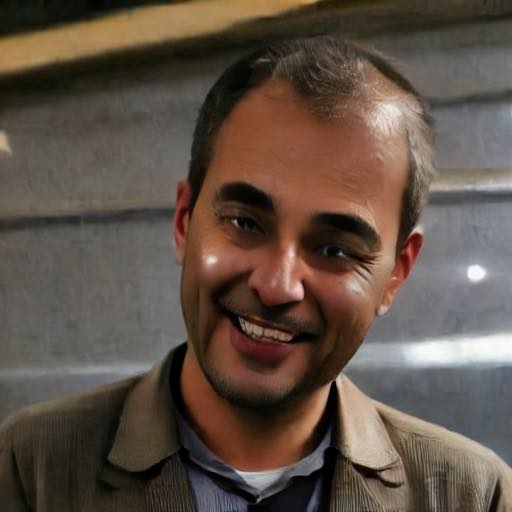} &
            \includegraphics[width=\linewidth]{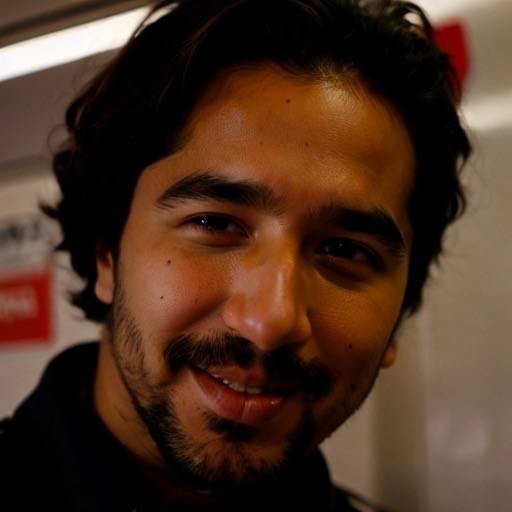} &
            \includegraphics[width=\linewidth]{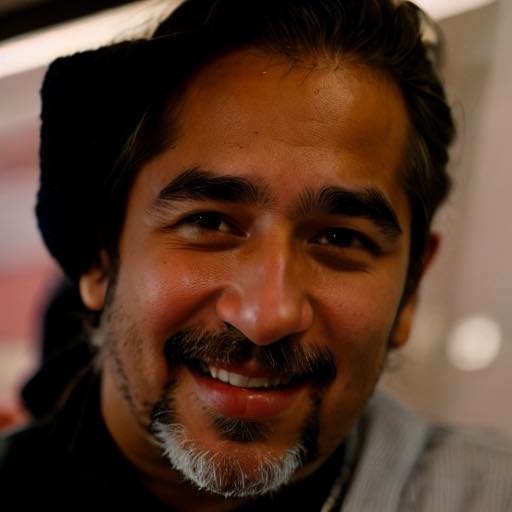} &
            \includegraphics[width=\linewidth]{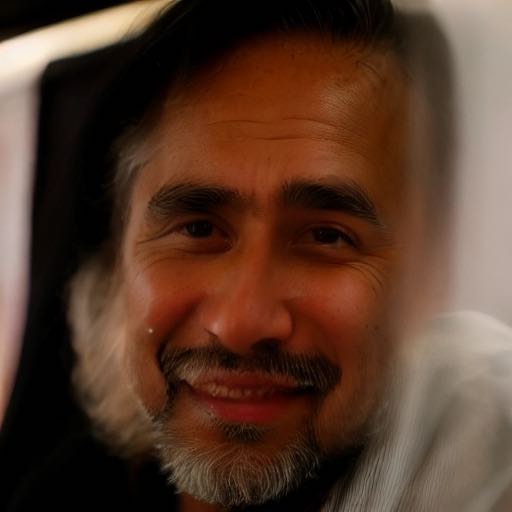} &
            \includegraphics[width=\linewidth]{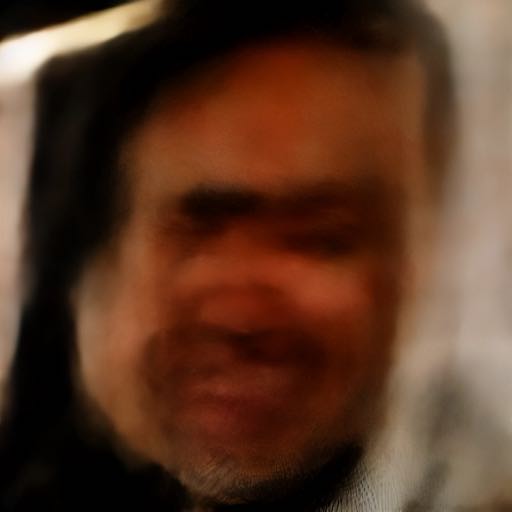} \\
        \end{tabularx}
        \caption{epiCRealism \cite{epicrealism}: A close-up of a man talking and laughing on New York subway. (Our method generates sharper details in 2 steps and 1 step.)}
    \end{subfigure}

    \vspace{10pt}

    \begin{subfigure}[b]{\textwidth}
        \centering
        \setlength\tabcolsep{0pt}
        \renewcommand{\arraystretch}{0}
        \begin{tabularx}{\textwidth}{@{}X@{\hskip 4pt}XXXX@{\hskip 4pt}XXXX@{}}
            \includegraphics[width=\linewidth]{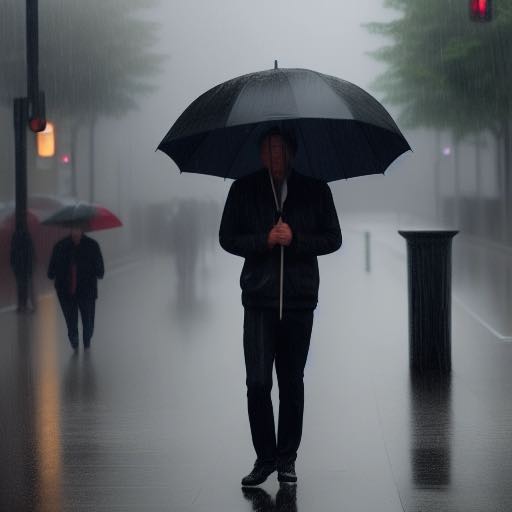} &
            \includegraphics[width=\linewidth]{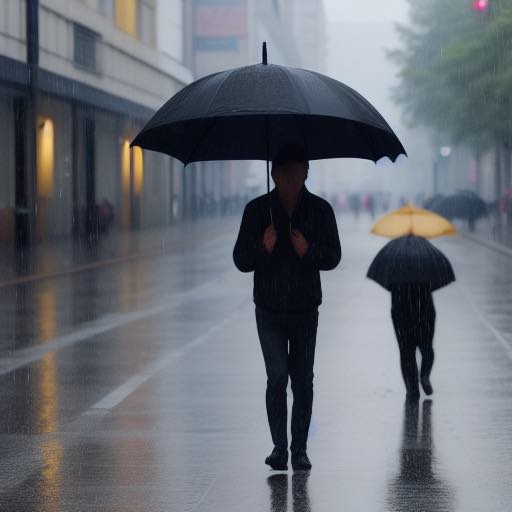} &
            \includegraphics[width=\linewidth]{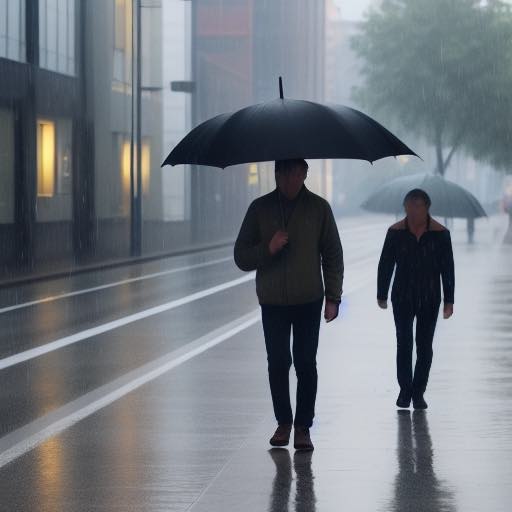} &
            \includegraphics[width=\linewidth]{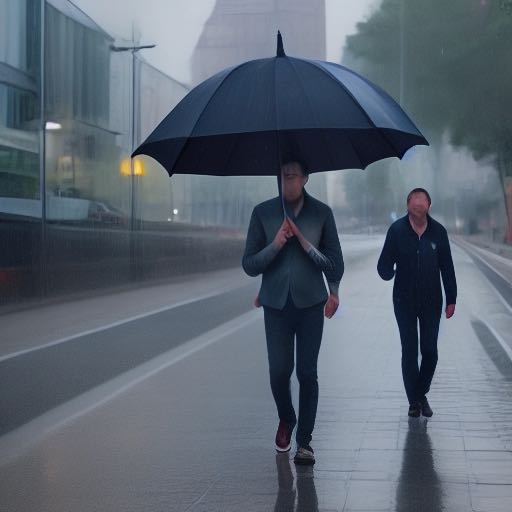} &
            \includegraphics[width=\linewidth]{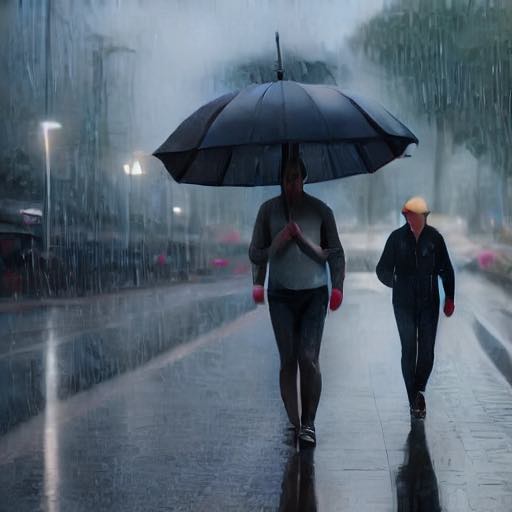} &
            \includegraphics[width=\linewidth]{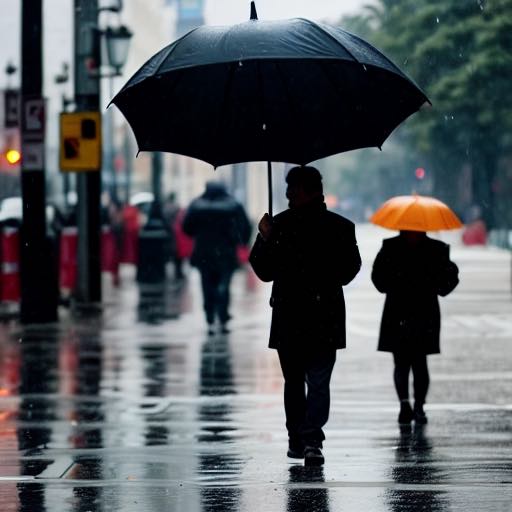} &
            \includegraphics[width=\linewidth]{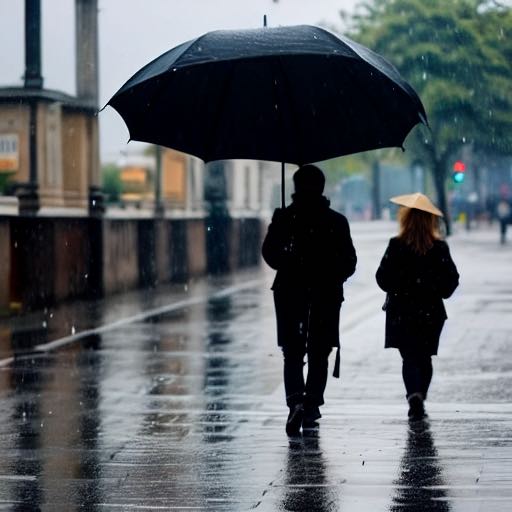} &
            \includegraphics[width=\linewidth]{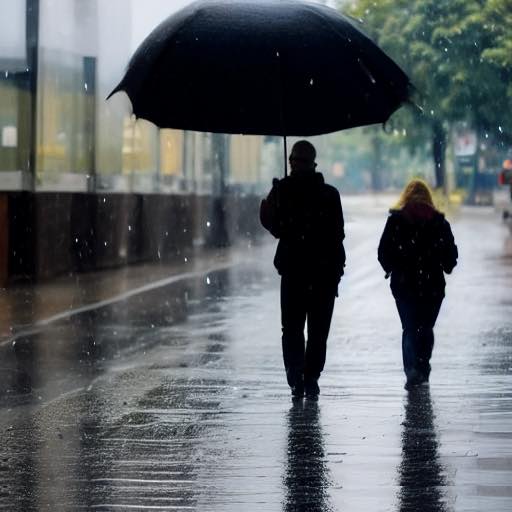} &
            \includegraphics[width=\linewidth]{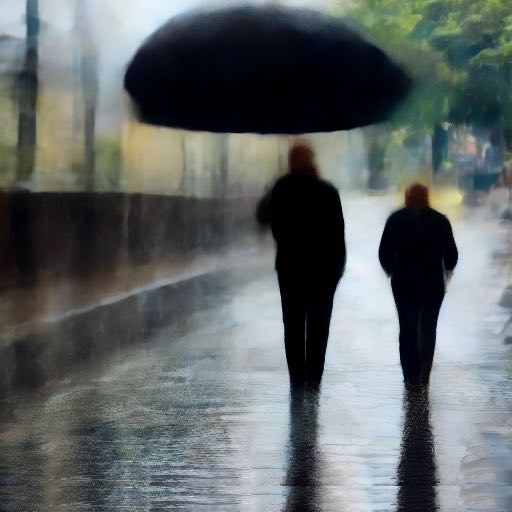} \\
            \includegraphics[width=\linewidth]{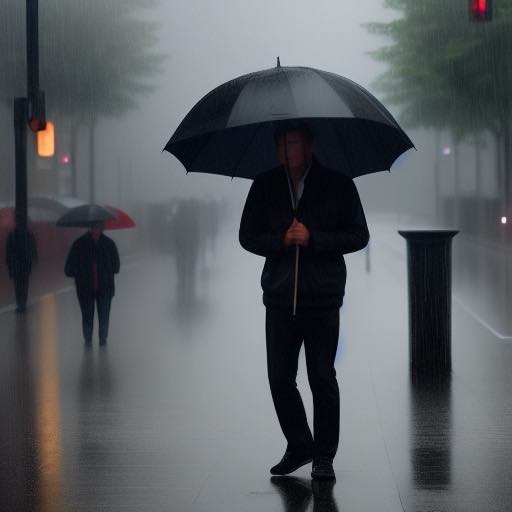} &
            \includegraphics[width=\linewidth]{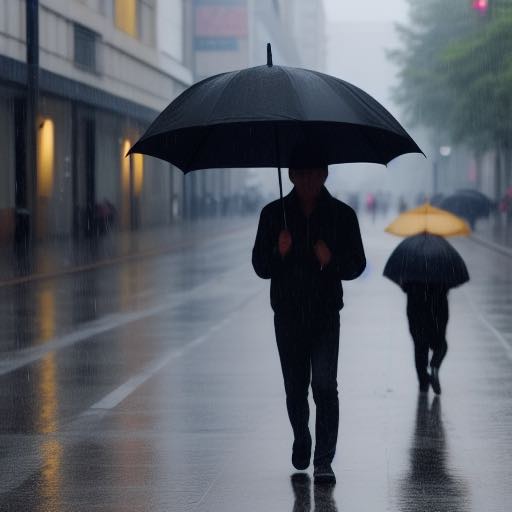} &
            \includegraphics[width=\linewidth]{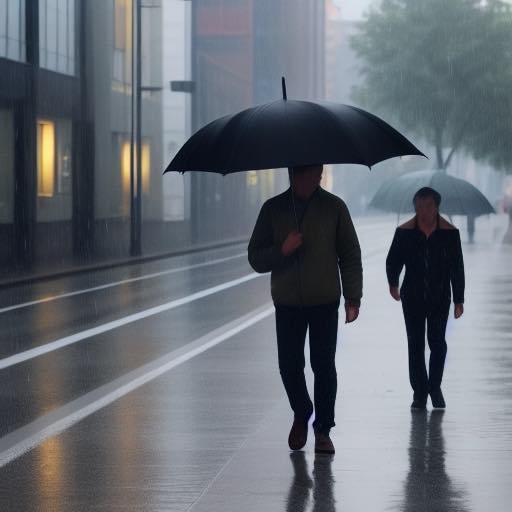} &
            \includegraphics[width=\linewidth]{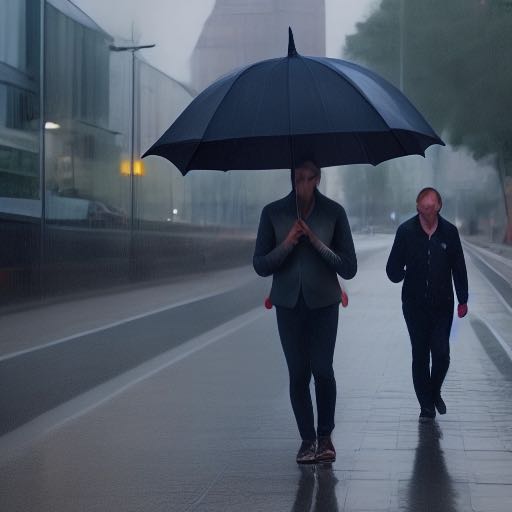} &
            \includegraphics[width=\linewidth]{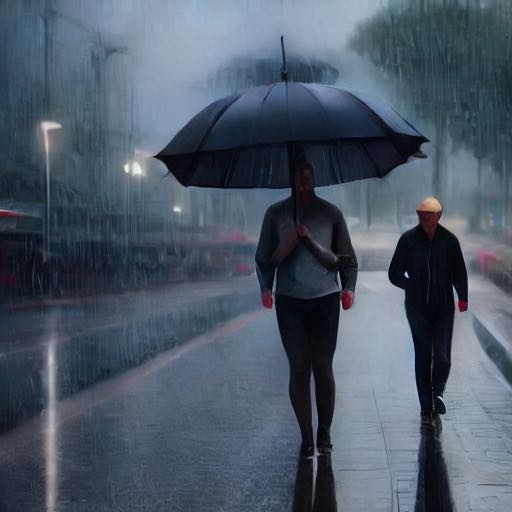} &
            \includegraphics[width=\linewidth]{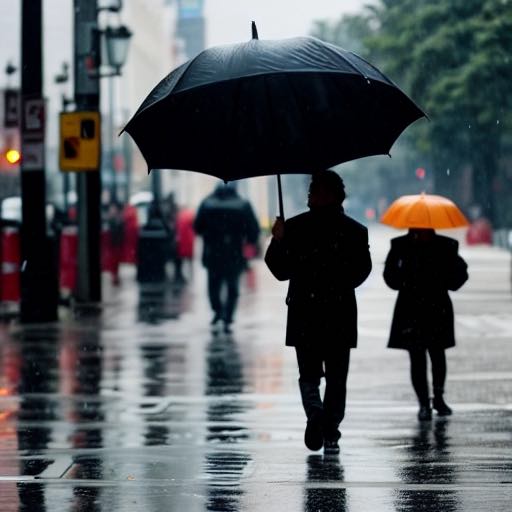} &
            \includegraphics[width=\linewidth]{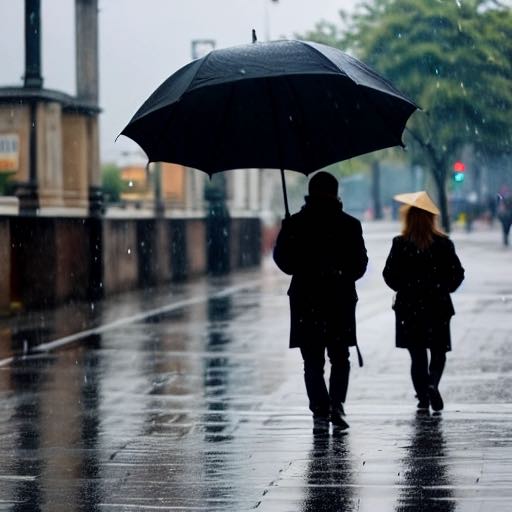} &
            \includegraphics[width=\linewidth]{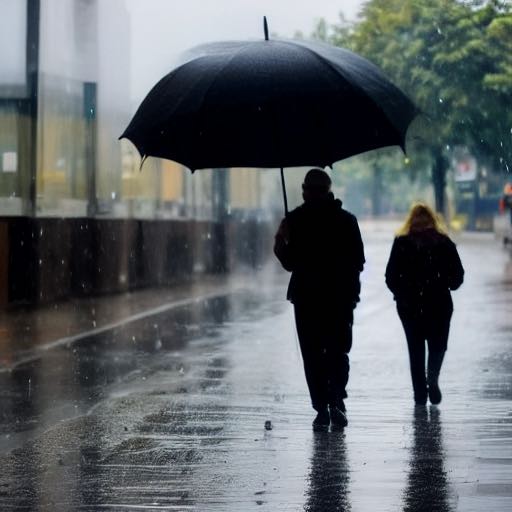} &
            \includegraphics[width=\linewidth]{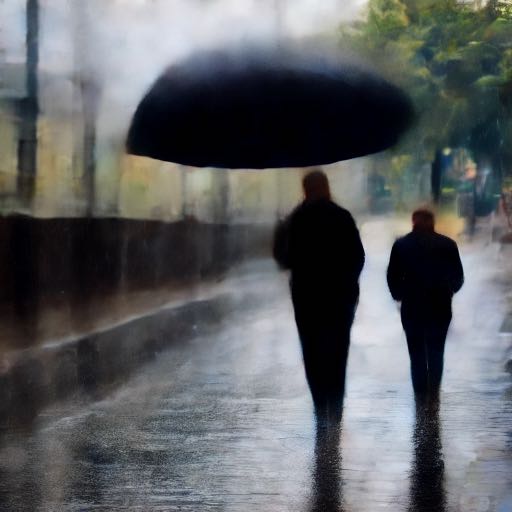} \\
            \includegraphics[width=\linewidth]{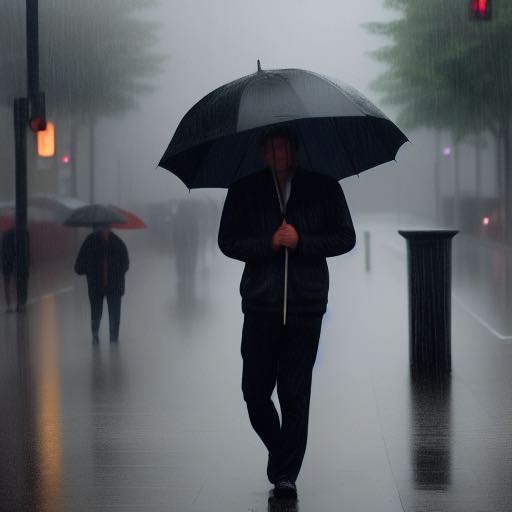} &
            \includegraphics[width=\linewidth]{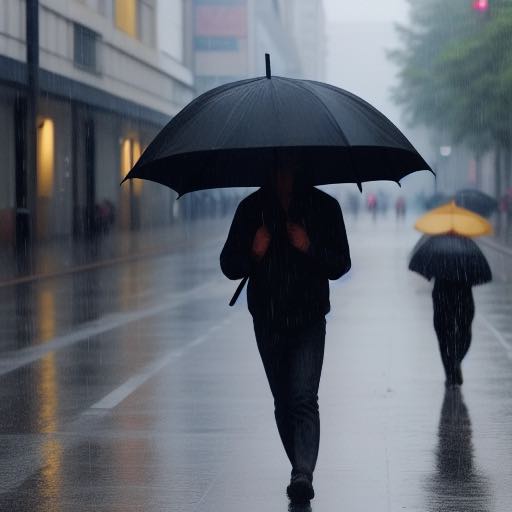} &
            \includegraphics[width=\linewidth]{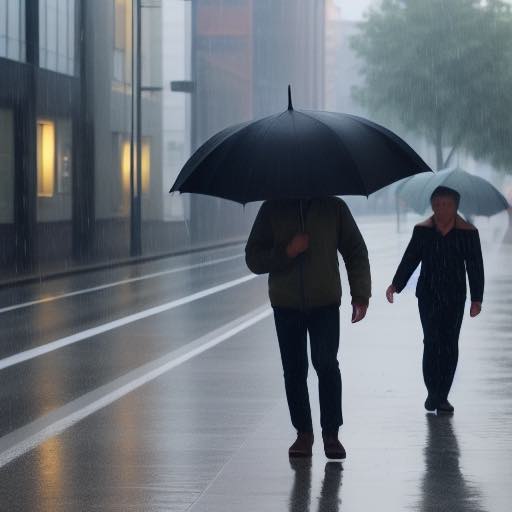} &
            \includegraphics[width=\linewidth]{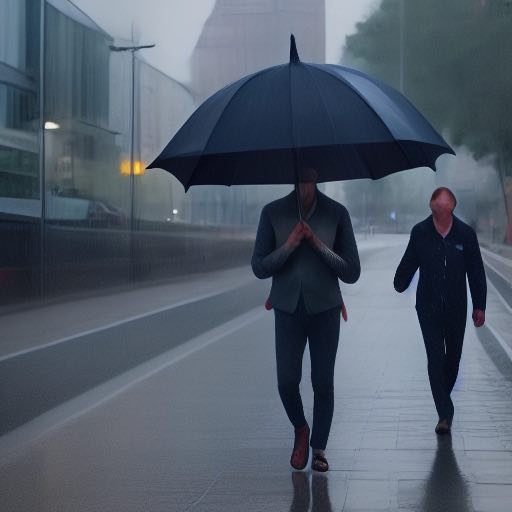} &
            \includegraphics[width=\linewidth]{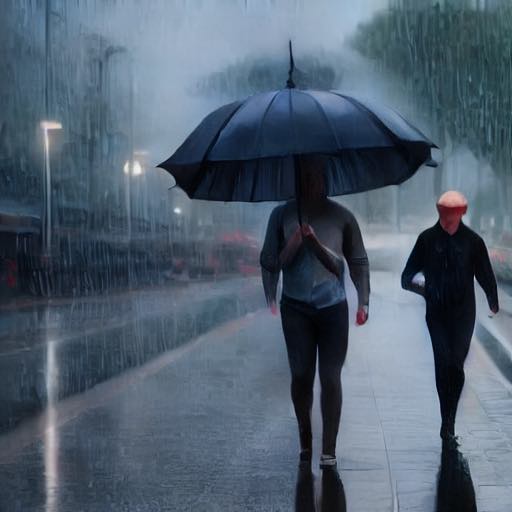} &
            \includegraphics[width=\linewidth]{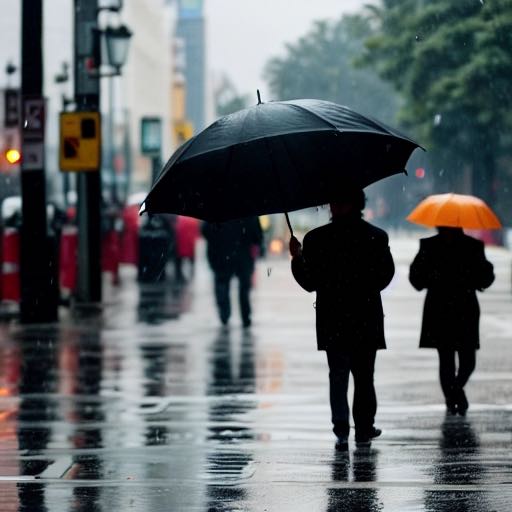} &
            \includegraphics[width=\linewidth]{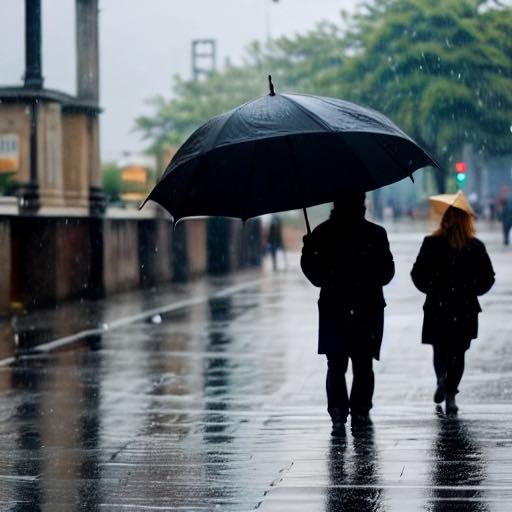} &
            \includegraphics[width=\linewidth]{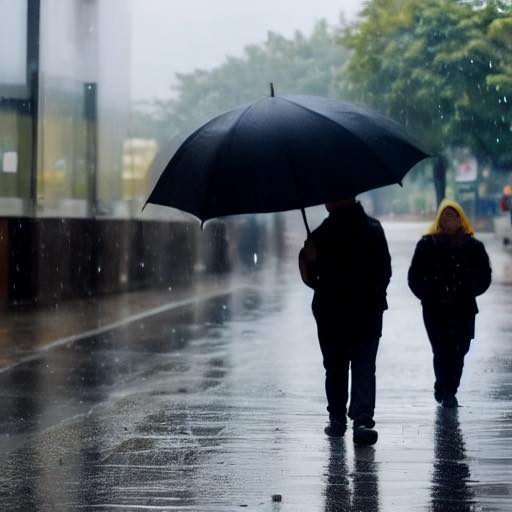} &
            \includegraphics[width=\linewidth]{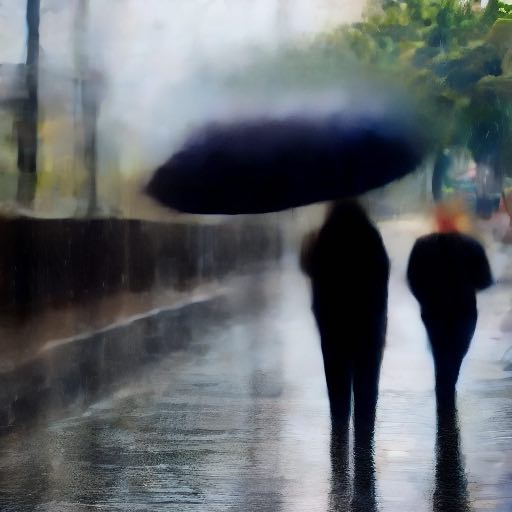} \\
        \end{tabularx}
        \caption{RealisticVision v5.1 \cite{realisticvision}: A man holding a black umbrella running in a rainy day. (Our method matches the original tone and style better.)}
    \end{subfigure}

    \vspace{10pt}

    \begin{subfigure}[b]{\textwidth}
        \centering
        \setlength\tabcolsep{0pt}
        \renewcommand{\arraystretch}{0}
        \begin{tabularx}{\textwidth}{@{}X@{\hskip 4pt}XXXX@{\hskip 4pt}XXXX@{}}
            \includegraphics[width=\linewidth]{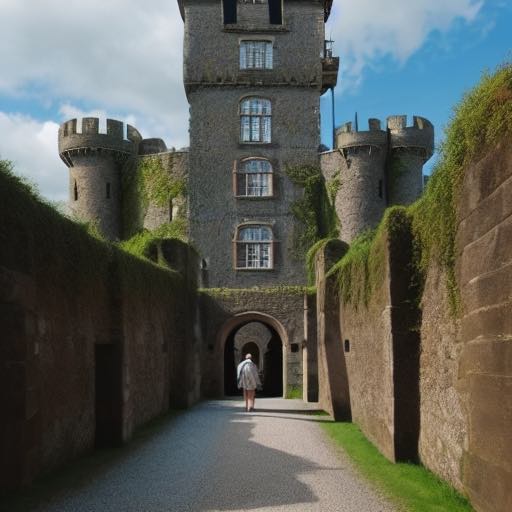} &
            \includegraphics[width=\linewidth]{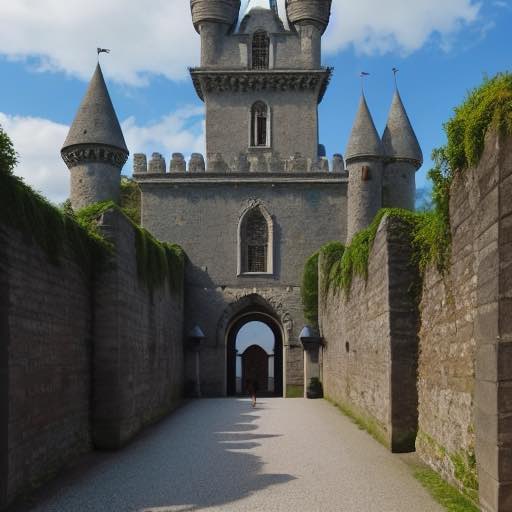} &
            \includegraphics[width=\linewidth]{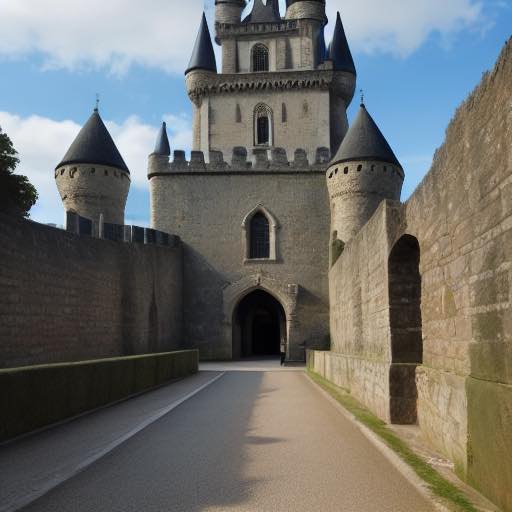} &
            \includegraphics[width=\linewidth]{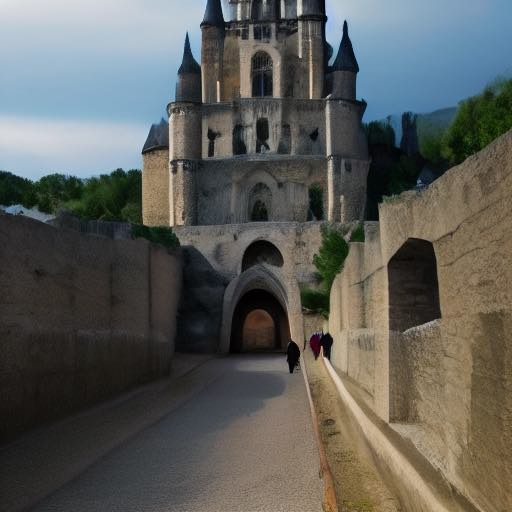} &
            \includegraphics[width=\linewidth]{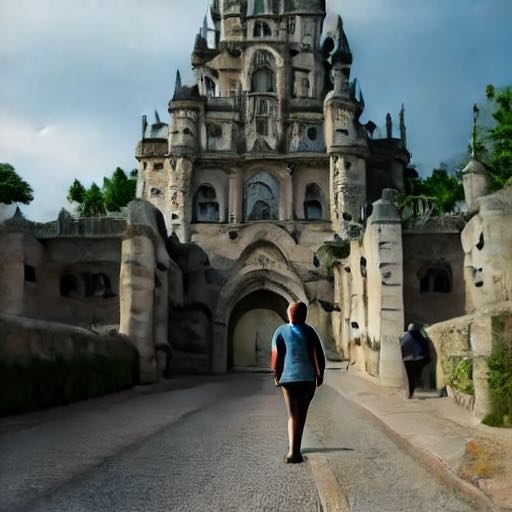} &
            \includegraphics[width=\linewidth]{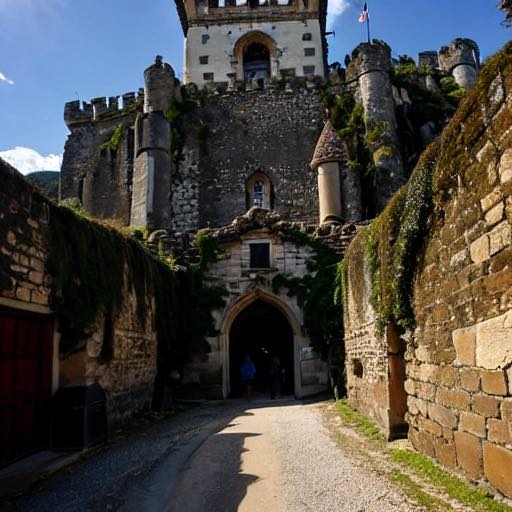} &
            \includegraphics[width=\linewidth]{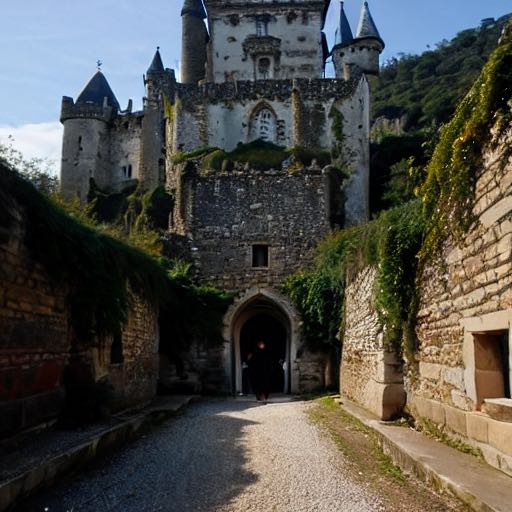} &
            \includegraphics[width=\linewidth]{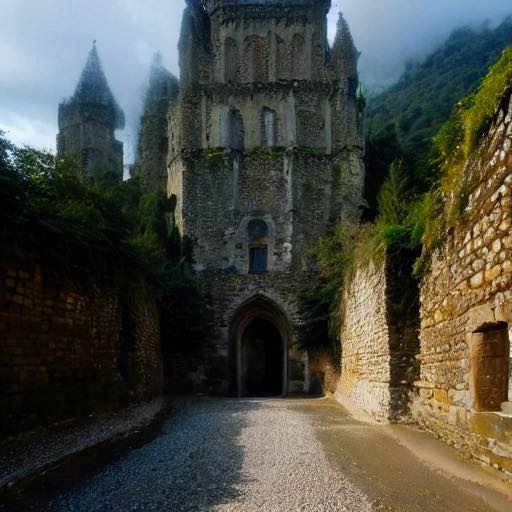} &
            \includegraphics[width=\linewidth]{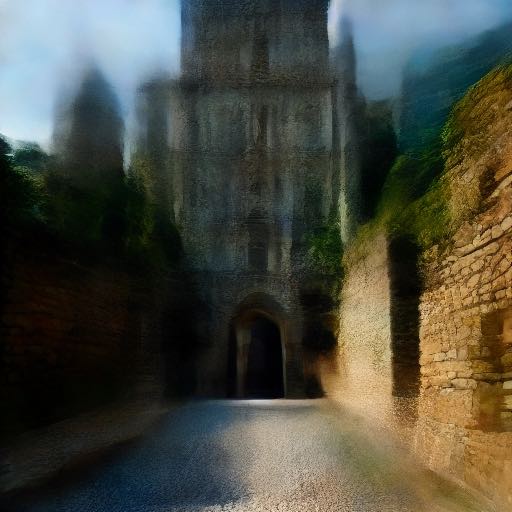} \\
            \includegraphics[width=\linewidth]{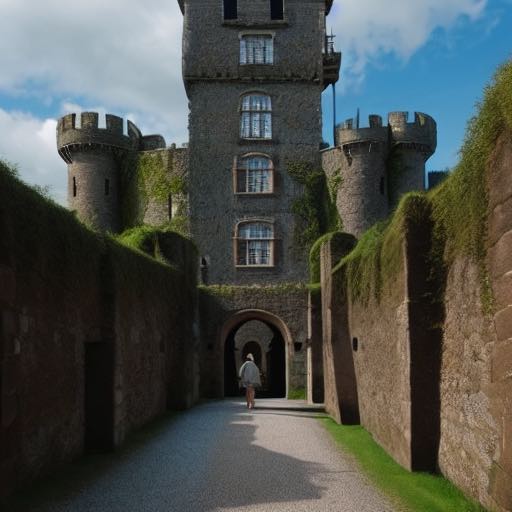} &
            \includegraphics[width=\linewidth]{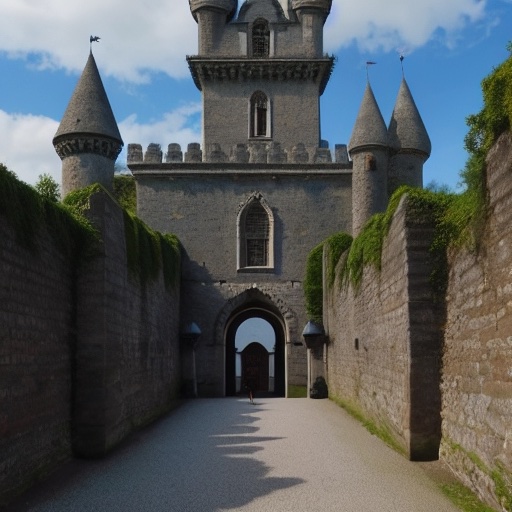} &
            \includegraphics[width=\linewidth]{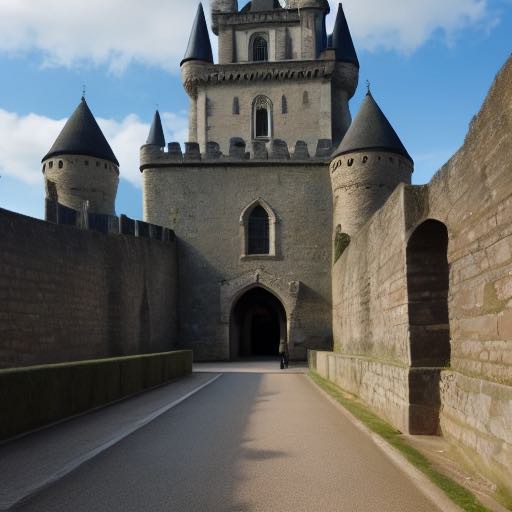} &
            \includegraphics[width=\linewidth]{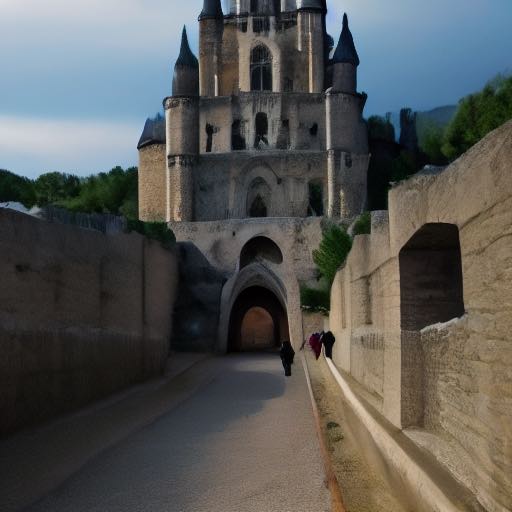} &
            \includegraphics[width=\linewidth]{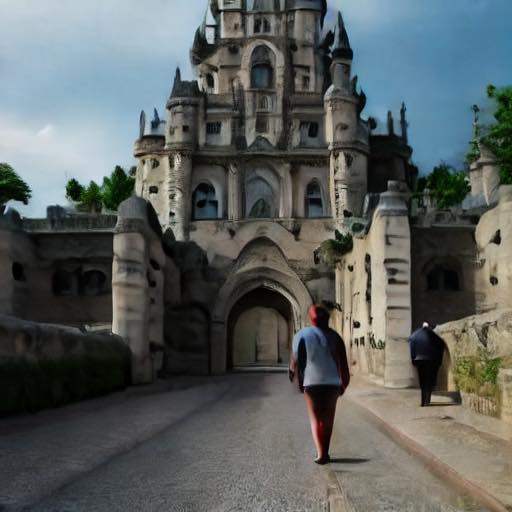} &
            \includegraphics[width=\linewidth]{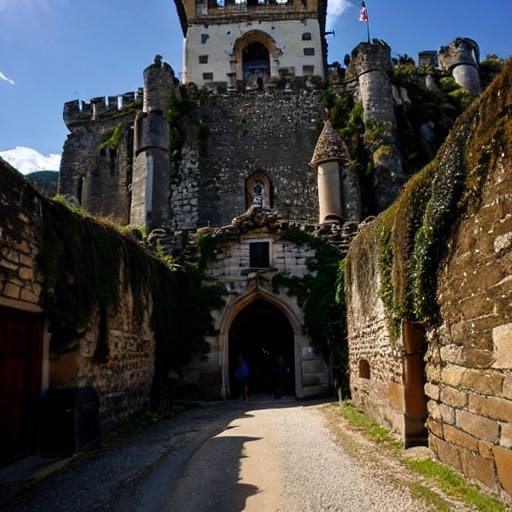} &
            \includegraphics[width=\linewidth]{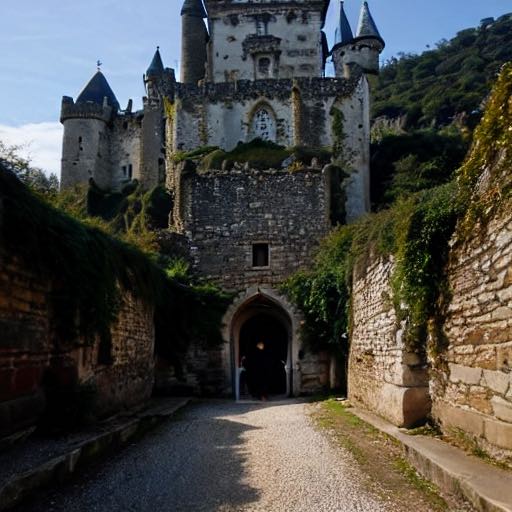} &
            \includegraphics[width=\linewidth]{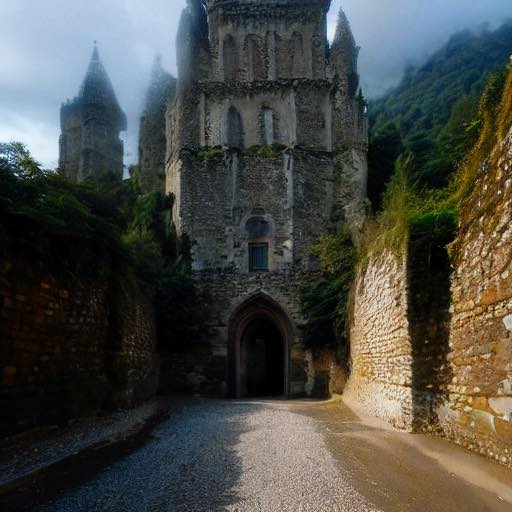} &
            \includegraphics[width=\linewidth]{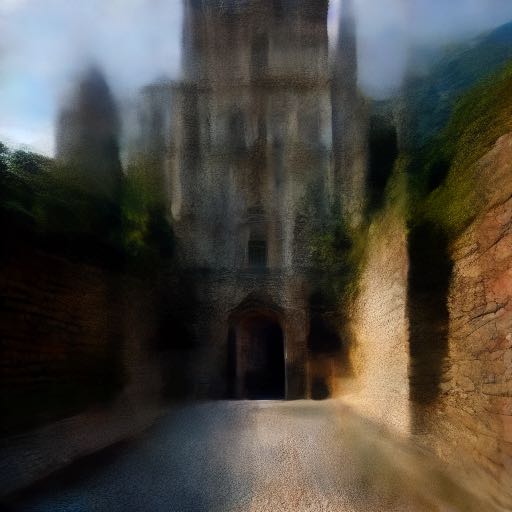} \\
            \includegraphics[width=\linewidth]{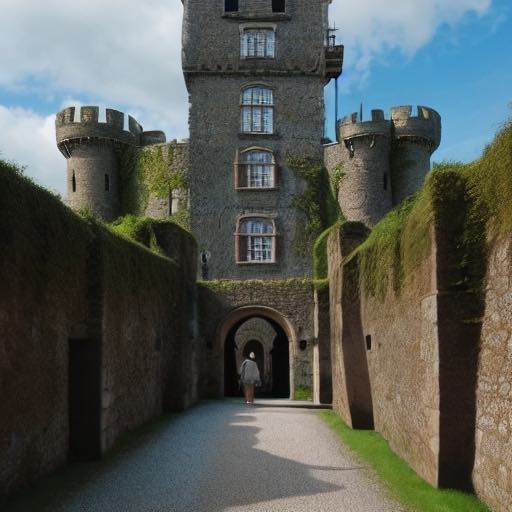} &
            \includegraphics[width=\linewidth]{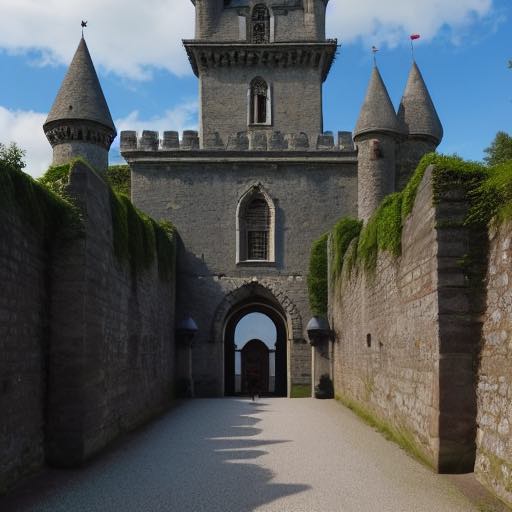} &
            \includegraphics[width=\linewidth]{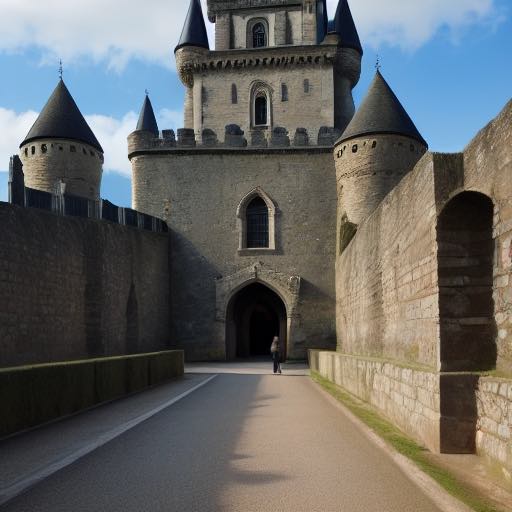} &
            \includegraphics[width=\linewidth]{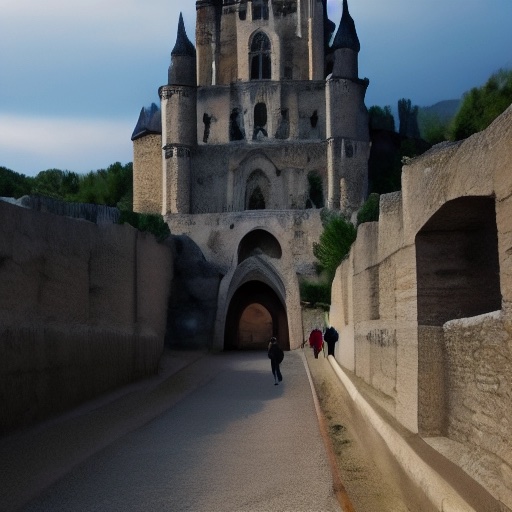} &
            \includegraphics[width=\linewidth]{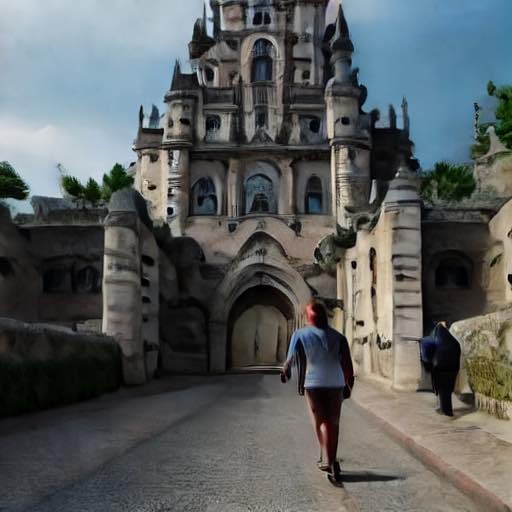} &
            \includegraphics[width=\linewidth]{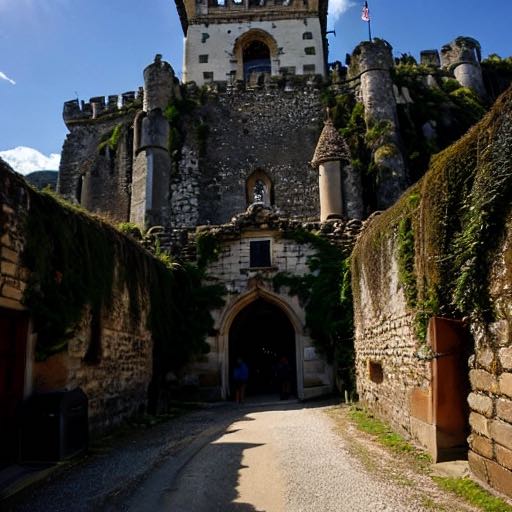} &
            \includegraphics[width=\linewidth]{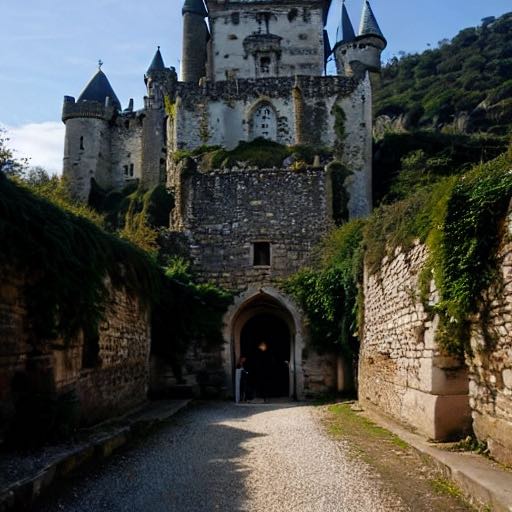} &
            \includegraphics[width=\linewidth]{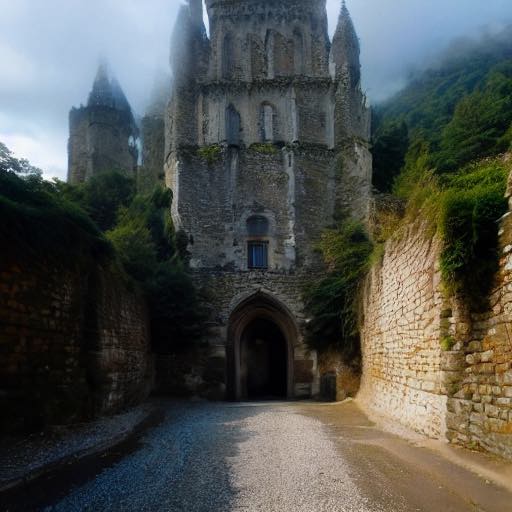} &
            \includegraphics[width=\linewidth]{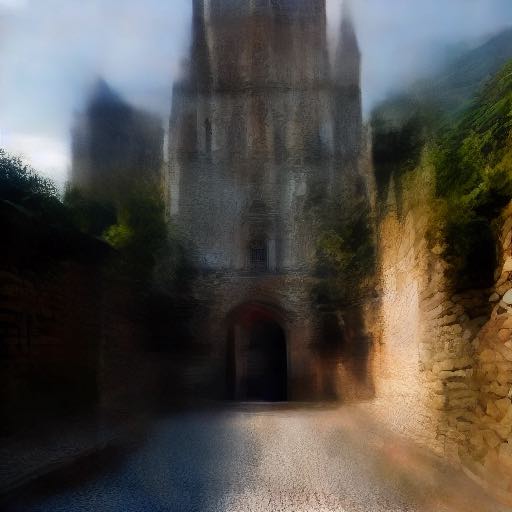}\\
        \end{tabularx}
        \caption{epiCRealism \cite{epicrealism}: Entering a big castle. (Our method generates sharper details in 2 steps and 1 step.)}
    \end{subfigure}
    
    \caption{Qualitative Comparison. We only show the first, middle, and last frames of the generated video clips in each column. Our model generates better results using 1-step, 2-step, and 4-step inference. Additionally, our model can better retain the style of the original model. This page focuses on realistic style generation. Please see the next page for anime-style generation.}
    \label{fig:qualitative}
\end{figure*}

\begin{figure*}
    \ContinuedFloat
    \centering
    \captionsetup{justification=raggedright,singlelinecheck=false}
    \small
    \setlength\tabcolsep{2pt}
    \begin{tabularx}{\textwidth}{|X@{\hskip 6pt}|X|X|X|X@{\hskip 6pt}|X|X|X|X|}
        \normalsize{\textbf{Original}} \cite{guo2023animatediff} & \multicolumn{4}{l|}{\normalsize{\textbf{Ours}}} & \multicolumn{4}{l|}{\normalsize{\textbf{AnimateLCM}} \cite{wang2024animatelcm}} \\
        \footnotesize{CFG7.5} & \multicolumn{4}{l|}{\footnotesize{No CFG}} & \multicolumn{4}{l|}{\footnotesize{No CFG}} \\
        32 Steps & 8 Steps & 4 Steps & 2 Steps & 1 Step & 8 Steps & 4 Step & 2 Steps & 1 Step \\
    \end{tabularx}

    \begin{subfigure}[b]{\textwidth}
        \centering
        \setlength\tabcolsep{0pt}
        \renewcommand{\arraystretch}{0}
        \begin{tabularx}{\textwidth}{@{}X@{\hskip 4pt}XXXX@{\hskip 4pt}XXXX@{}}
            \includegraphics[width=\linewidth]{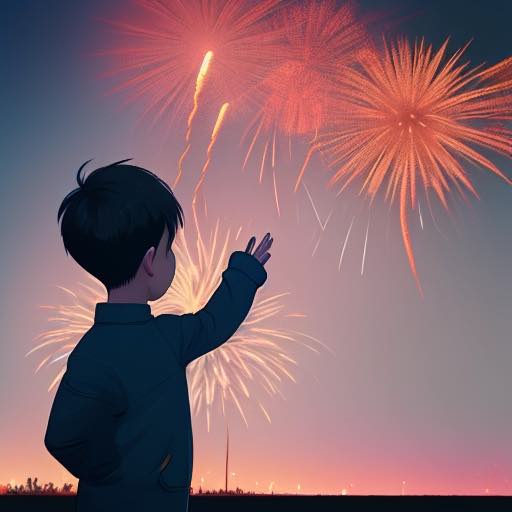} &
            \includegraphics[width=\linewidth]{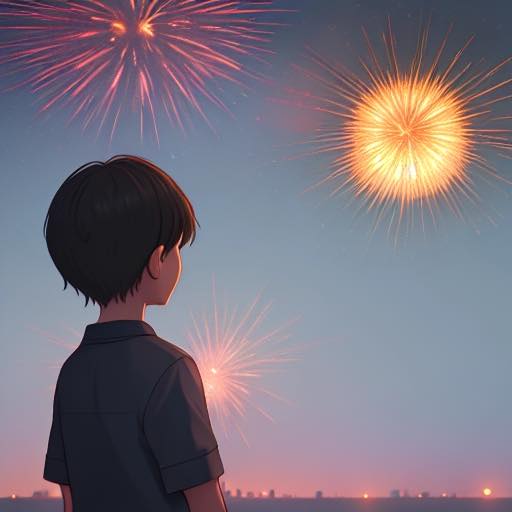} &
            \includegraphics[width=\linewidth]{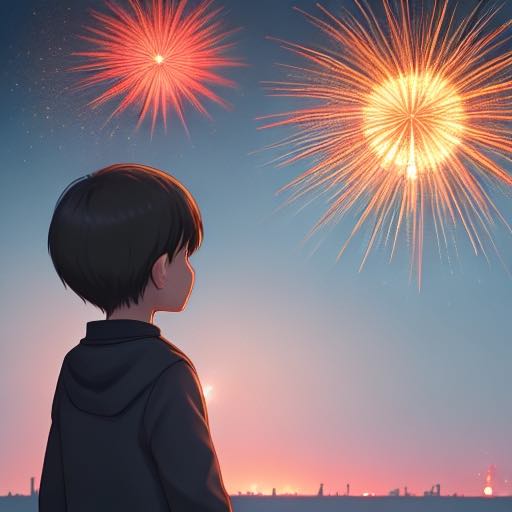} &
            \includegraphics[width=\linewidth]{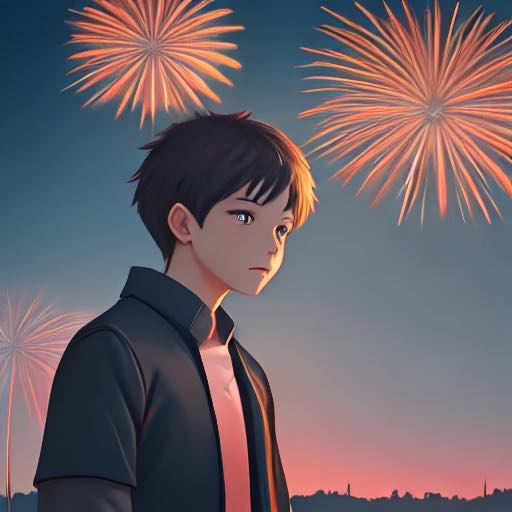} &
            \includegraphics[width=\linewidth]{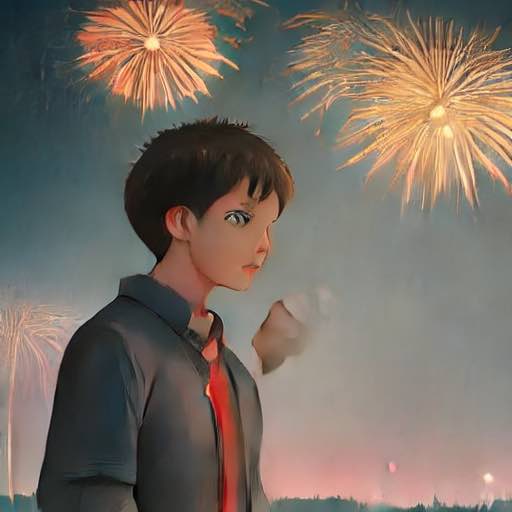} &
            \includegraphics[width=\linewidth]{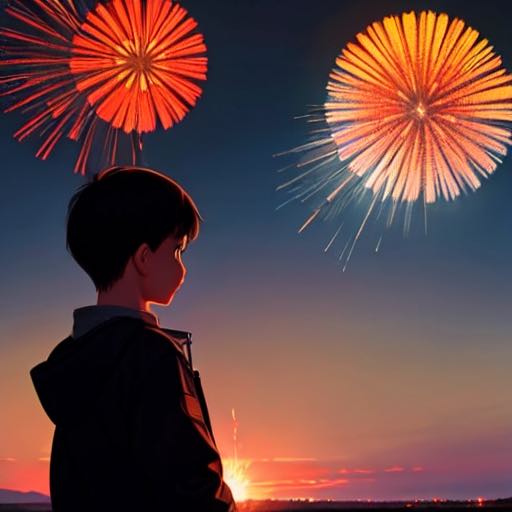} &
            \includegraphics[width=\linewidth]{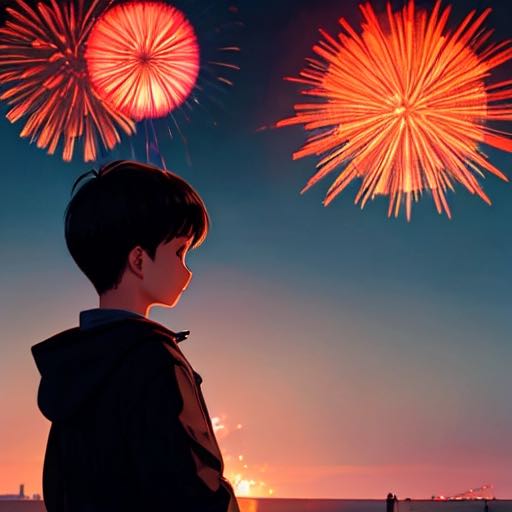} &
            \includegraphics[width=\linewidth]{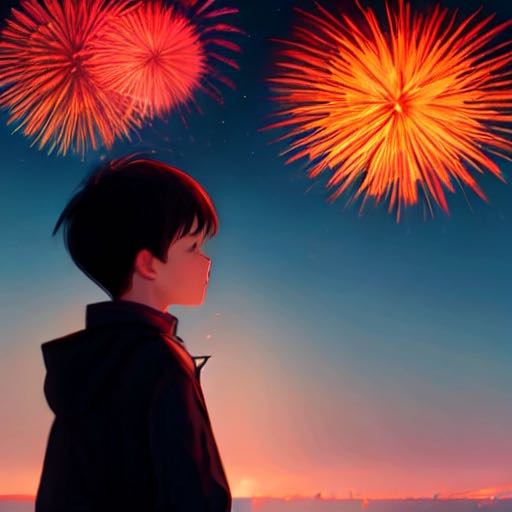} &
            \includegraphics[width=\linewidth]{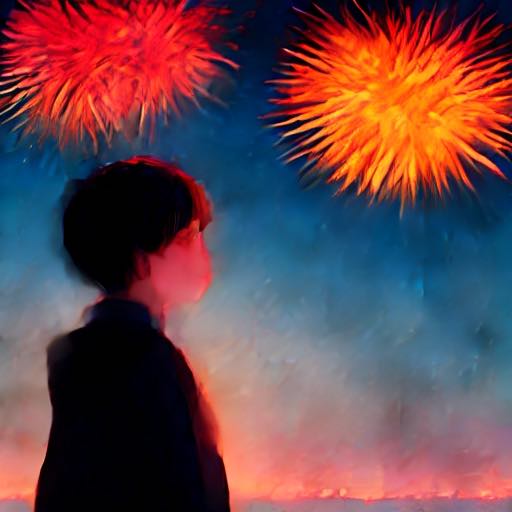} \\
            \includegraphics[width=\linewidth]{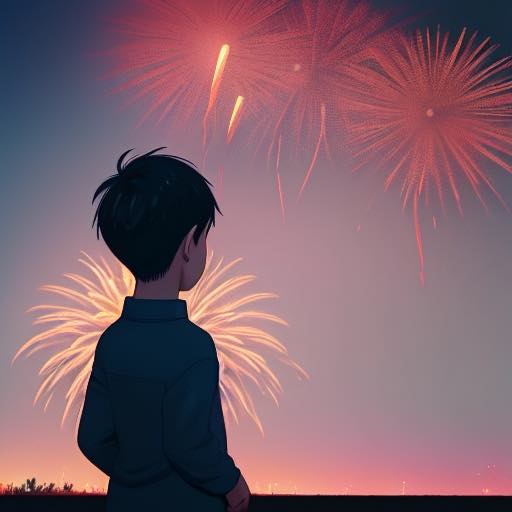} &
            \includegraphics[width=\linewidth]{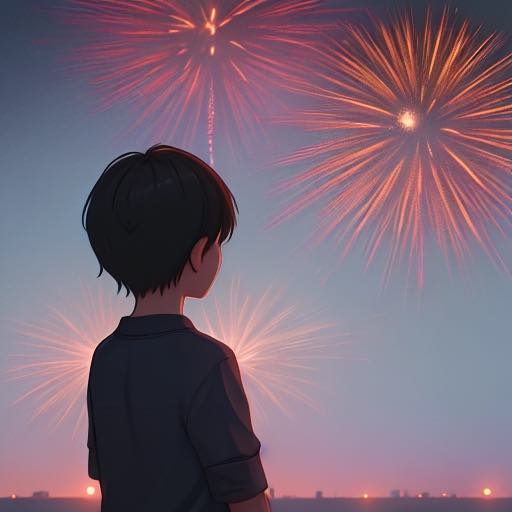} &
            \includegraphics[width=\linewidth]{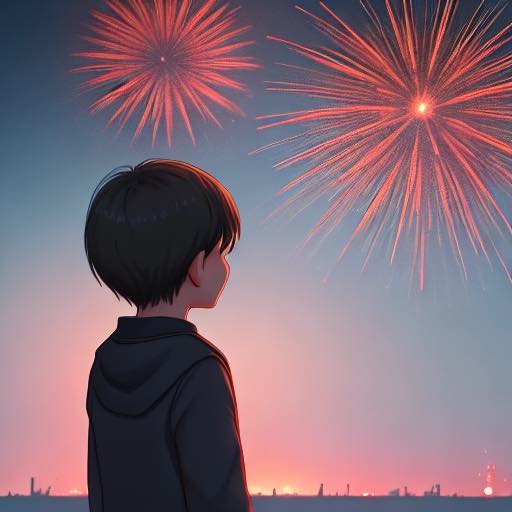} &
            \includegraphics[width=\linewidth]{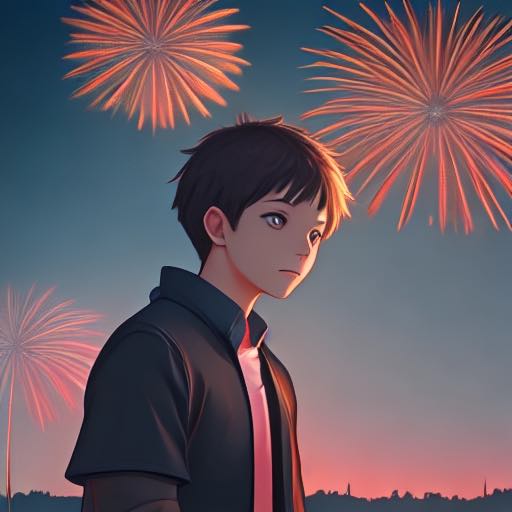} &
            \includegraphics[width=\linewidth]{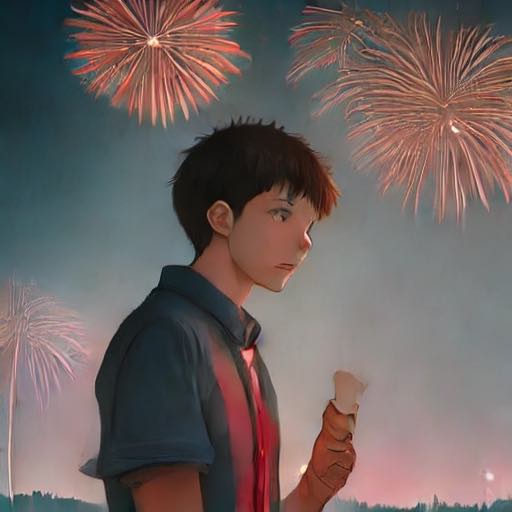} &
            \includegraphics[width=\linewidth]{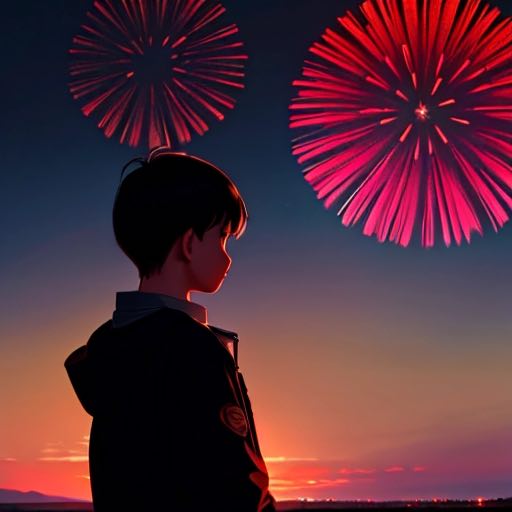} &
            \includegraphics[width=\linewidth]{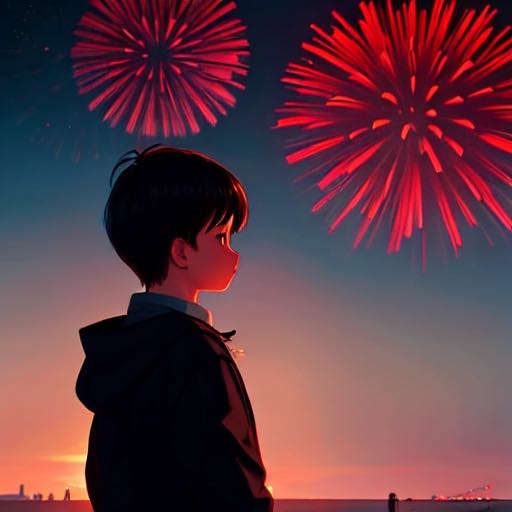} &
            \includegraphics[width=\linewidth]{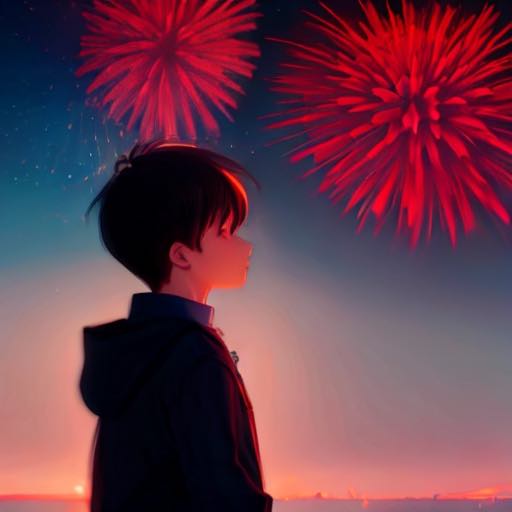} &
            \includegraphics[width=\linewidth]{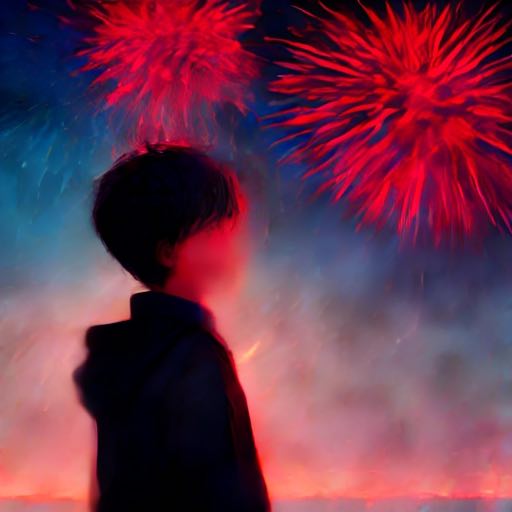} \\
            \includegraphics[width=\linewidth]{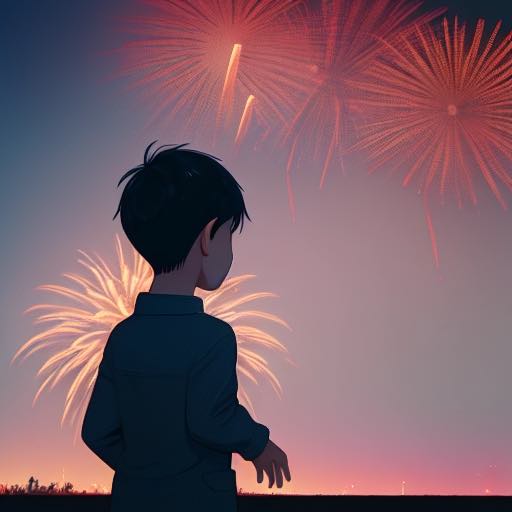} &
            \includegraphics[width=\linewidth]{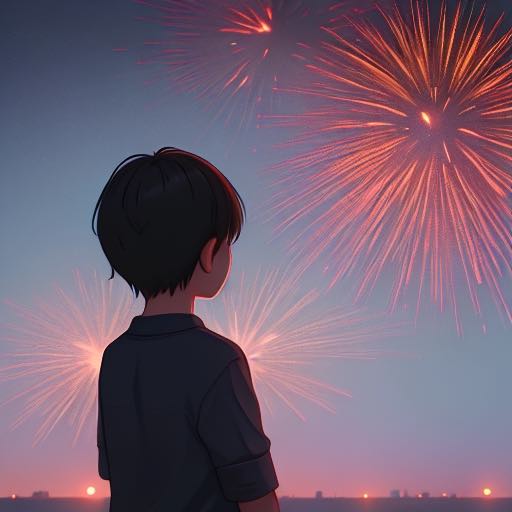} &
            \includegraphics[width=\linewidth]{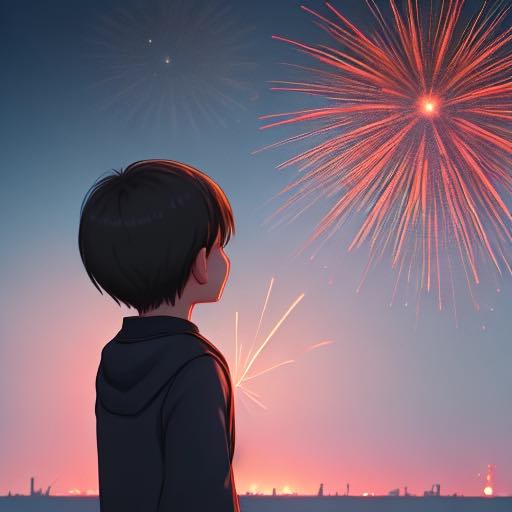} &
            \includegraphics[width=\linewidth]{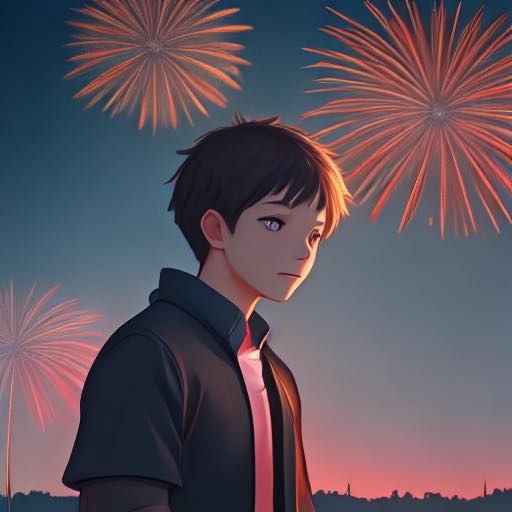} &
            \includegraphics[width=\linewidth]{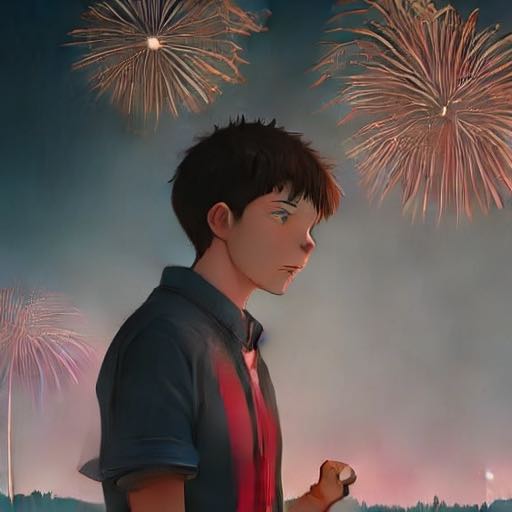} &
            \includegraphics[width=\linewidth]{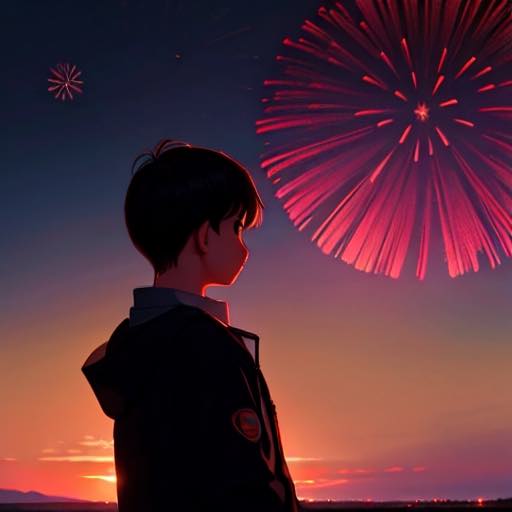} &
            \includegraphics[width=\linewidth]{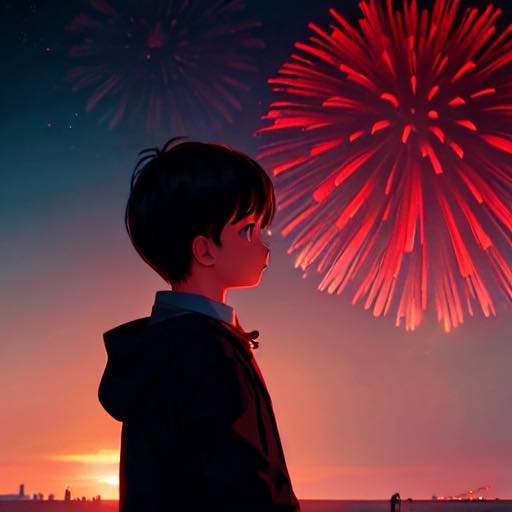} &
            \includegraphics[width=\linewidth]{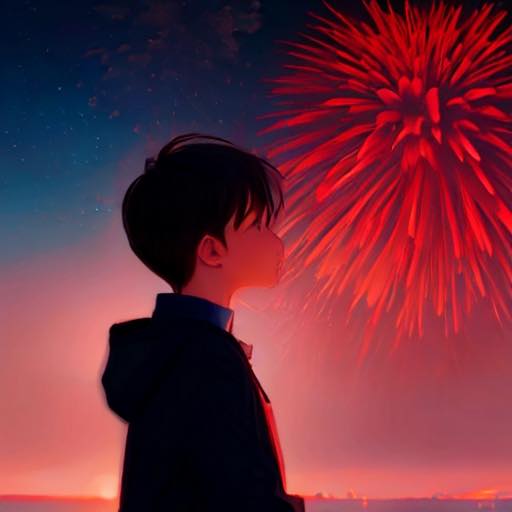} &
            \includegraphics[width=\linewidth]{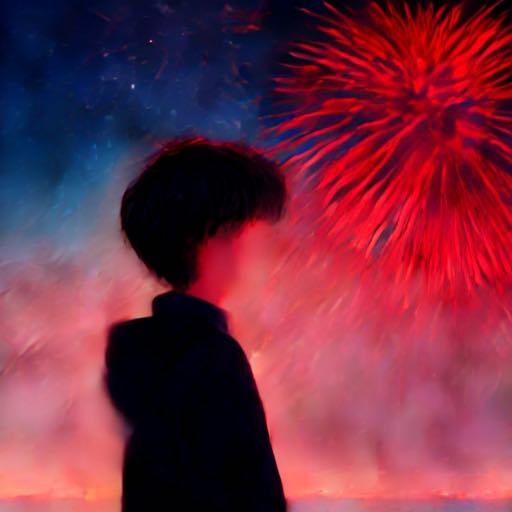} \\
        \end{tabularx}
        \caption{IMP v1.0 \cite{imp}: A boy looking at the sky, firework in the background. (Our method matches the original tone and style better.)}
    \end{subfigure}

    \vspace{10pt}

    \begin{subfigure}[b]{\textwidth}
        \centering
        \setlength\tabcolsep{0pt}
        \renewcommand{\arraystretch}{0}
        \begin{tabularx}{\textwidth}{@{}X@{\hskip 4pt}XXXX@{\hskip 4pt}XXXX@{}}
            \includegraphics[width=\linewidth]{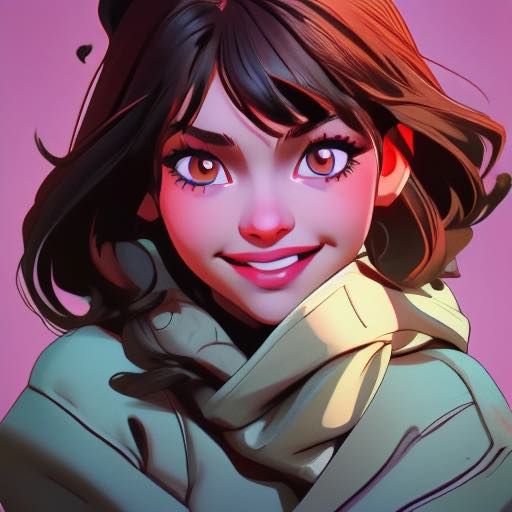} &
            \includegraphics[width=\linewidth]{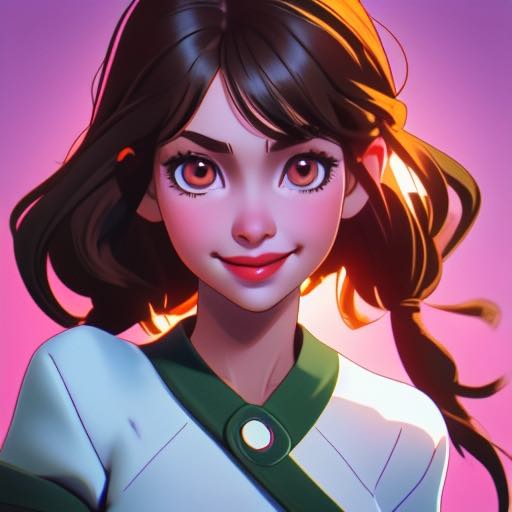} &
            \includegraphics[width=\linewidth]{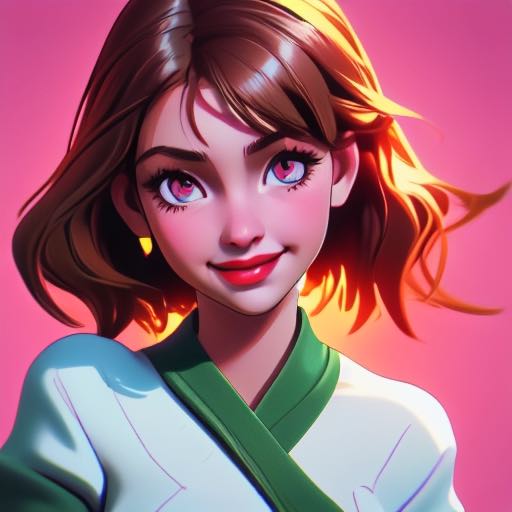} &
            \includegraphics[width=\linewidth]{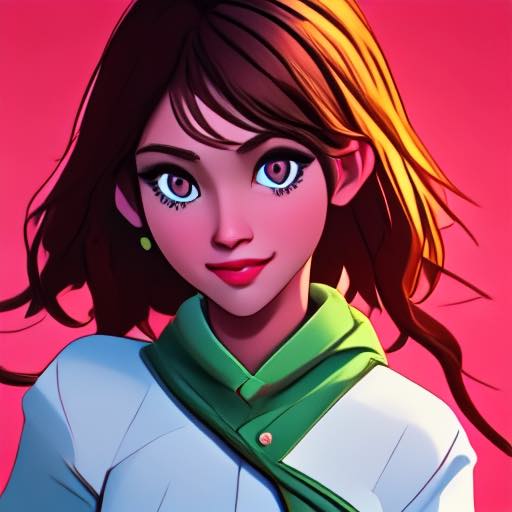} &
            \includegraphics[width=\linewidth]{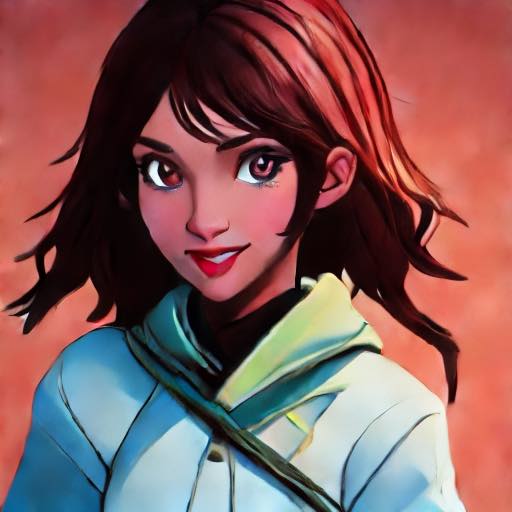} &
            \includegraphics[width=\linewidth]{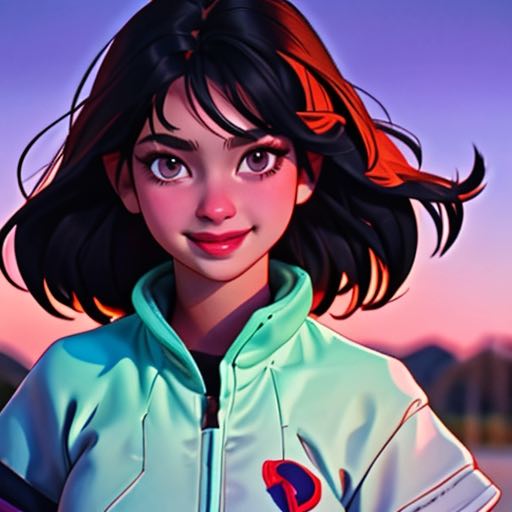} &
            \includegraphics[width=\linewidth]{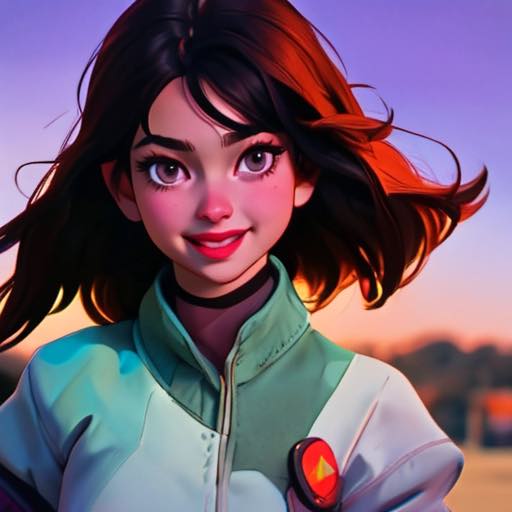} &
            \includegraphics[width=\linewidth]{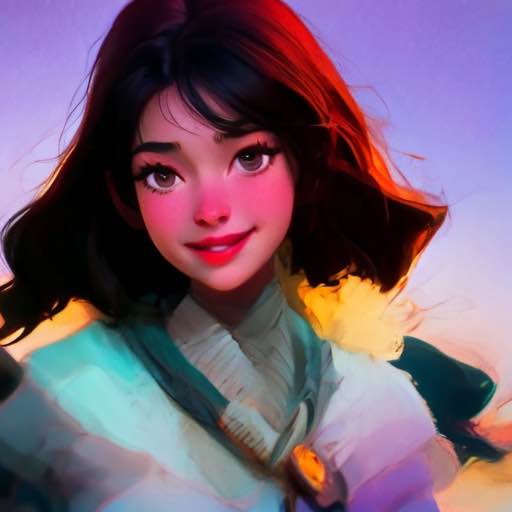} &
            \includegraphics[width=\linewidth]{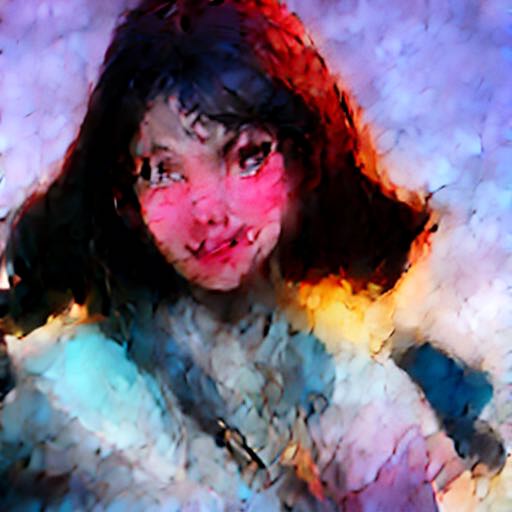} \\
            \includegraphics[width=\linewidth]{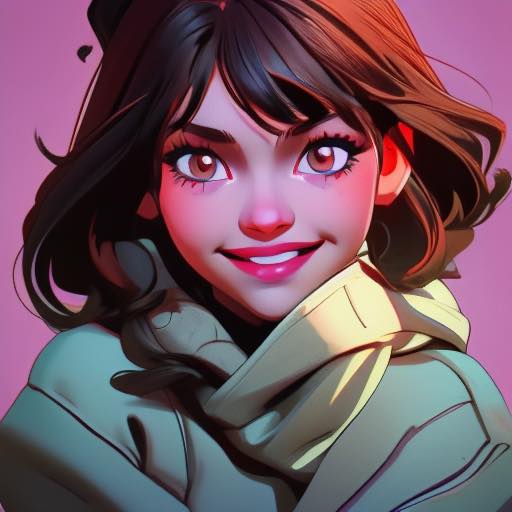} &
            \includegraphics[width=\linewidth]{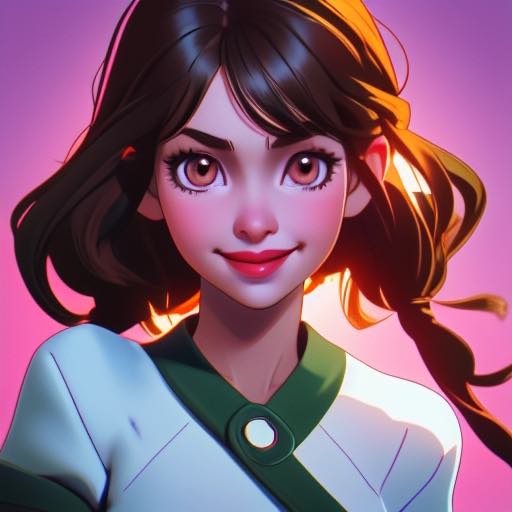} &
            \includegraphics[width=\linewidth]{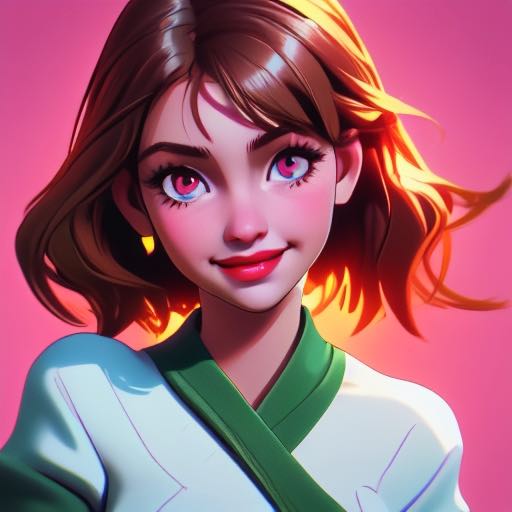} &
            \includegraphics[width=\linewidth]{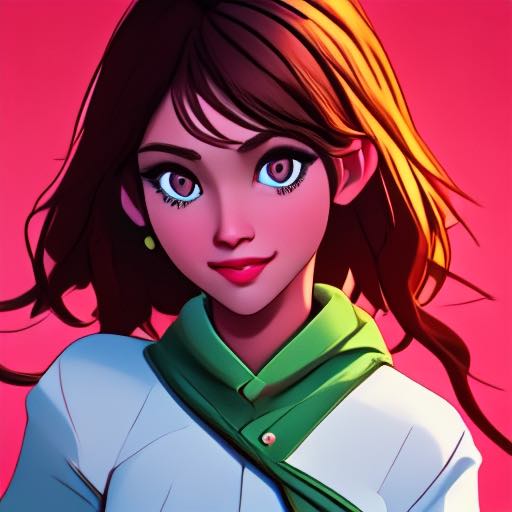} &
            \includegraphics[width=\linewidth]{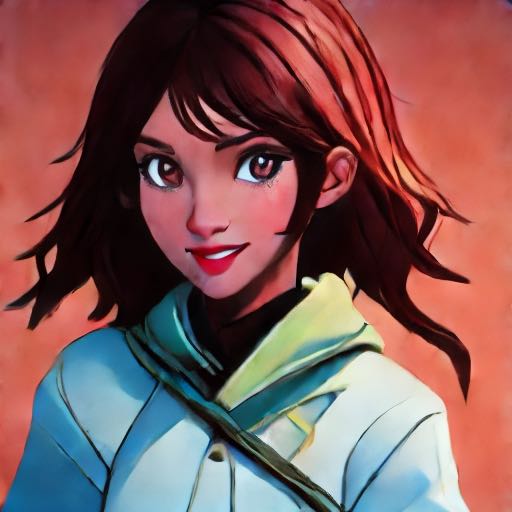} &
            \includegraphics[width=\linewidth]{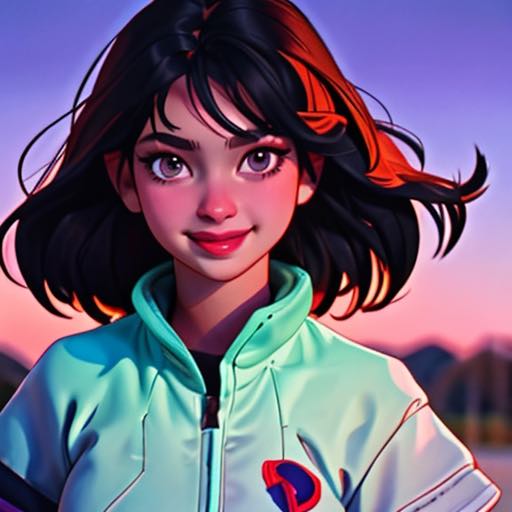} &
            \includegraphics[width=\linewidth]{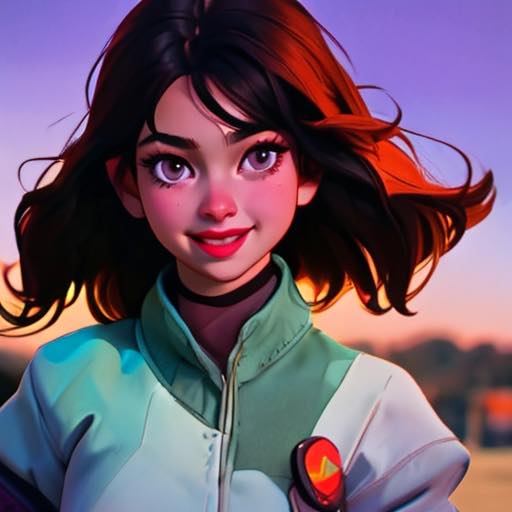} &
            \includegraphics[width=\linewidth]{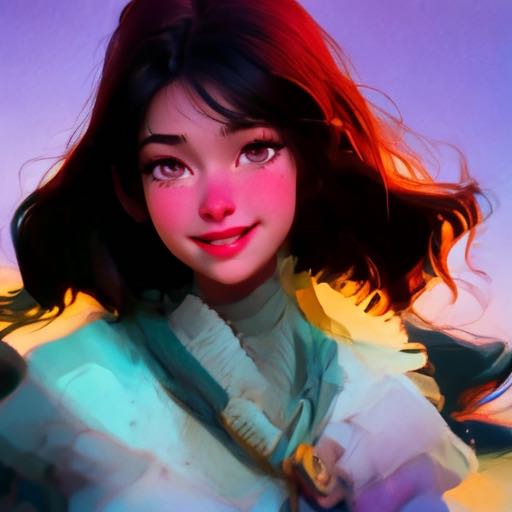} &
            \includegraphics[width=\linewidth]{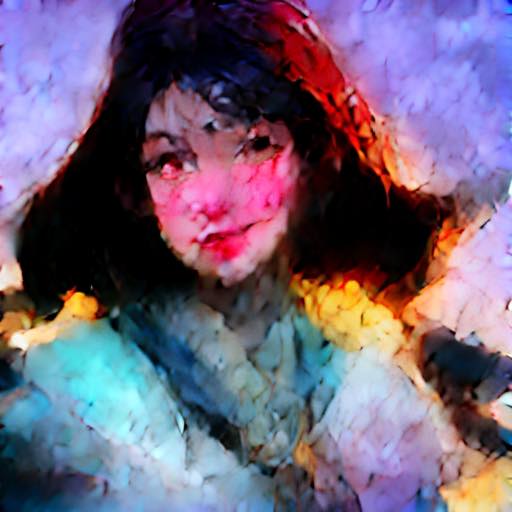} \\
            \includegraphics[width=\linewidth]{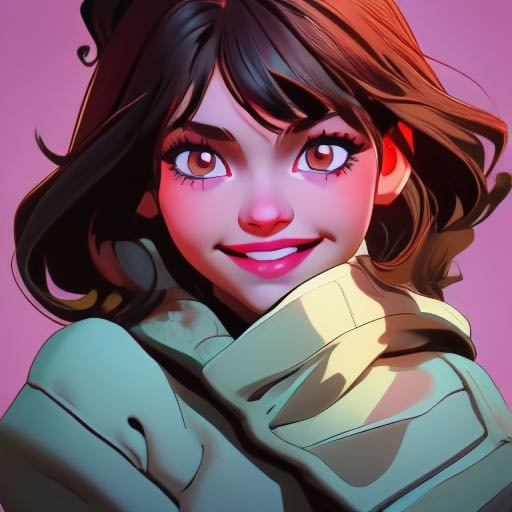} &
            \includegraphics[width=\linewidth]{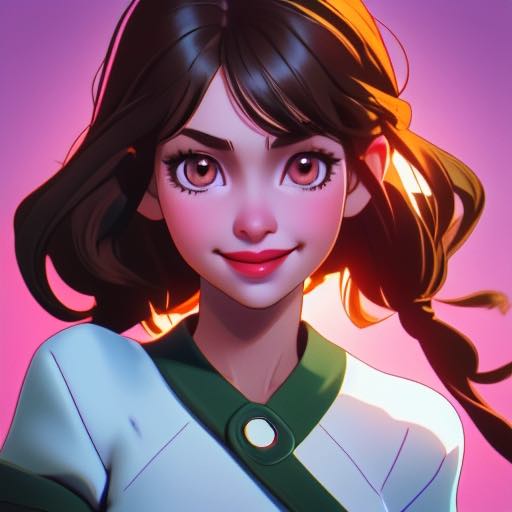} &
            \includegraphics[width=\linewidth]{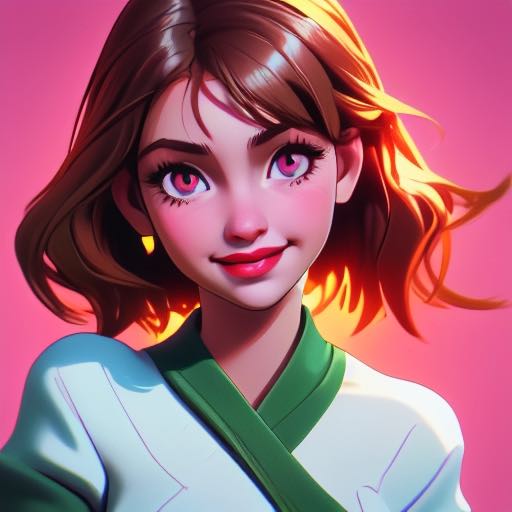} &
            \includegraphics[width=\linewidth]{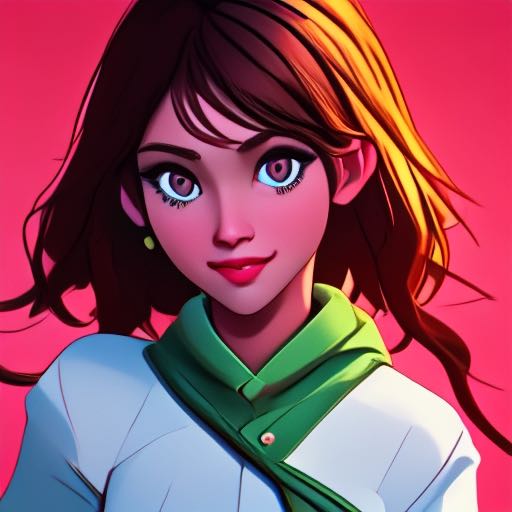} &
            \includegraphics[width=\linewidth]{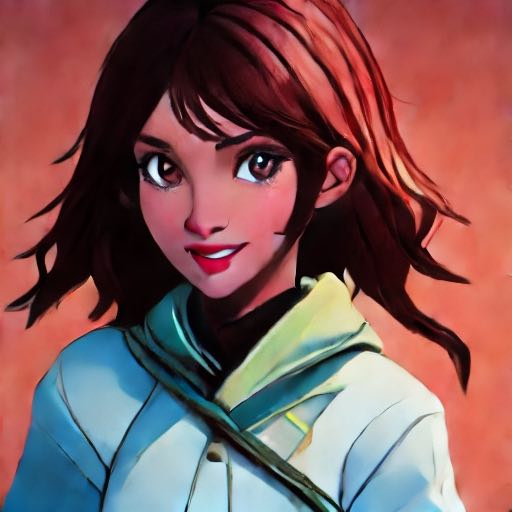} &
            \includegraphics[width=\linewidth]{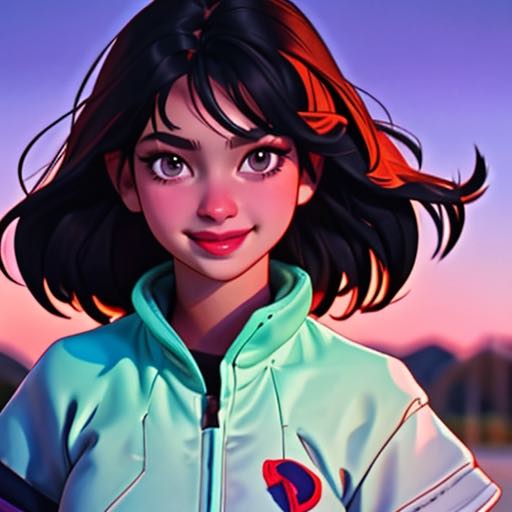} &
            \includegraphics[width=\linewidth]{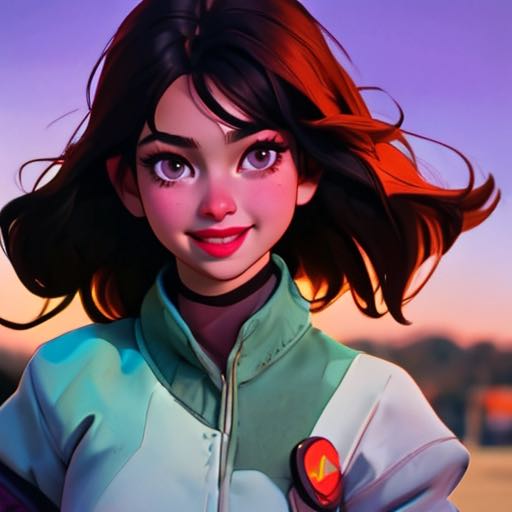} &
            \includegraphics[width=\linewidth]{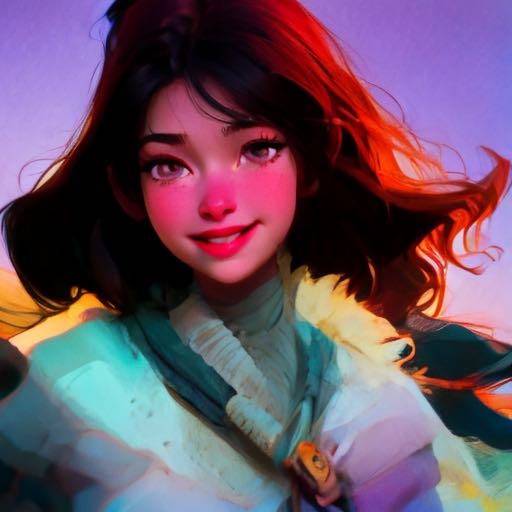} &
            \includegraphics[width=\linewidth]{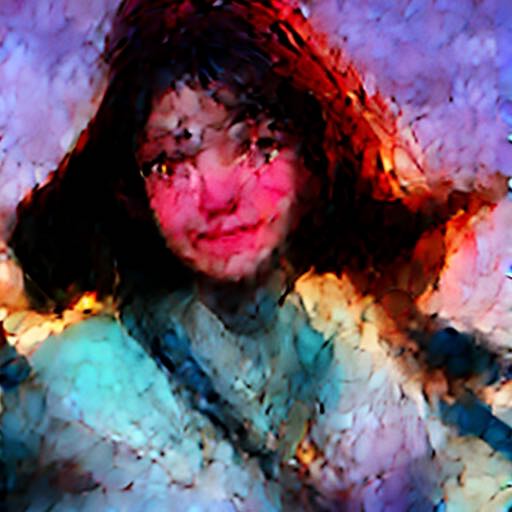} \\
        \end{tabularx}
        \caption{ToonYou Beta 6\cite{toonyou}: A girl smiling. (Our method matches the original tone and style better.)}
    \end{subfigure}

    \vspace{10pt}

    \begin{subfigure}[b]{\textwidth}
        \centering
        \setlength\tabcolsep{0pt}
        \renewcommand{\arraystretch}{0}
        \begin{tabularx}{\textwidth}{@{}X@{\hskip 4pt}XXXX@{\hskip 4pt}XXXX@{}}
            \includegraphics[width=\linewidth]{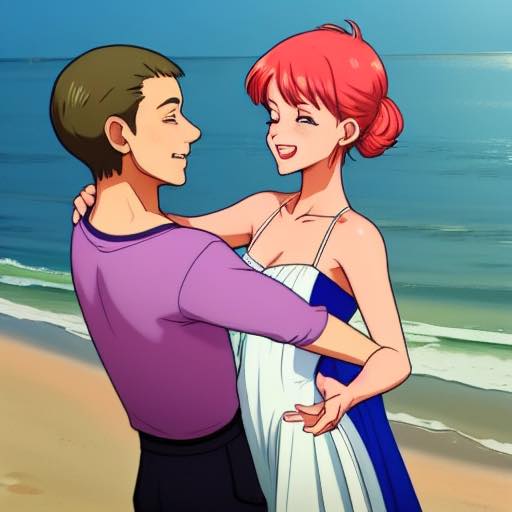} &
            \includegraphics[width=\linewidth]{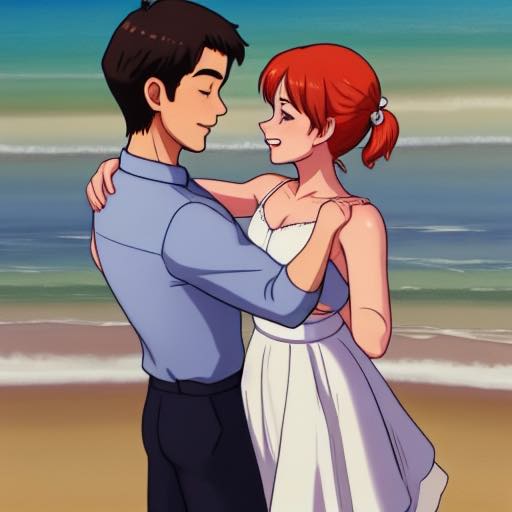} &
            \includegraphics[width=\linewidth]{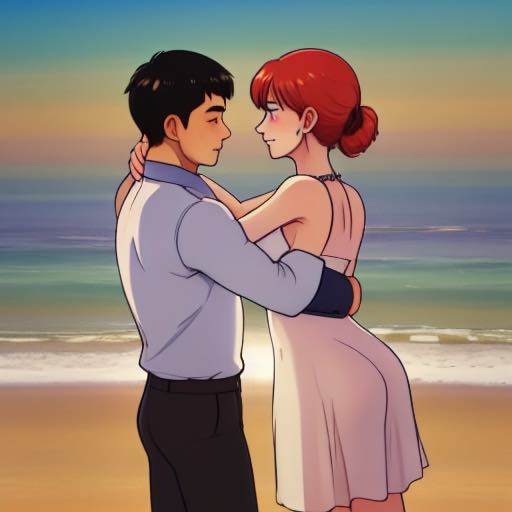} &
            \includegraphics[width=\linewidth]{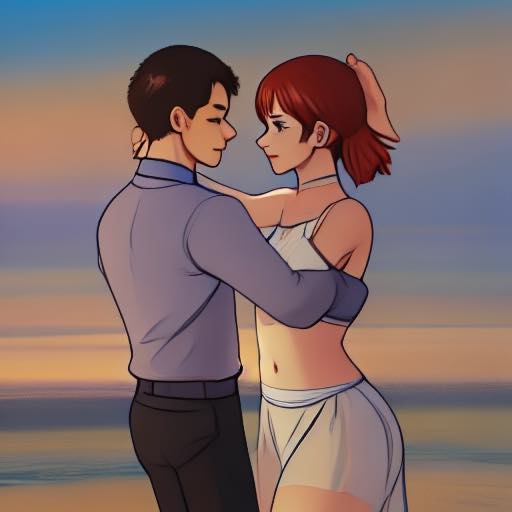} &
            \includegraphics[width=\linewidth]{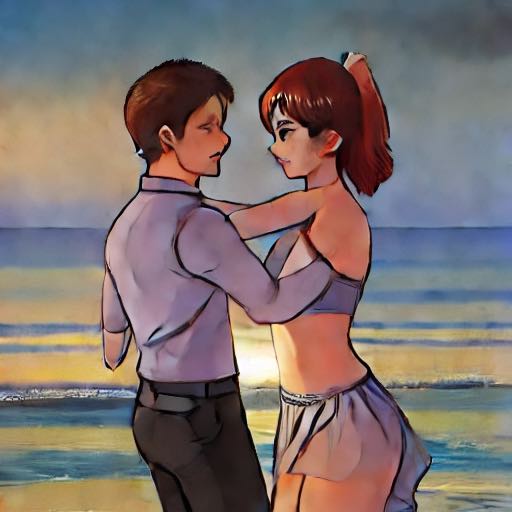} &
            \includegraphics[width=\linewidth]{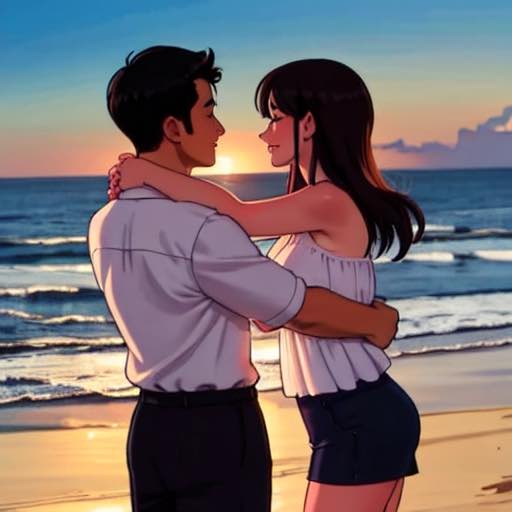} &
            \includegraphics[width=\linewidth]{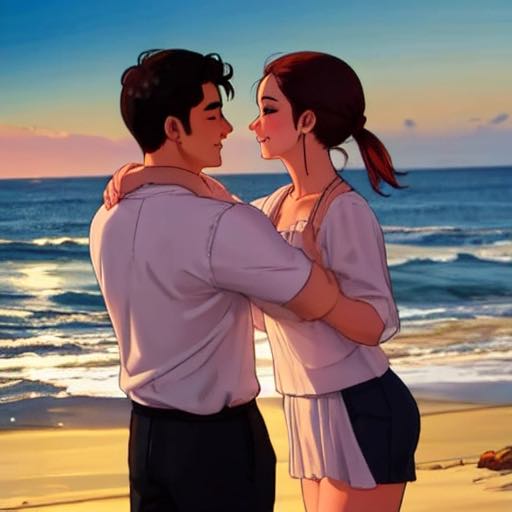} &
            \includegraphics[width=\linewidth]{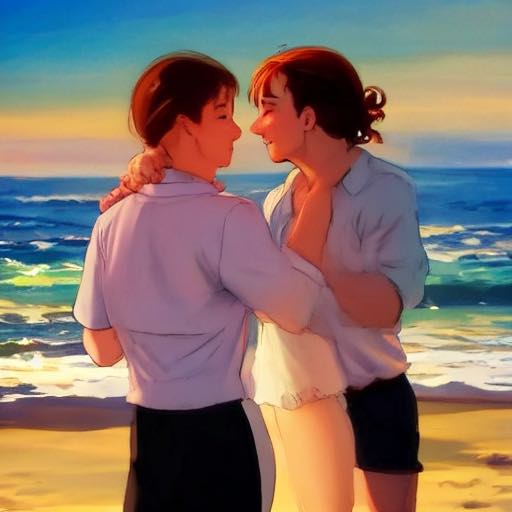} &
            \includegraphics[width=\linewidth]{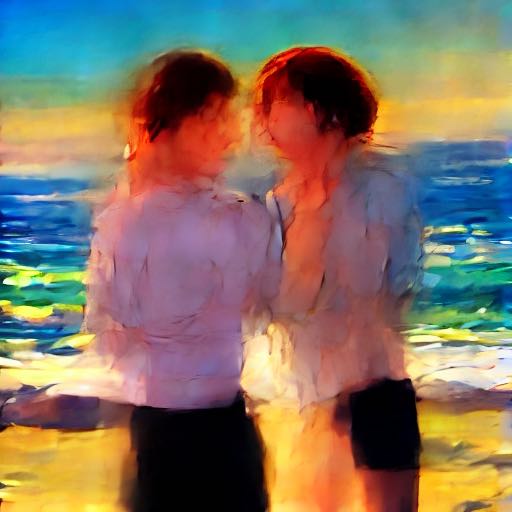}\\
            \includegraphics[width=\linewidth]{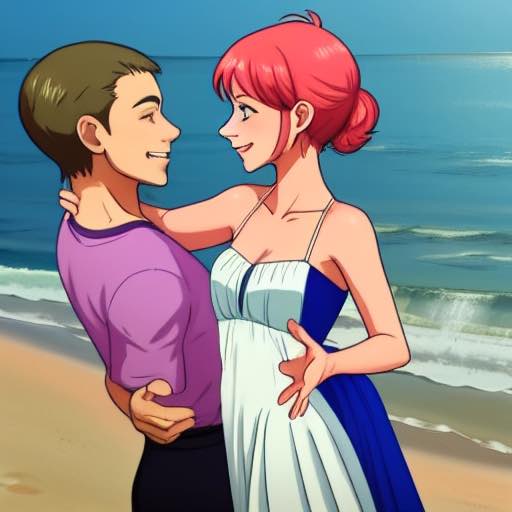} &
            \includegraphics[width=\linewidth]{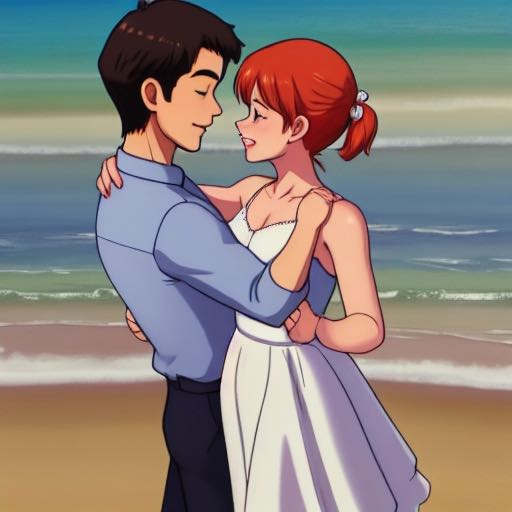} &
            \includegraphics[width=\linewidth]{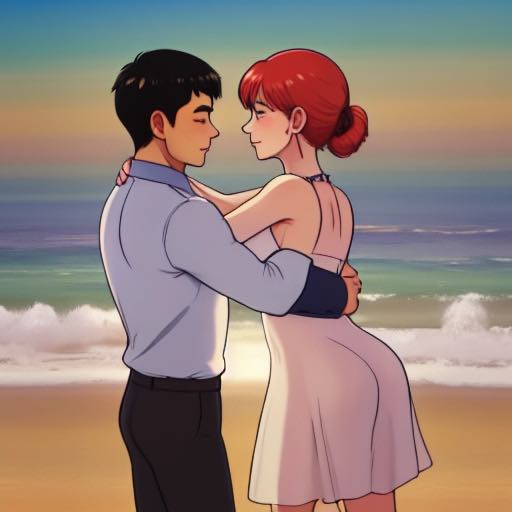} &
            \includegraphics[width=\linewidth]{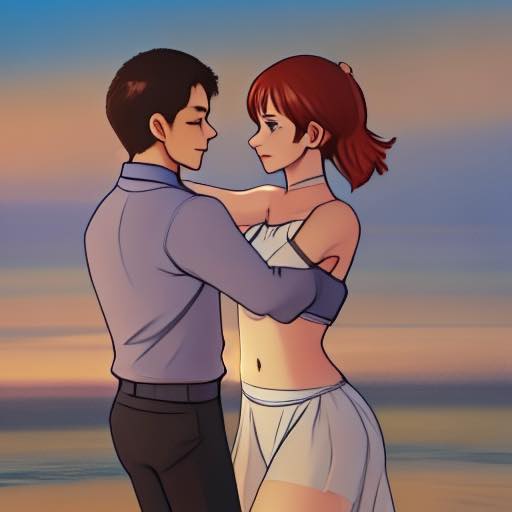} &
            \includegraphics[width=\linewidth]{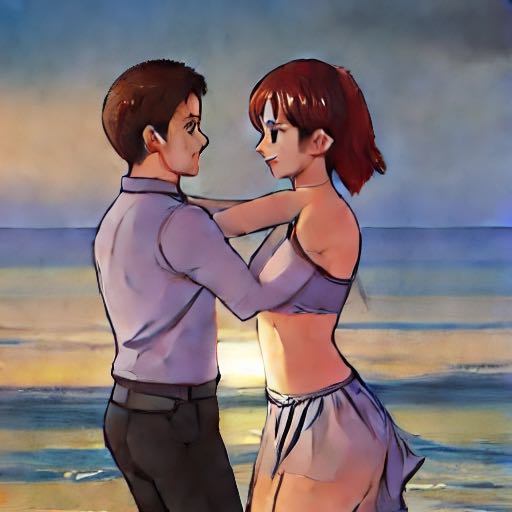} &
            \includegraphics[width=\linewidth]{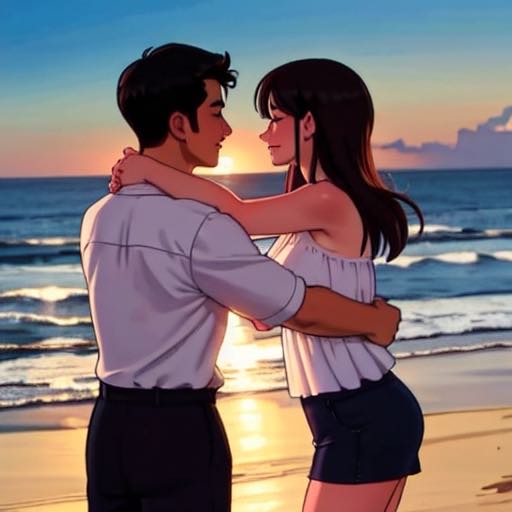} &
            \includegraphics[width=\linewidth]{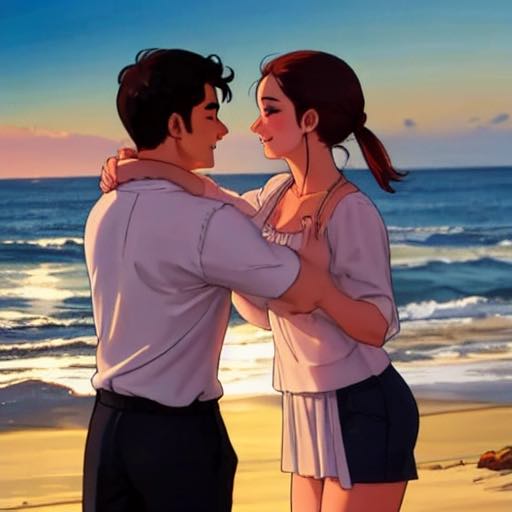} &
            \includegraphics[width=\linewidth]{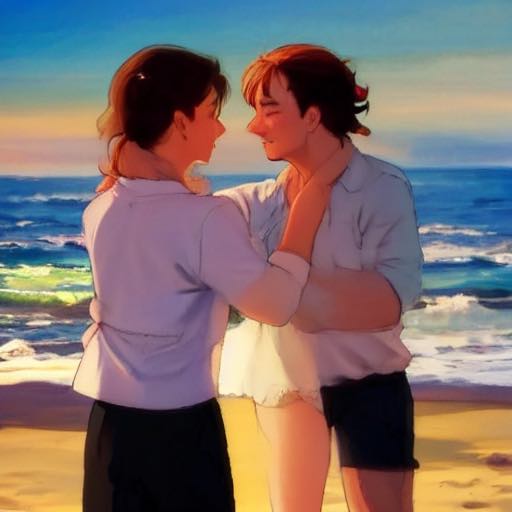} &
            \includegraphics[width=\linewidth]{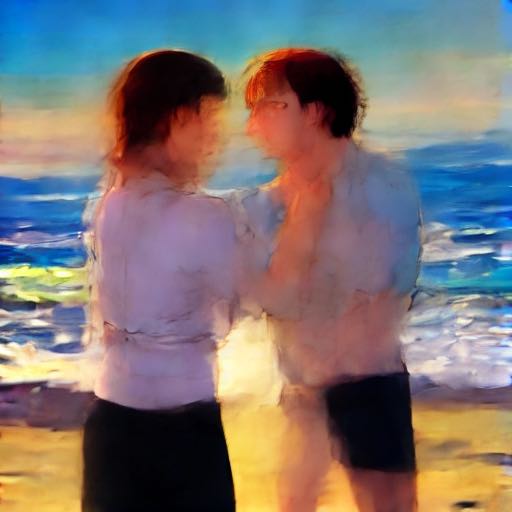}\\
            \includegraphics[width=\linewidth]{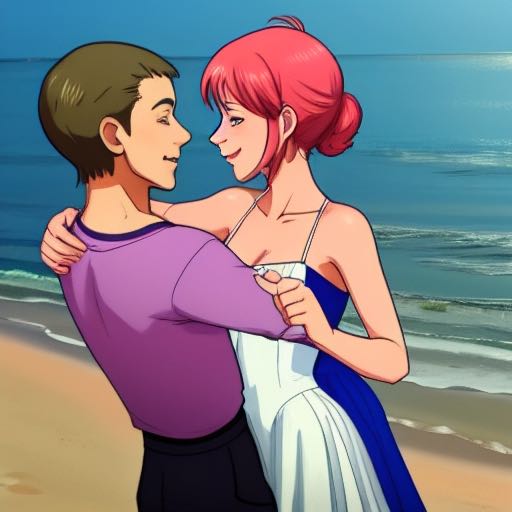} &
            \includegraphics[width=\linewidth]{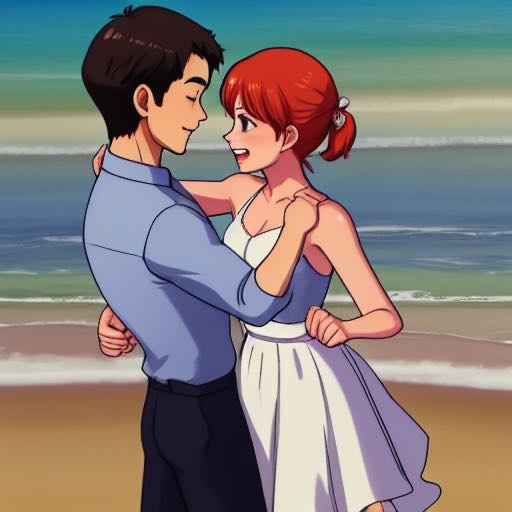} &
            \includegraphics[width=\linewidth]{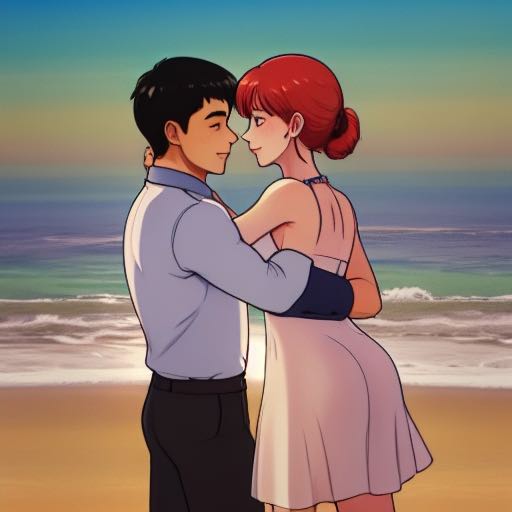} &
            \includegraphics[width=\linewidth]{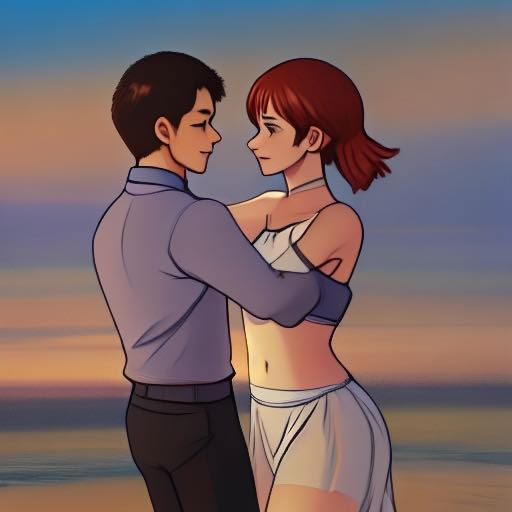} &
            \includegraphics[width=\linewidth]{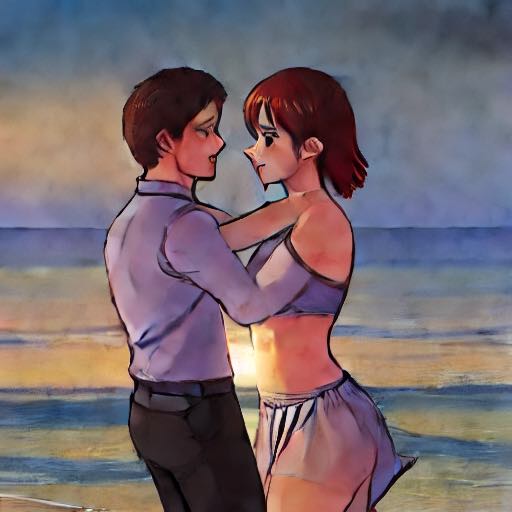} &
            \includegraphics[width=\linewidth]{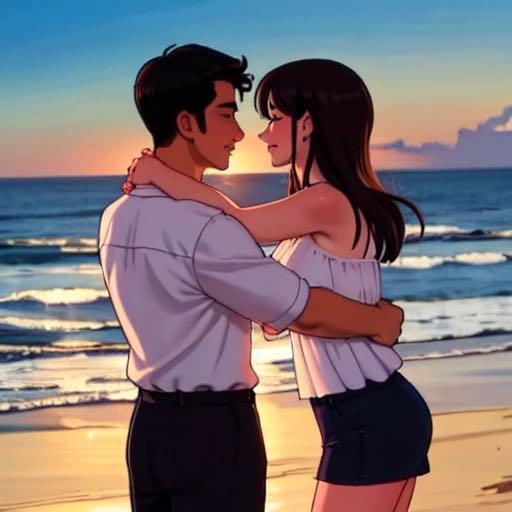} &
            \includegraphics[width=\linewidth]{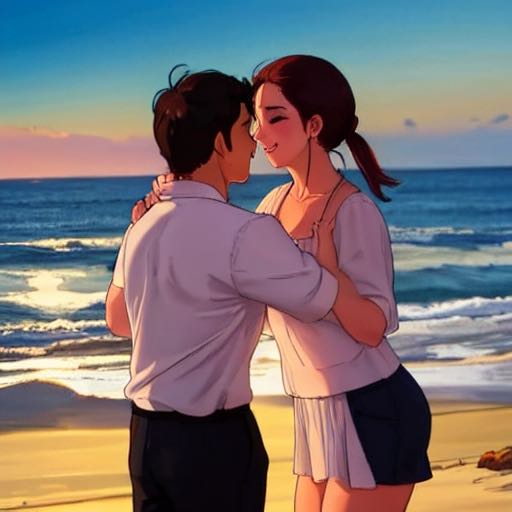} &
            \includegraphics[width=\linewidth]{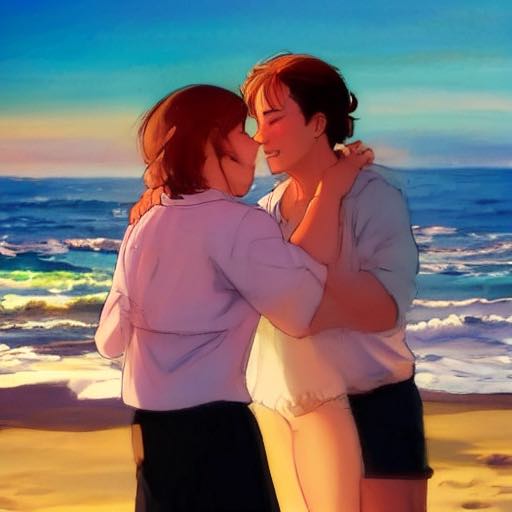} &
            \includegraphics[width=\linewidth]{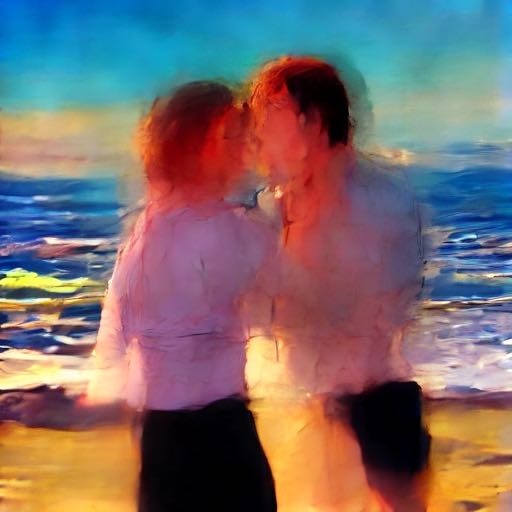}\\
        \end{tabularx}
        \caption{Mistoon Anime \cite{mistoon}: A couple dancing at the beach. (On an unseen base model, our method matches the original style, clothing, and hair color better.)}
        \label{fig:qualitative-mistoon}
    \end{subfigure}
    
    \caption{Qualitative Comparison. Continuing from the last page, we show an anime-style generation comparison on this page. We also try to apply our model on an unseen base model: Mistoon Anime \cite{mistoon} in \cref{fig:qualitative-mistoon}. Though there is style degradation as the inference step reduces, our model produces more similar results compared to the original in terms of overall anime style, clothing, and hair color of the characters.}
    \label{fig:qualitative}
\end{figure*}

\begin{figure*}
    \small
    \setlength\tabcolsep{4pt}
    \begin{tabularx}{\linewidth}{|X@{\hskip 8pt}|X|X|X@{\hskip 8pt}|X|X|X|}
        \normalsize{\textbf{Original} \cite{guo2023animatediff}} & \multicolumn{3}{l|}{\normalsize{\textbf{Cross-Model Distillation}}} & \multicolumn{3}{l|}{\normalsize{\textbf{Single-Model Distillation}}} \\
        32 Steps & 8 Steps & 4 Steps & 2 Steps & 8 Steps & 4 Steps & 2 Steps
    \end{tabularx}

    \setlength\tabcolsep{0pt}
    \captionsetup{justification=raggedright,singlelinecheck=false}
    \begin{subfigure}[b]{\textwidth}
        \begin{tabularx}{\linewidth}{X@{\hskip 4pt}XXX@{\hskip 4pt}XXX}
            \includegraphics[width=\linewidth]{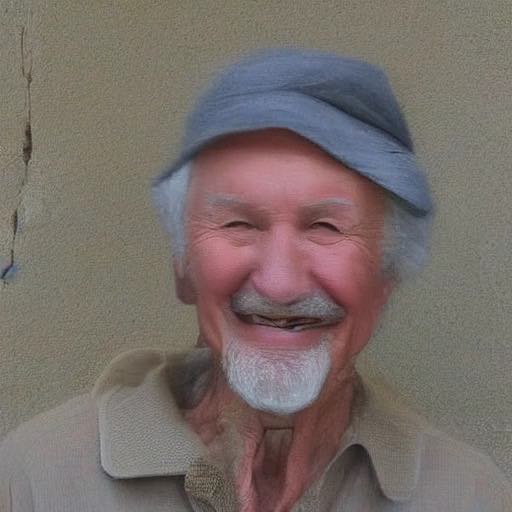} &
            \includegraphics[width=\linewidth]{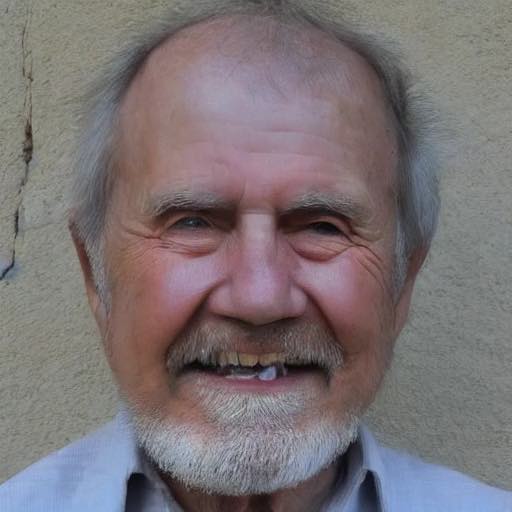} &
            \includegraphics[width=\linewidth]{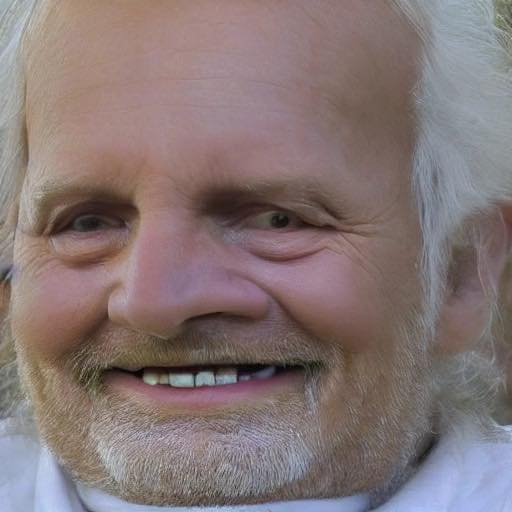} &
            \includegraphics[width=\linewidth]{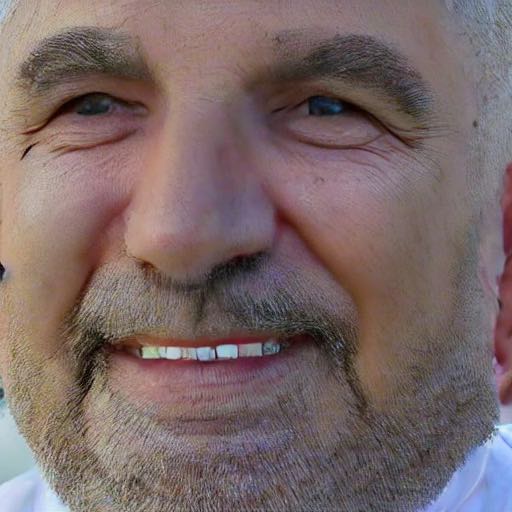} &
            \includegraphics[width=\linewidth]{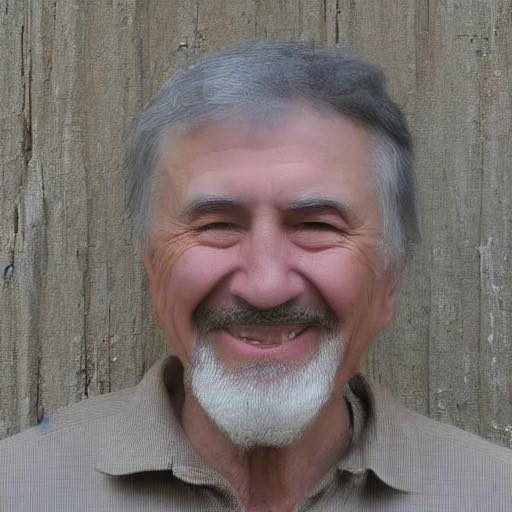} &
            \includegraphics[width=\linewidth]{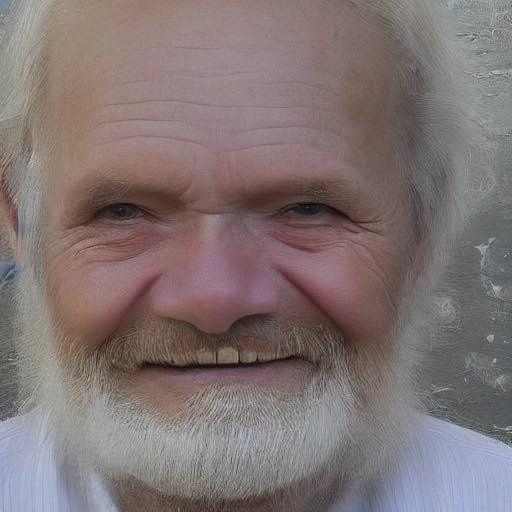} &
            \includegraphics[width=\linewidth]{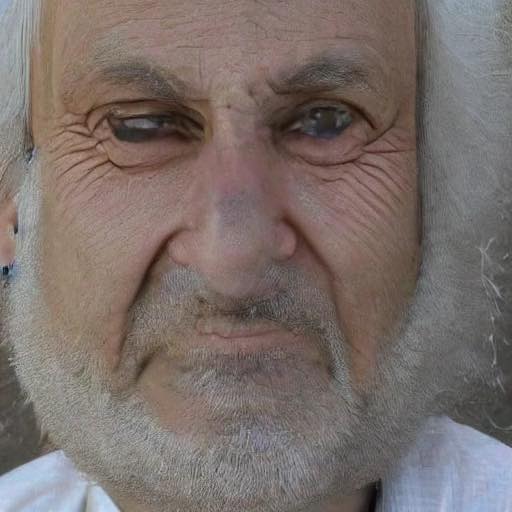} \\
        \end{tabularx}
        \vspace{-6pt}
        \caption{Stable Diffusion v1.5 \cite{rombach2022highresolution}: An old man smiling.}
    \end{subfigure}

    \begin{subfigure}[b]{\textwidth}
        \begin{tabularx}{\linewidth}{X@{\hskip 4pt}XXX@{\hskip 4pt}XXX}
            \includegraphics[width=\linewidth]{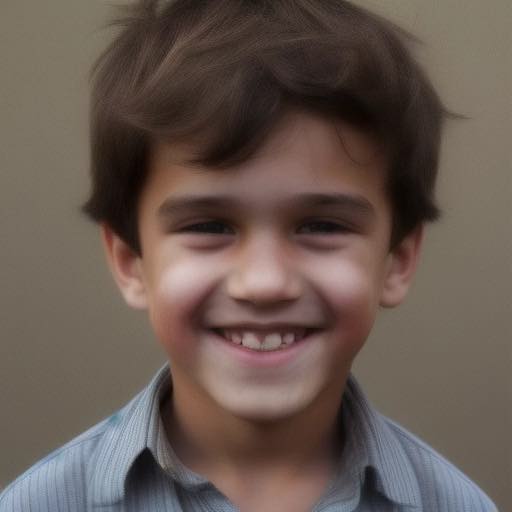} &
            \includegraphics[width=\linewidth]{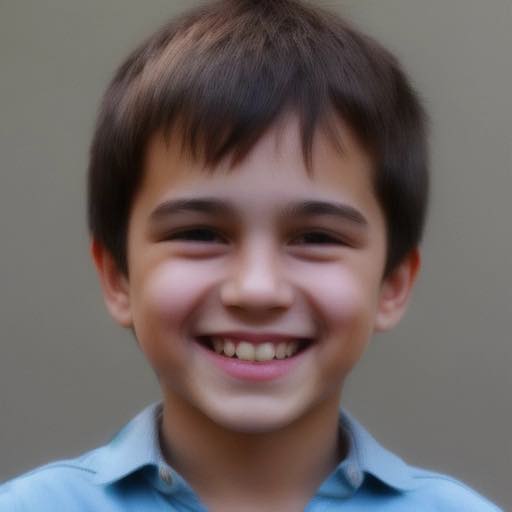} &
            \includegraphics[width=\linewidth]{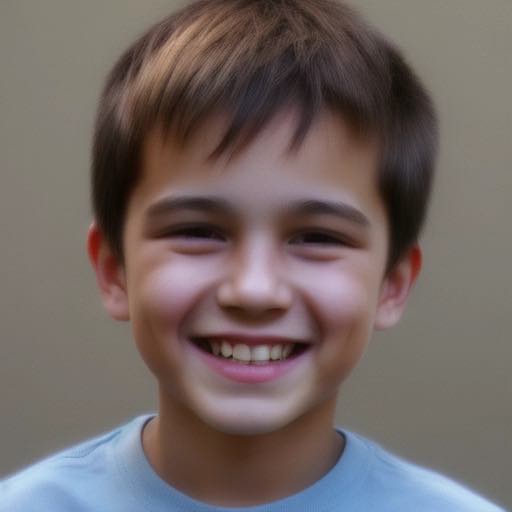} &
            \includegraphics[width=\linewidth]{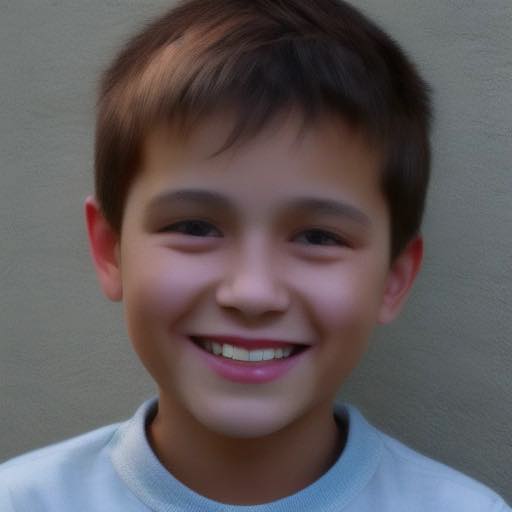} &
            \includegraphics[width=\linewidth]{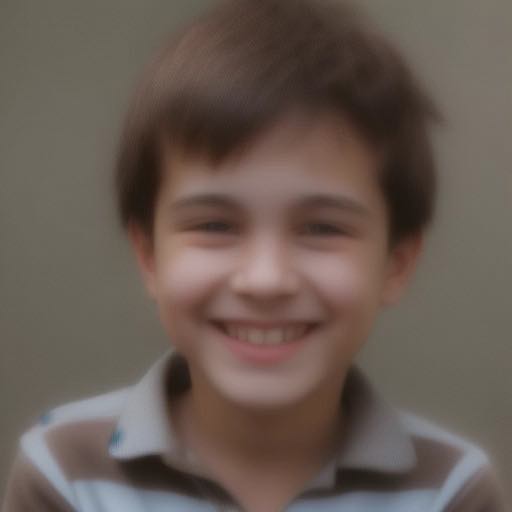} &
            \includegraphics[width=\linewidth]{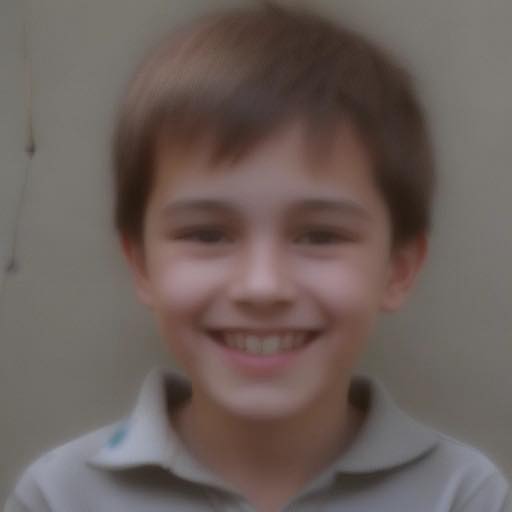} &
            \includegraphics[width=\linewidth]{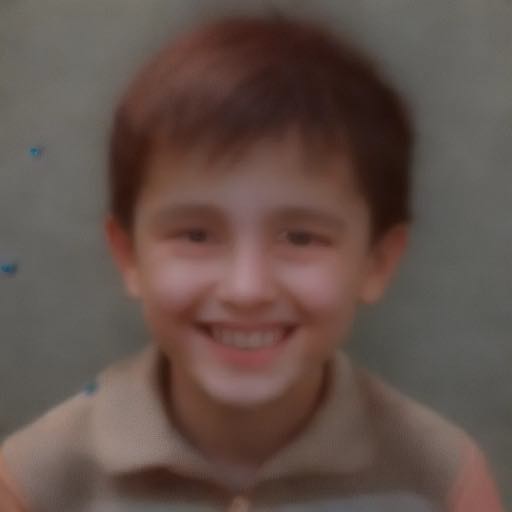} \\
        \end{tabularx}
        \vspace{-6pt}
        \caption{RealisticVision v5.1 \cite{realisticvision}: A boy smiling.}
    \end{subfigure}

    \begin{subfigure}[b]{\textwidth}
        \begin{tabularx}{\linewidth}{X@{\hskip 4pt}XXX@{\hskip 4pt}XXX}
            \includegraphics[width=\linewidth]{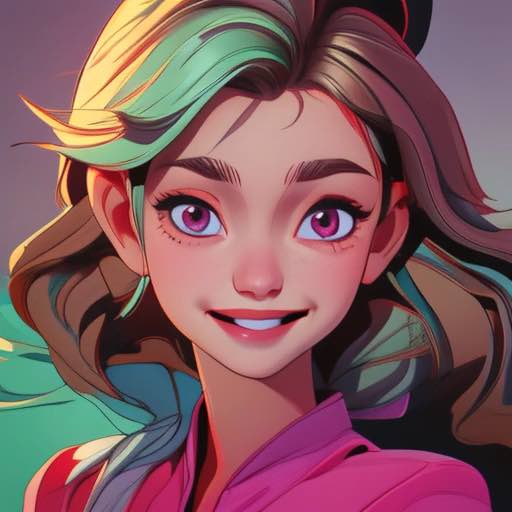} &
            \includegraphics[width=\linewidth]{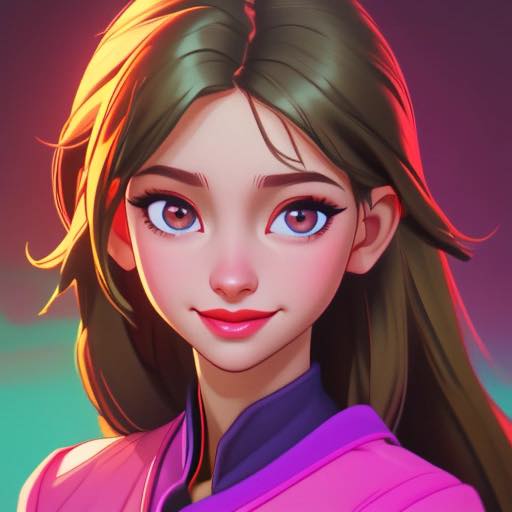} &
            \includegraphics[width=\linewidth]{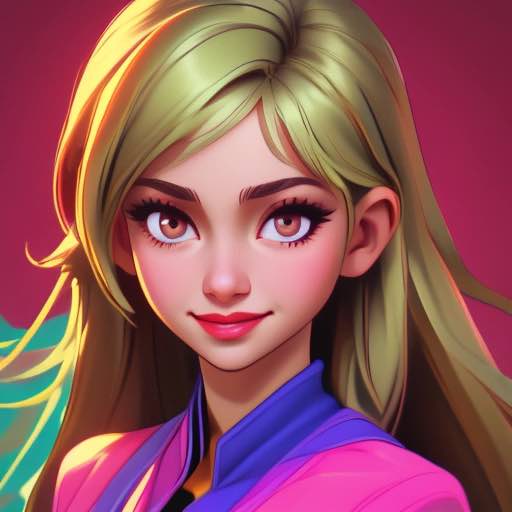} &
            \includegraphics[width=\linewidth]{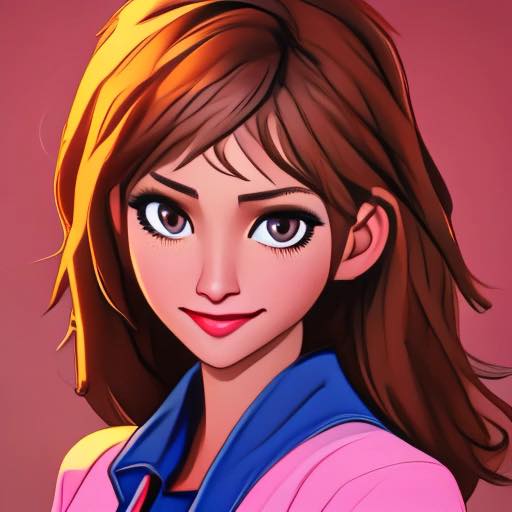} &
            \includegraphics[width=\linewidth]{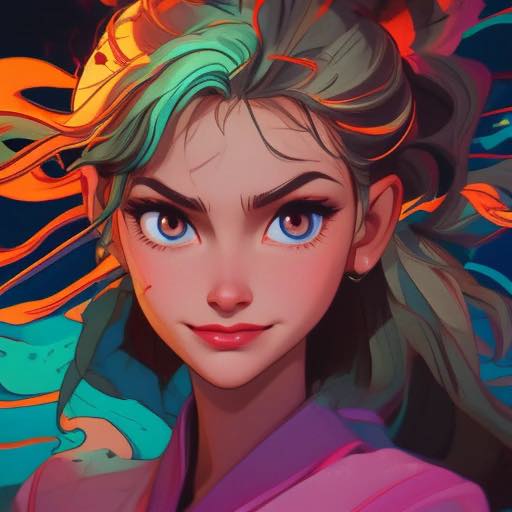} &
            \includegraphics[width=\linewidth]{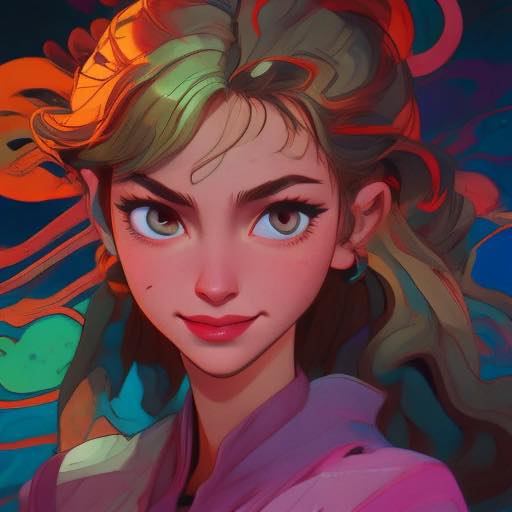} &
            \includegraphics[width=\linewidth]{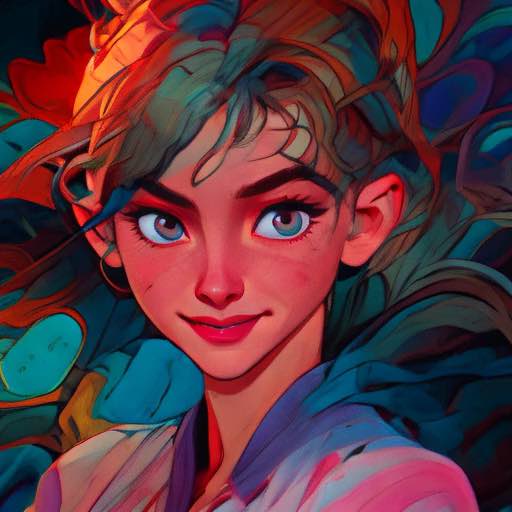} \\
        \end{tabularx}
        \vspace{-6pt}
        \caption{ToonYou Beta 6 \cite{toonyou}: A girl smiling.}
    \end{subfigure}

    \caption{Comparison between cross-model and single-model distillation. Single-model distillation is trained only on SD v1.5 \cite{rombach2022highresolution} base model with the WebVid-10M \cite{Bain2021FrozenIT} dataset. Single-model distillation fails to retain quality on other base models. We show the first frame of the generated video clips.}
    \label{fig:cross-ablation}
\end{figure*}

\vspace{10pt}

\begin{figure*}
    \footnotesize
    \setlength\tabcolsep{2pt}
    \begin{tabularx}{\linewidth}{|X|X|X|X|X|X|X|X|}
        \makecell{AbsoluteReality\\v1.8.1 \cite{absolutereality}} & \makecell{DreamShaper\\v8 \cite{dreamshaper}} & \makecell{DynaVision\\v2 \cite{dynavision}} & \makecell{Exquisite Details\\Art \cite{exquisite}} & \makecell{MajicMix\\Realistic v7 \cite{majicmix_realistic}} & \makecell{MajicMix\\Reverie v1 \cite{majicmix_reverie}} & \makecell{RCNZ Cartoon\\v2 \cite{rcnz}} & \makecell{ReV Animated\\v1.2.2 \cite{rev}}
    \end{tabularx}

    \captionsetup{justification=raggedright,singlelinecheck=false}
    \setlength\tabcolsep{0pt}
    \begin{subfigure}[b]{\textwidth}
        \begin{tabularx}{\linewidth}{XXXXXXXX}
            \includegraphics[width=\linewidth]{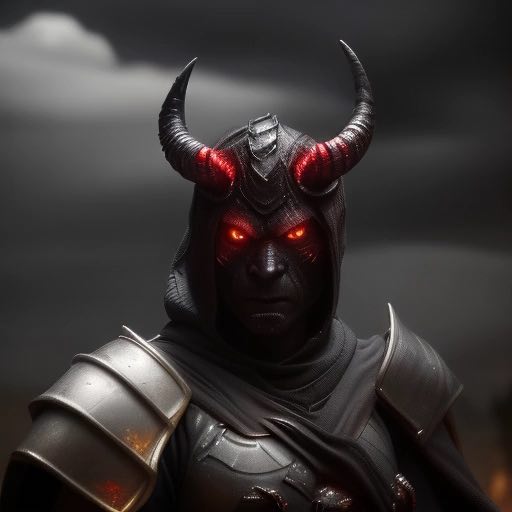} &
            \includegraphics[width=\linewidth]{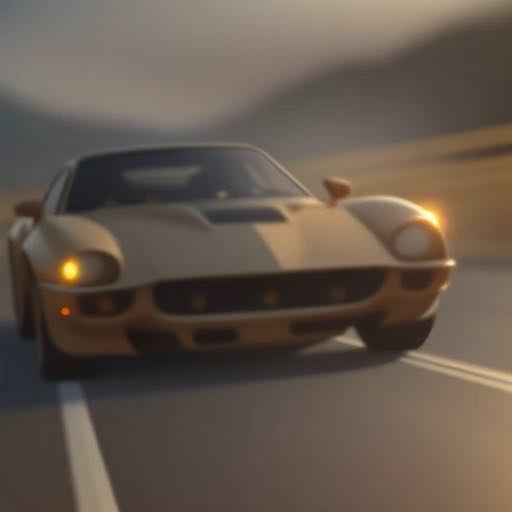} &
            \includegraphics[width=\linewidth]{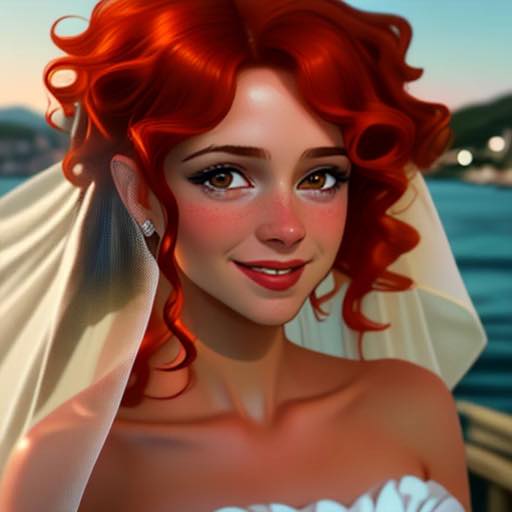} &
            \includegraphics[width=\linewidth]{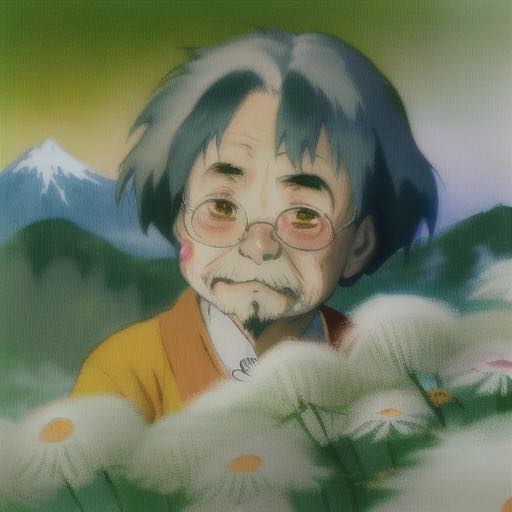} &
            \includegraphics[width=\linewidth]{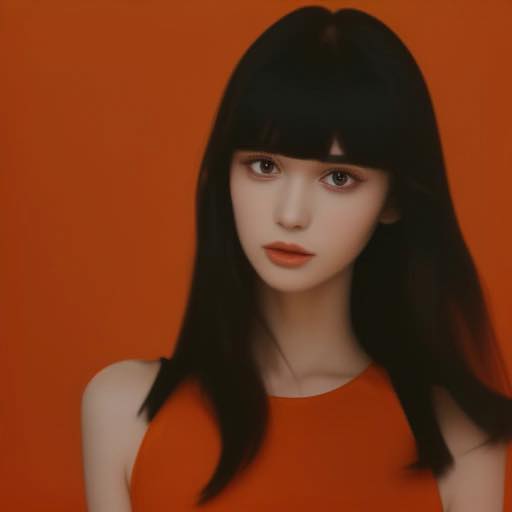} &
            \includegraphics[width=\linewidth]{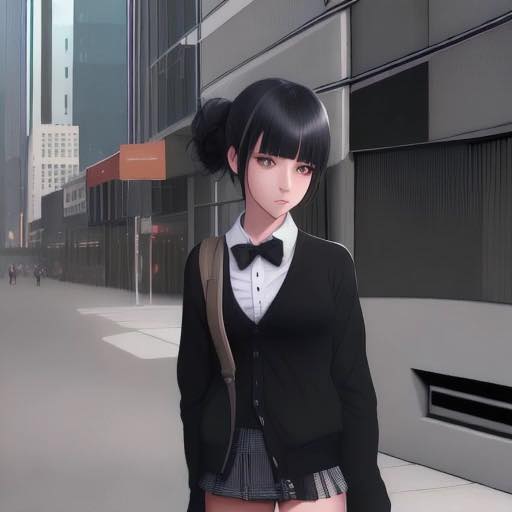} &
            \includegraphics[width=\linewidth]{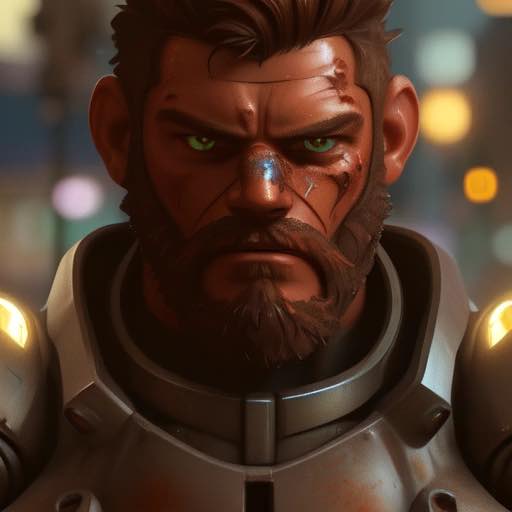} &
            \includegraphics[width=\linewidth]{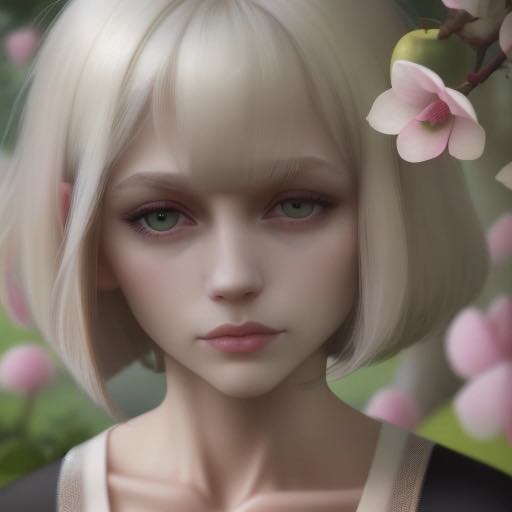}\\
        \end{tabularx}
        \vspace{-6pt}
        \caption{AnimateDiff \cite{guo2023animatediff} using 32 steps with Euler sampler.}
    \end{subfigure}

    \begin{subfigure}[b]{\textwidth}
        \begin{tabularx}{\linewidth}{XXXXXXXX}
            \includegraphics[width=\linewidth]{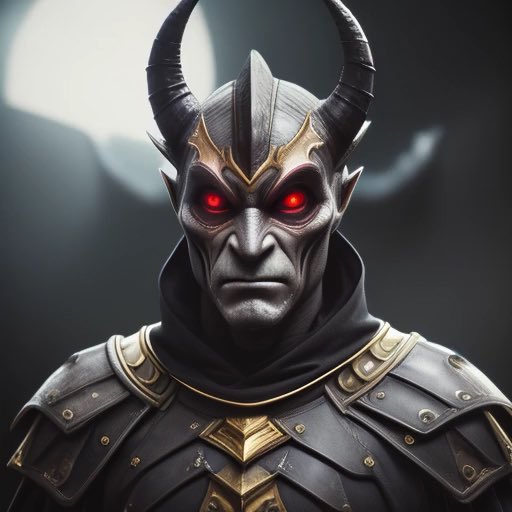} &
            \includegraphics[width=\linewidth]{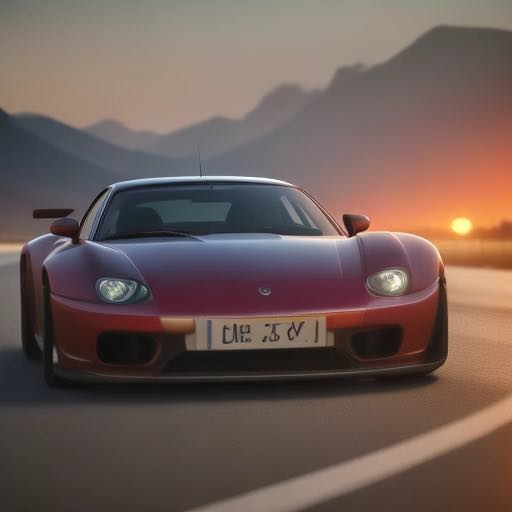} &
            \includegraphics[width=\linewidth]{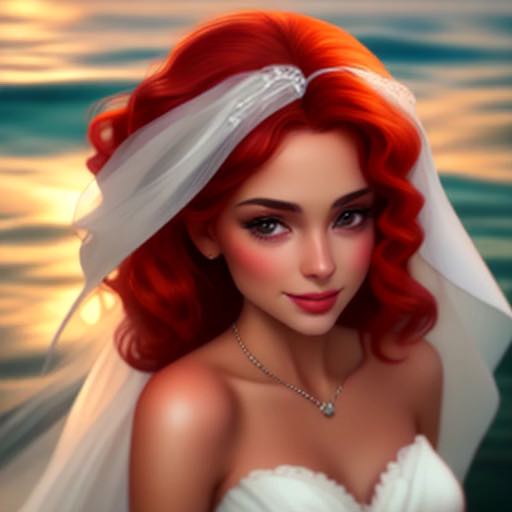} &
            \includegraphics[width=\linewidth]{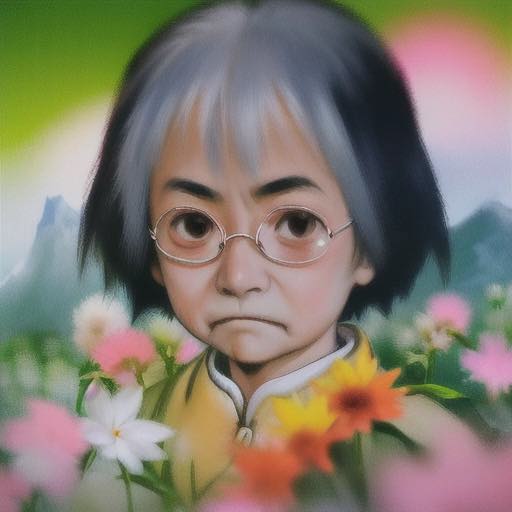} &
            \includegraphics[width=\linewidth]{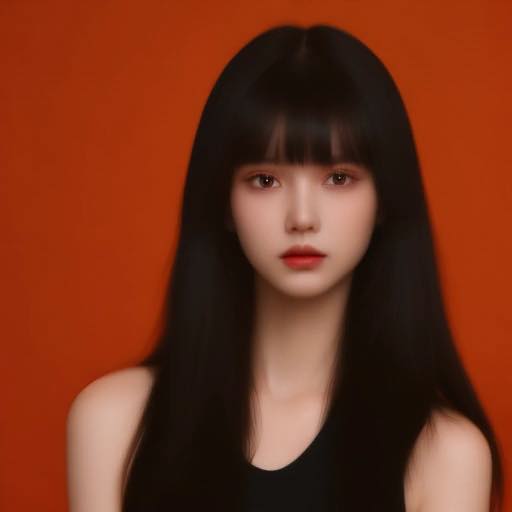} &
            \includegraphics[width=\linewidth]{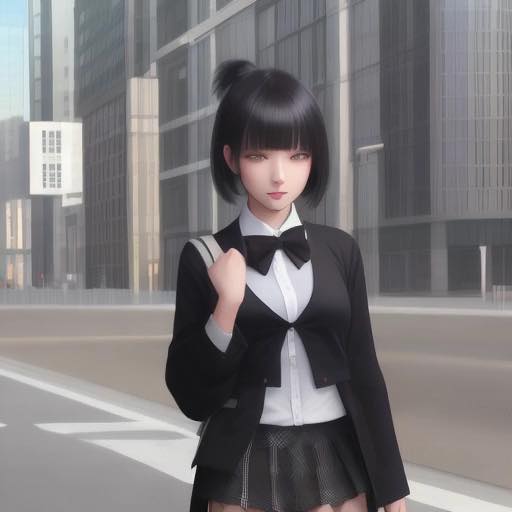} &
            \includegraphics[width=\linewidth]{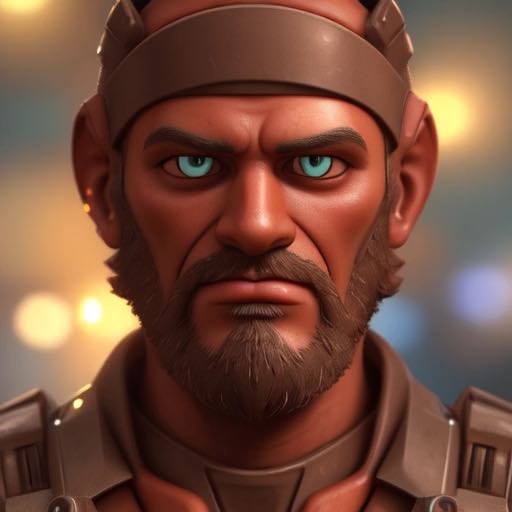} &
            \includegraphics[width=\linewidth]{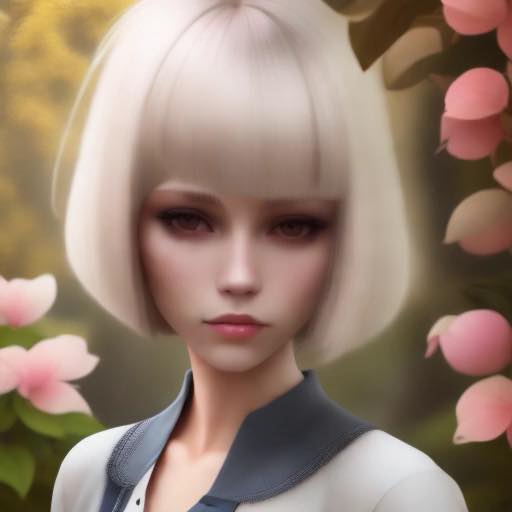}\\
        \end{tabularx}
        \vspace{-6pt}
        \caption{Our method using 4 steps.}
    \end{subfigure}

    \begin{subfigure}[b]{\textwidth}
        \begin{tabularx}{\linewidth}{XXXXXXXX}
            \includegraphics[width=\linewidth]{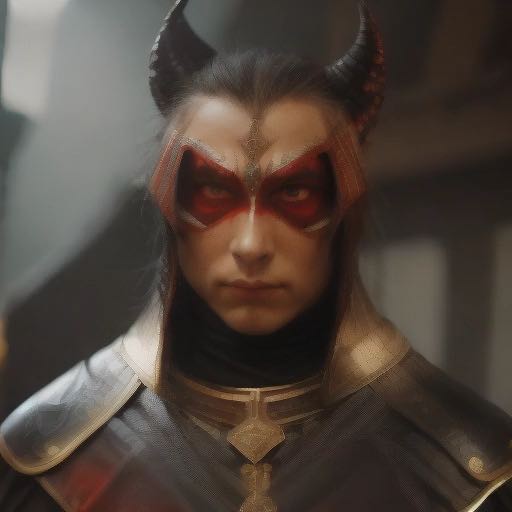} &
            \includegraphics[width=\linewidth]{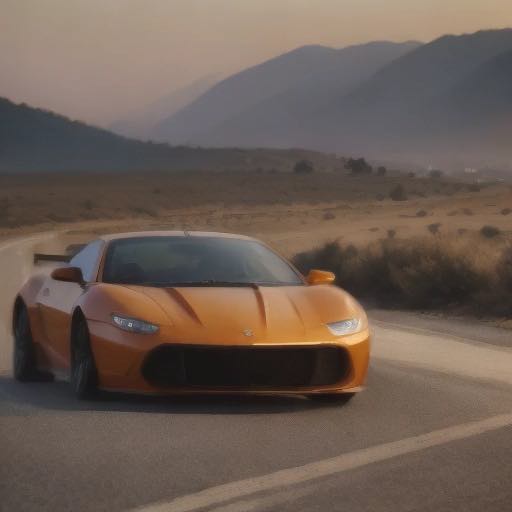} &
            \includegraphics[width=\linewidth]{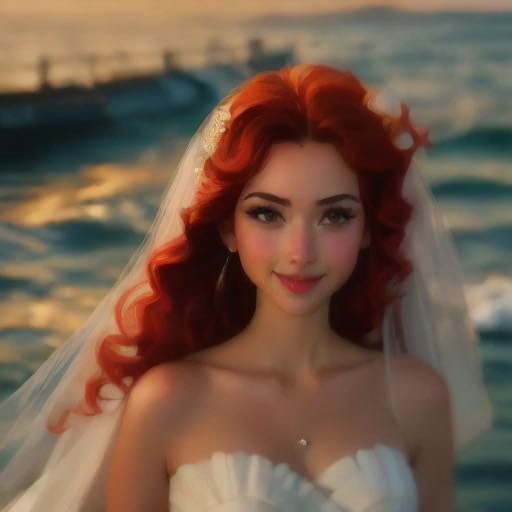} &
            \includegraphics[width=\linewidth]{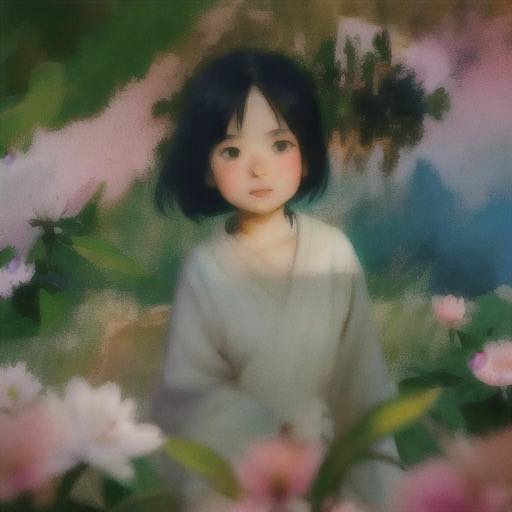} &
            \includegraphics[width=\linewidth]{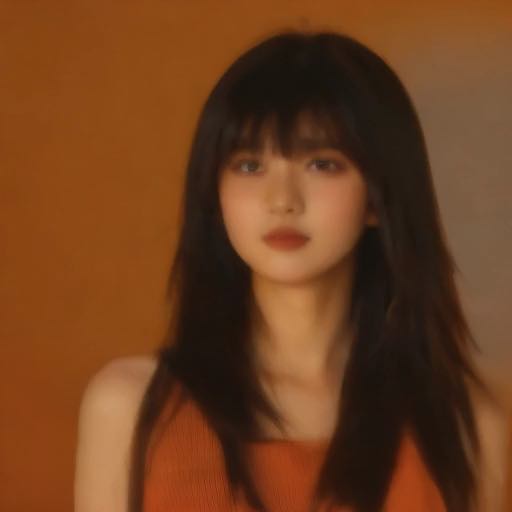} &
            \includegraphics[width=\linewidth]{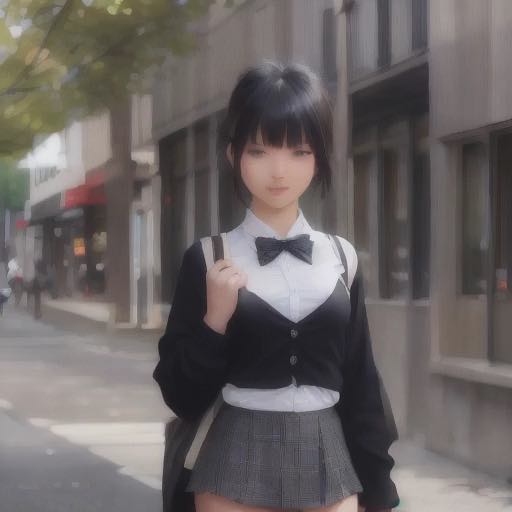} &
            \includegraphics[width=\linewidth]{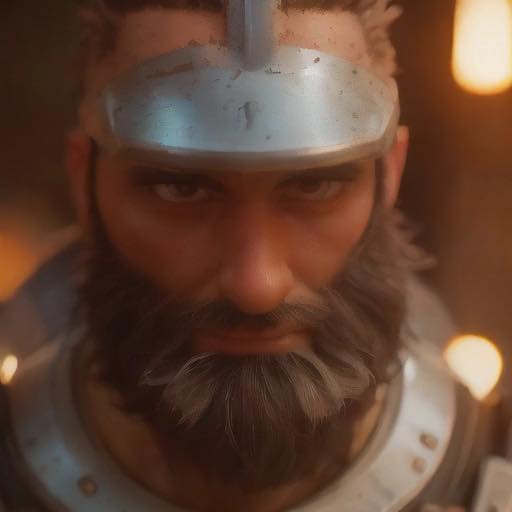} &
            \includegraphics[width=\linewidth]{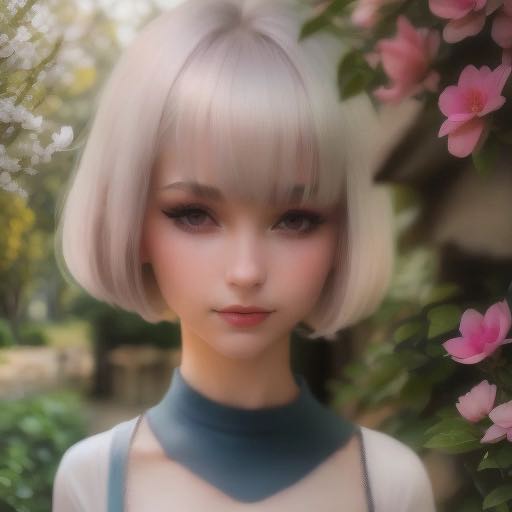}\\
        \end{tabularx}
        \vspace{-6pt}
        \caption{AnimateLCM \cite{wang2024animatelcm} using 4 steps.}
    \end{subfigure}

    \caption{Distillation results on unseen base models. All the image base models here are unseen during the distillation of our model and the AnimateLCM model \cite{wang2024animatelcm}. Our results are better in detail and are closer to the original styles. We use different prompts that best match the image base models' specialty, but the same prompt and seed are used across model comparisons. We show the first frame of the generated video clips.}
    \label{fig:unseen}
\end{figure*}

\clearpage

\begin{figure*}
    \centering
    \newcolumntype{Y}{>{\centering\arraybackslash}X}
    \setlength\tabcolsep{0pt}
    \small
    \begin{tabularx}{\linewidth}{|Y|Y|Y|Y|Y|Y|Y|Y|}
        \multicolumn{2}{|c|}{Zoom} & \multicolumn{2}{c|}{Pan} & \multicolumn{2}{c|}{Tilt} & \multicolumn{2}{c|}{Roll} \\
        In & Out & Left & Right & Up & Down & Left & Right
    \end{tabularx}
    \setlength\tabcolsep{0pt}
    \renewcommand{\arraystretch}{0}
    \begin{tabularx}{\linewidth}{XXXXXXXX}
        \includegraphics[width=\linewidth]{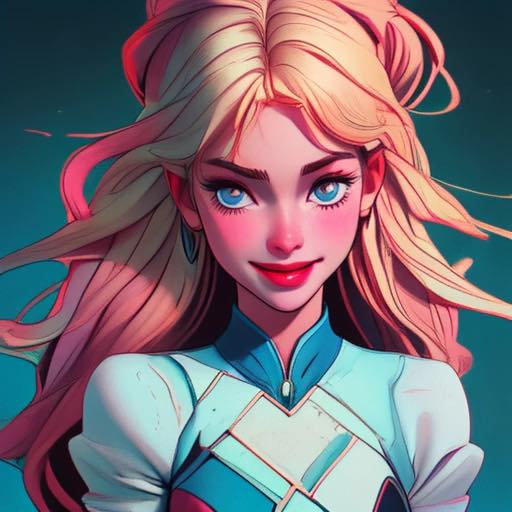} &
        \includegraphics[width=\linewidth]{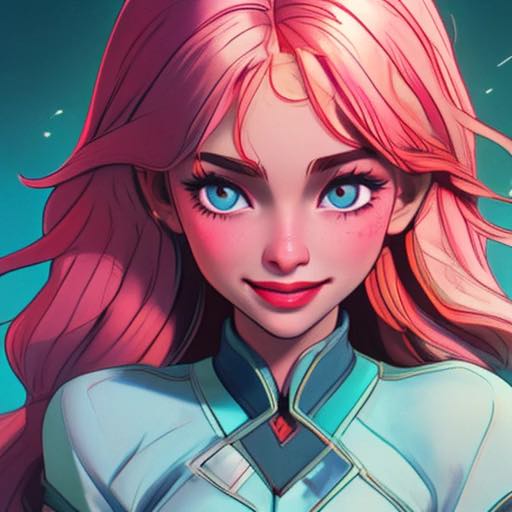} &
        \includegraphics[width=\linewidth]{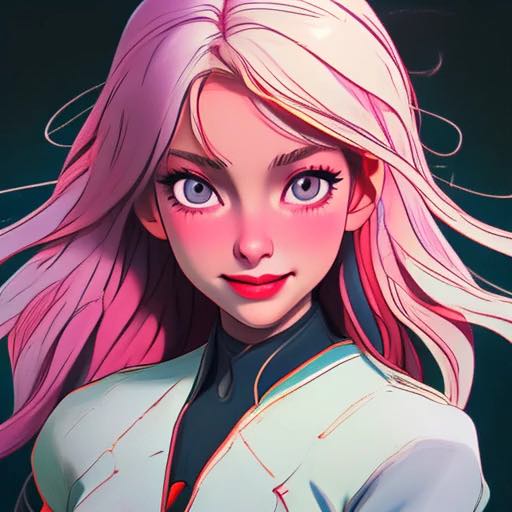} &
        \includegraphics[width=\linewidth]{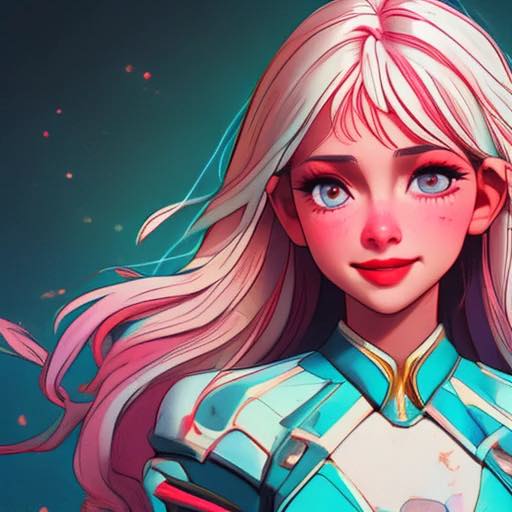} &
        \includegraphics[width=\linewidth]{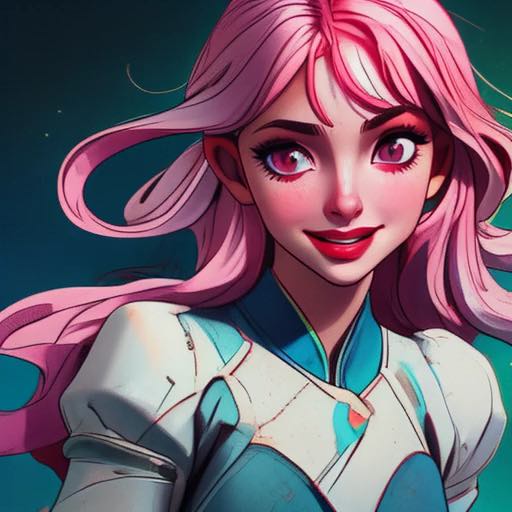} &
        \includegraphics[width=\linewidth]{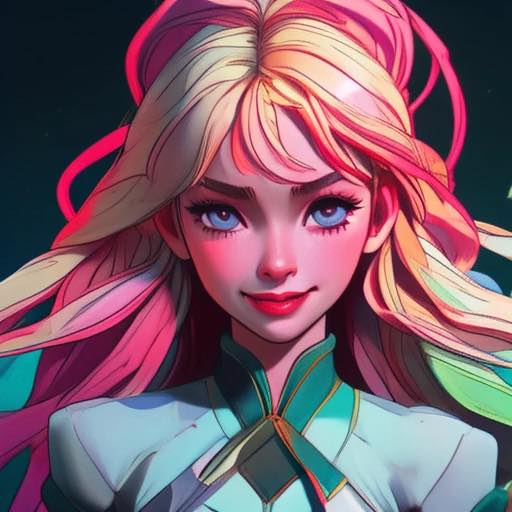} &
        \includegraphics[width=\linewidth]{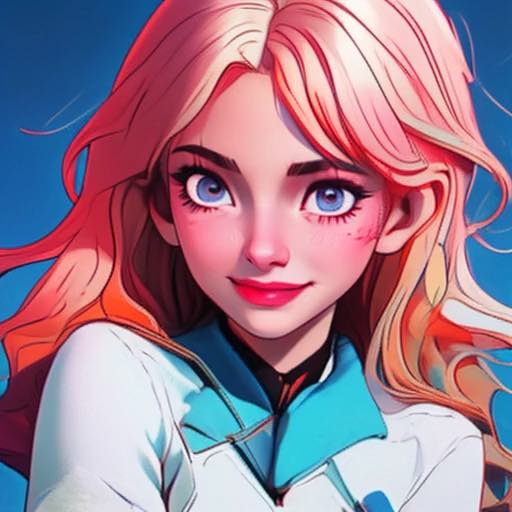} &
        \includegraphics[width=\linewidth]{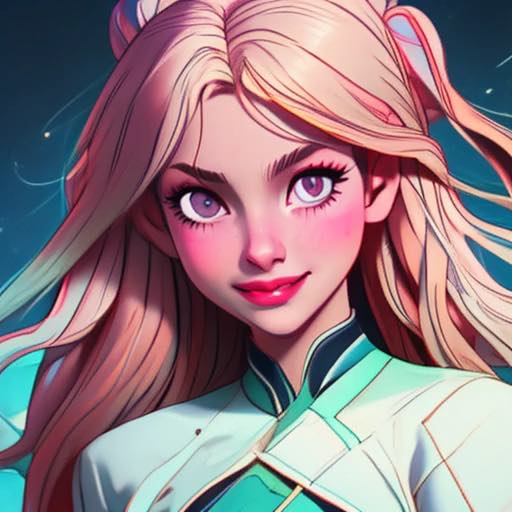} \\
        \includegraphics[width=\linewidth]{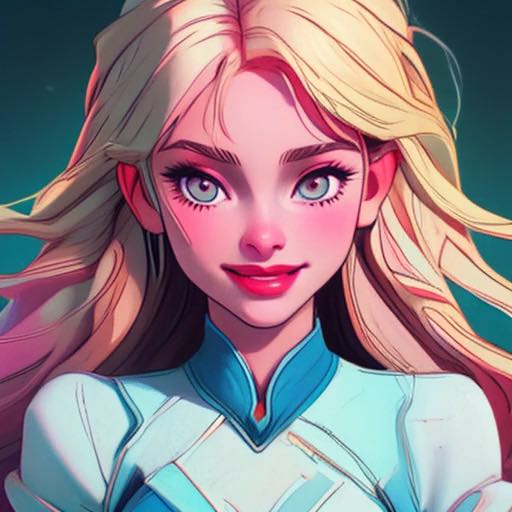} &
        \includegraphics[width=\linewidth]{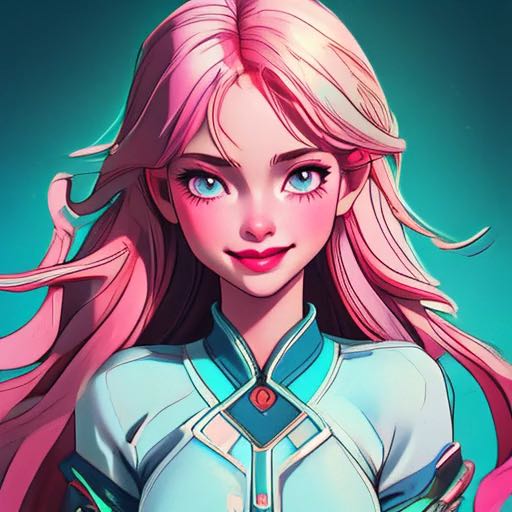} &
        \includegraphics[width=\linewidth]{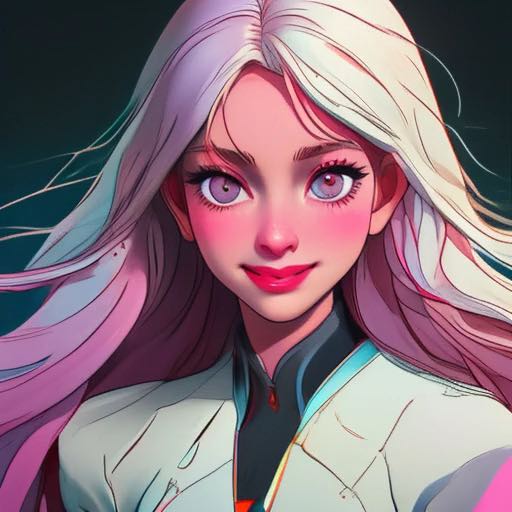} &
        \includegraphics[width=\linewidth]{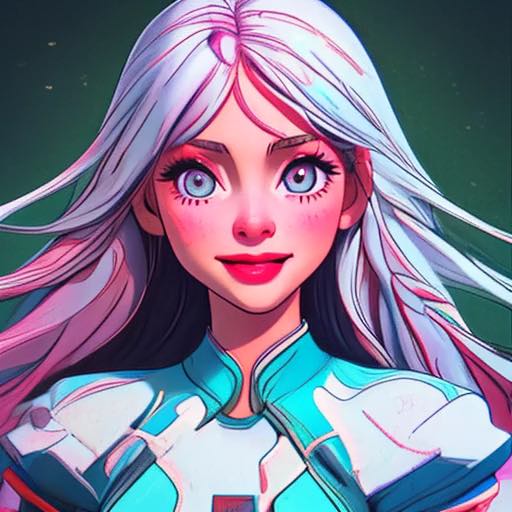} &
        \includegraphics[width=\linewidth]{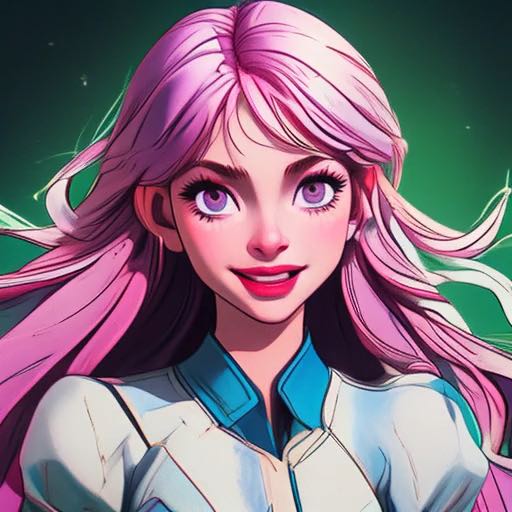} &
        \includegraphics[width=\linewidth]{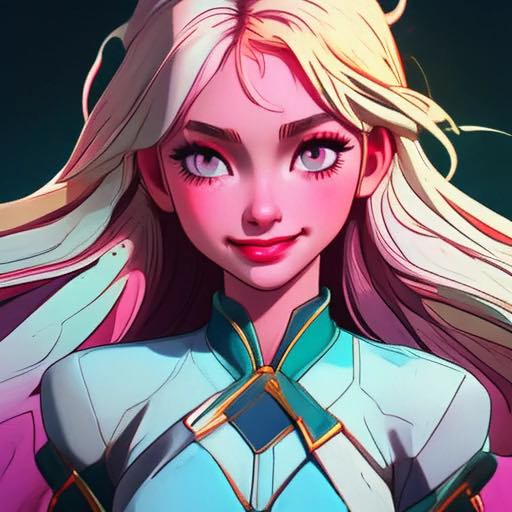} &
        \includegraphics[width=\linewidth]{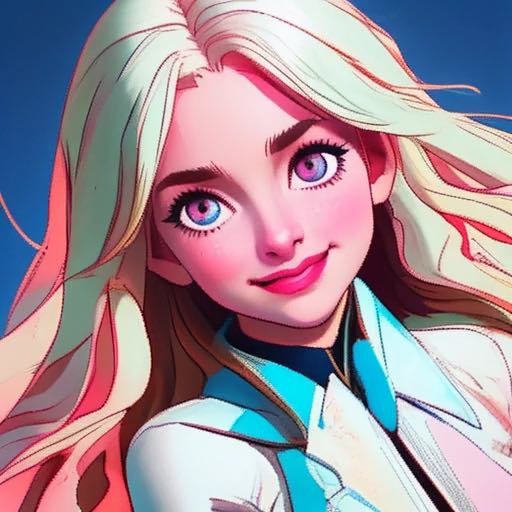} &
        \includegraphics[width=\linewidth]{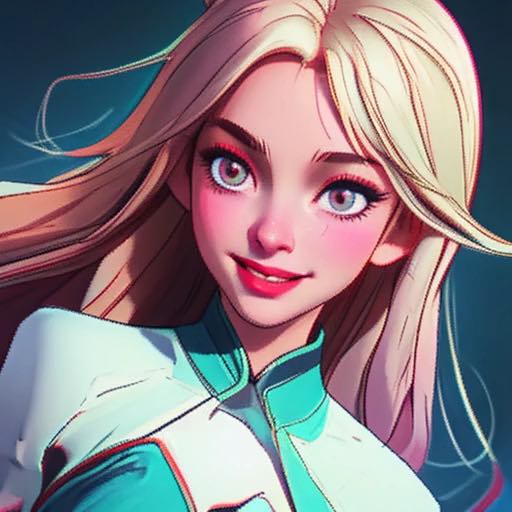}
    \end{tabularx}
    \caption{Our model is compatible with Motion LoRA modules \cite{guo2023animatediff} for fine-grained motion control. Here is our 4-step model on ToonYou \cite{toonyou} with prompt: ``A girl smiling''. The first row is the starting frame and the second row is the final frame.}
    \label{fig:motion-lora}
\end{figure*}

\begin{figure*}
    \centering
    \includegraphics[width=\linewidth]{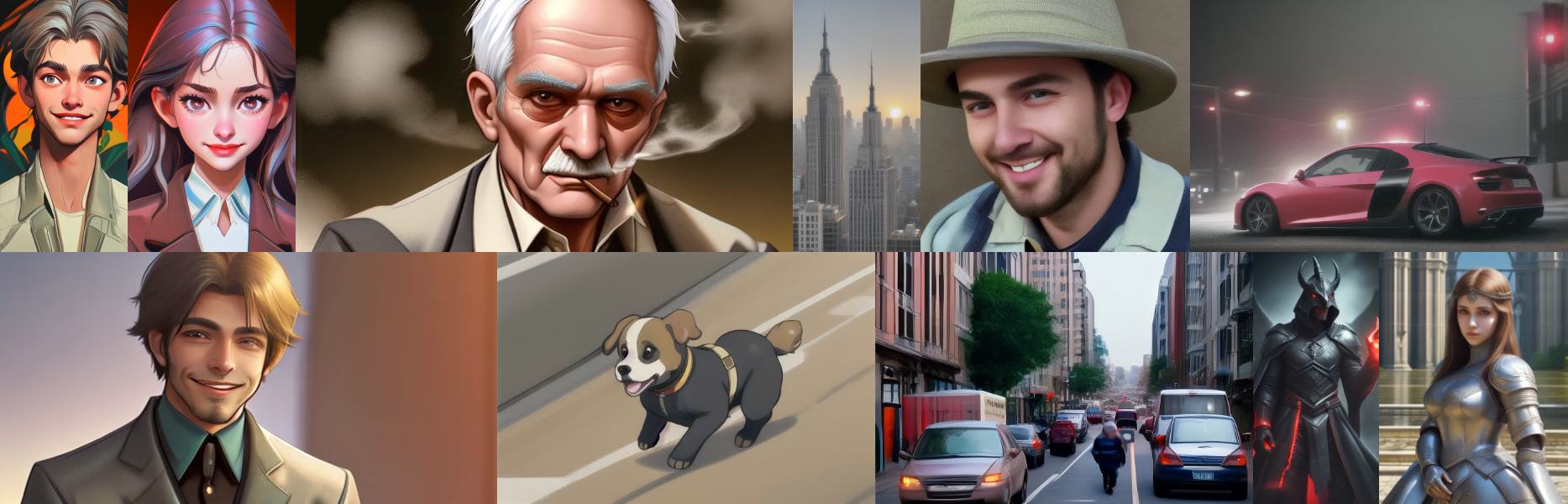}
    \caption{Text-to-video generation of different aspect ratios. Examples shown here are 2-step and 4-step models generating 1:2, 2:3, 3:2, and 2:1 aspect ratios. We show a random frame from the generated video clips.}
    \label{fig:aspect-ratio}
\end{figure*}

\begin{figure*}
    \begin{subfigure}{0.496\textwidth}
        \centering
        \setlength\tabcolsep{0pt}
        \renewcommand{\arraystretch}{0}
        \begin{tabularx}{\linewidth}{XXXXX}
            \includegraphics[width=\linewidth]{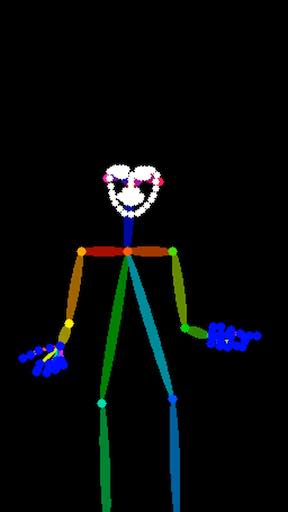} &
            \includegraphics[width=\linewidth]{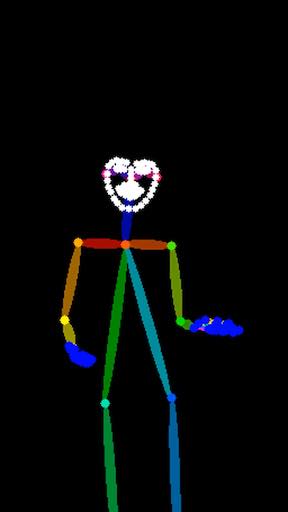} &
            \includegraphics[width=\linewidth]{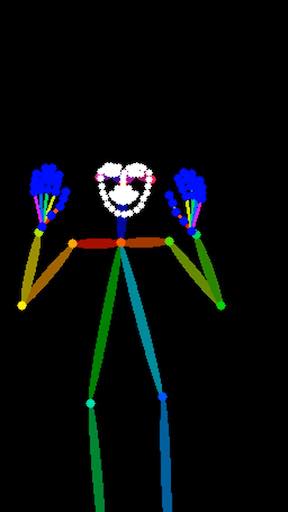} &
            \includegraphics[width=\linewidth]{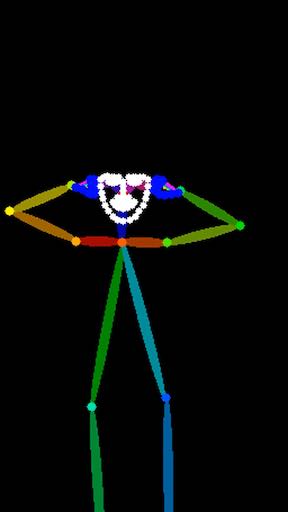} &
            \includegraphics[width=\linewidth]{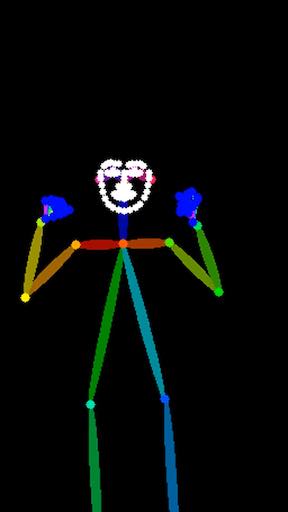} \\
            \includegraphics[width=\linewidth]{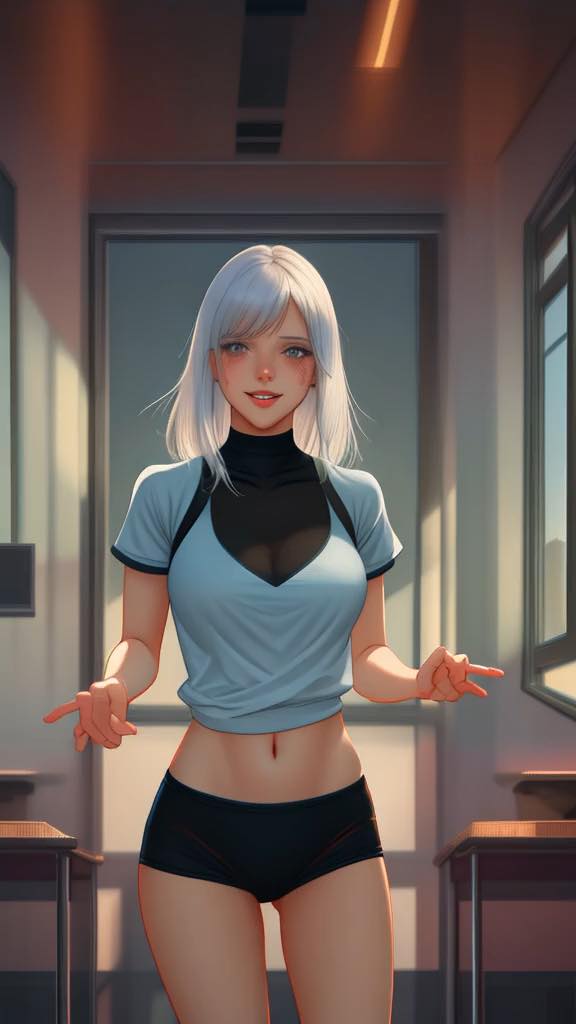} &
            \includegraphics[width=\linewidth]{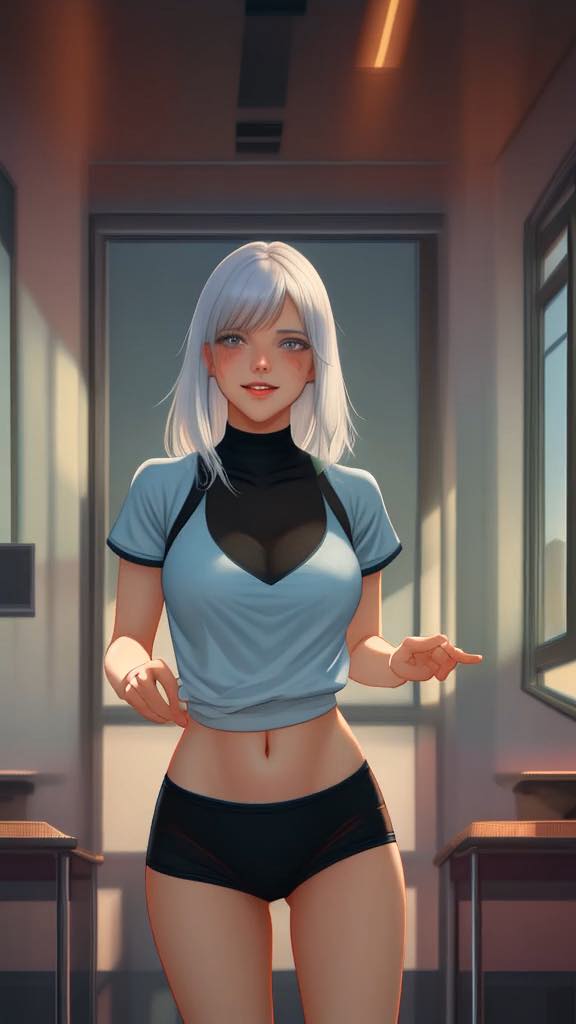} &
            \includegraphics[width=\linewidth]{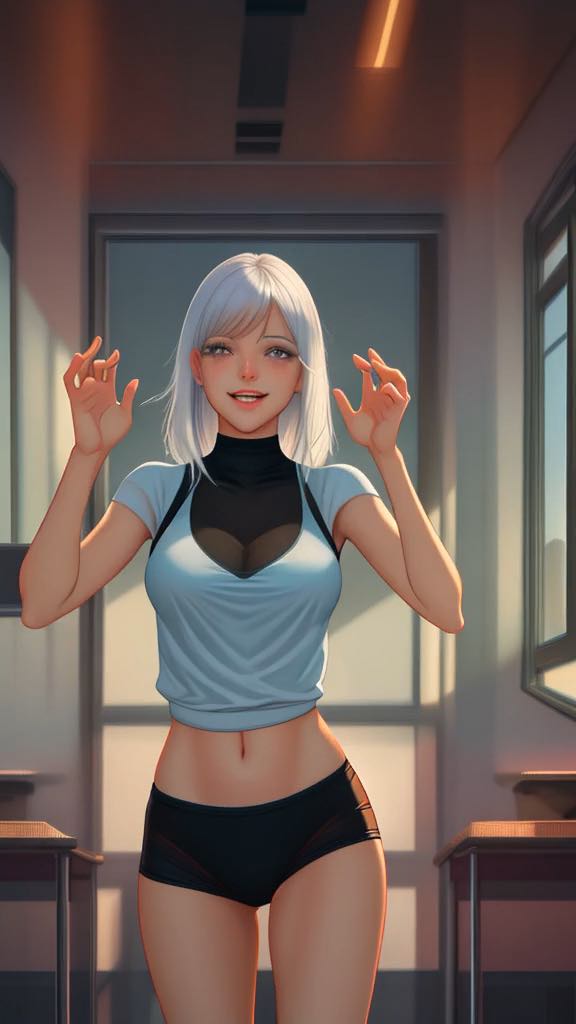} &
            \includegraphics[width=\linewidth]{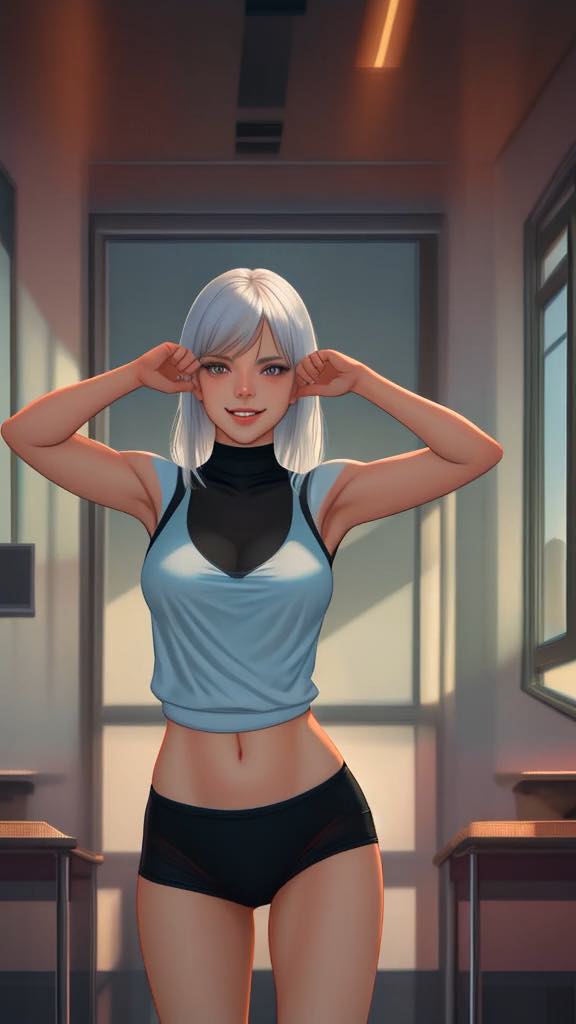} &
            \includegraphics[width=\linewidth]{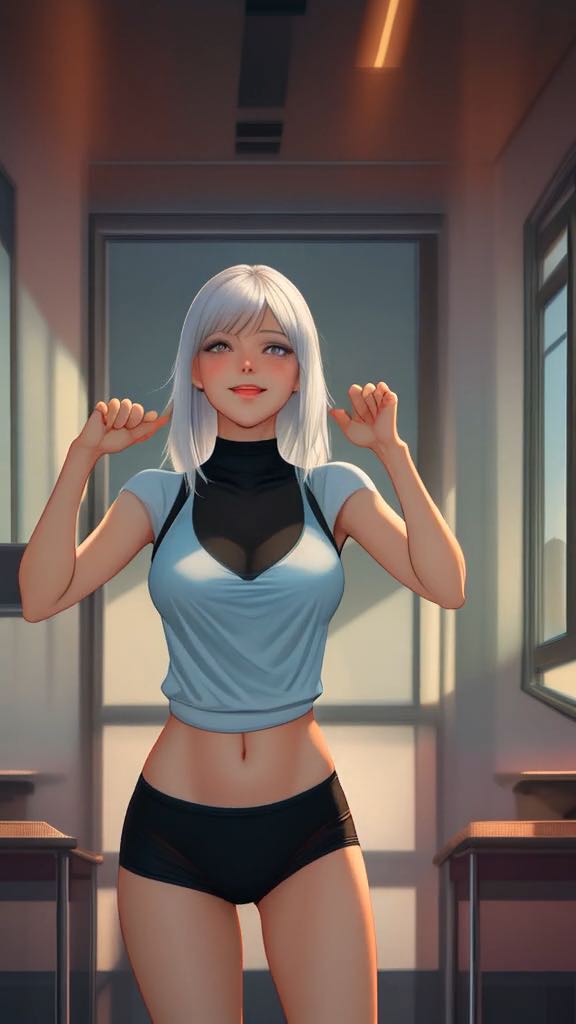}
        \end{tabularx}
        \caption{4 Steps, IMP v1.0 \cite{imp}, DWPose \cite{yang2023effective}}
    \end{subfigure}
    \hfill
    \begin{subfigure}{0.496\textwidth}
        \centering
        \setlength\tabcolsep{0pt}
        \renewcommand{\arraystretch}{0}
        \begin{tabularx}{\linewidth}{XXXXX}
            \includegraphics[width=\linewidth]{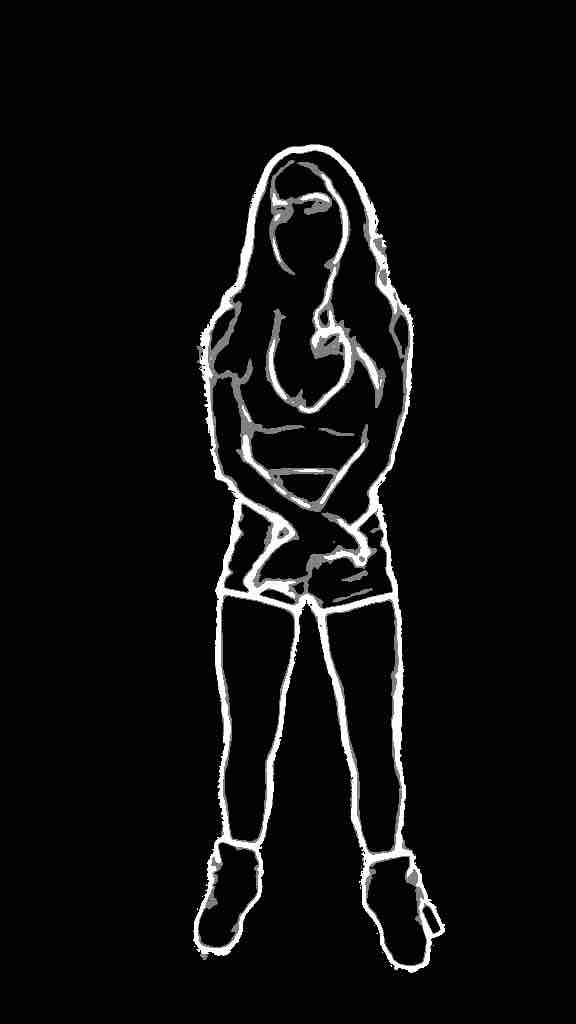} &
            \includegraphics[width=\linewidth]{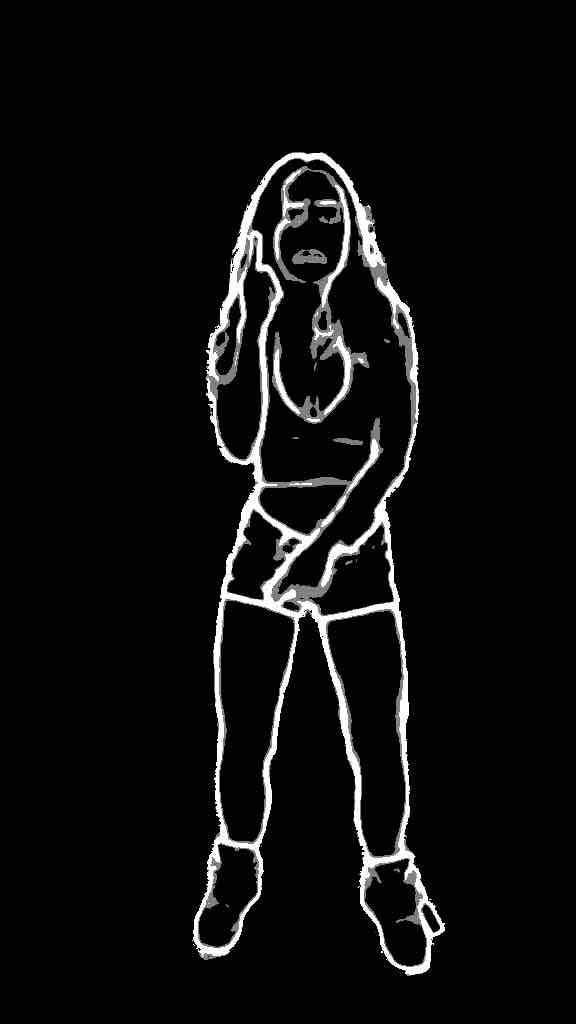} &
            \includegraphics[width=\linewidth]{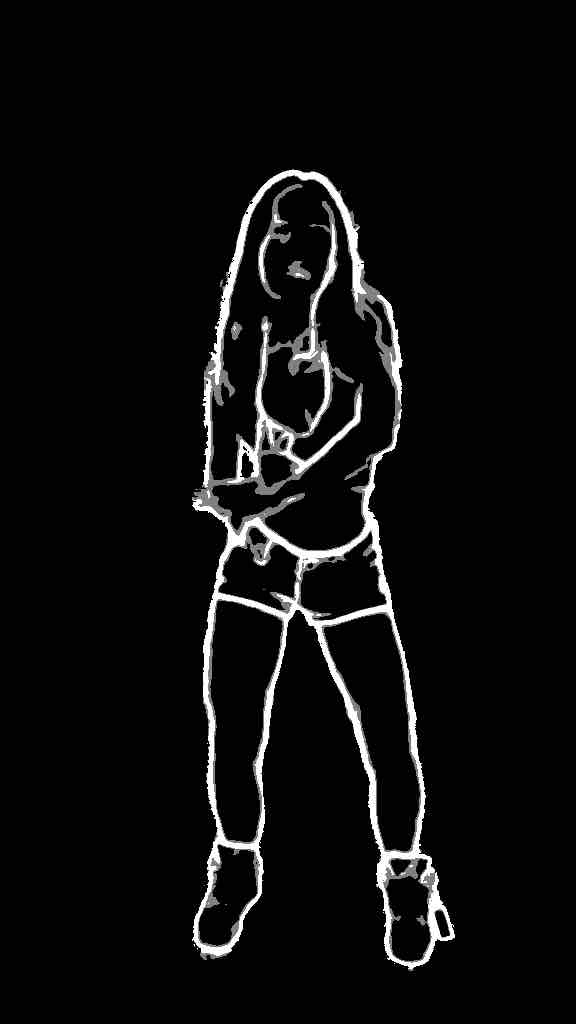} &
            \includegraphics[width=\linewidth]{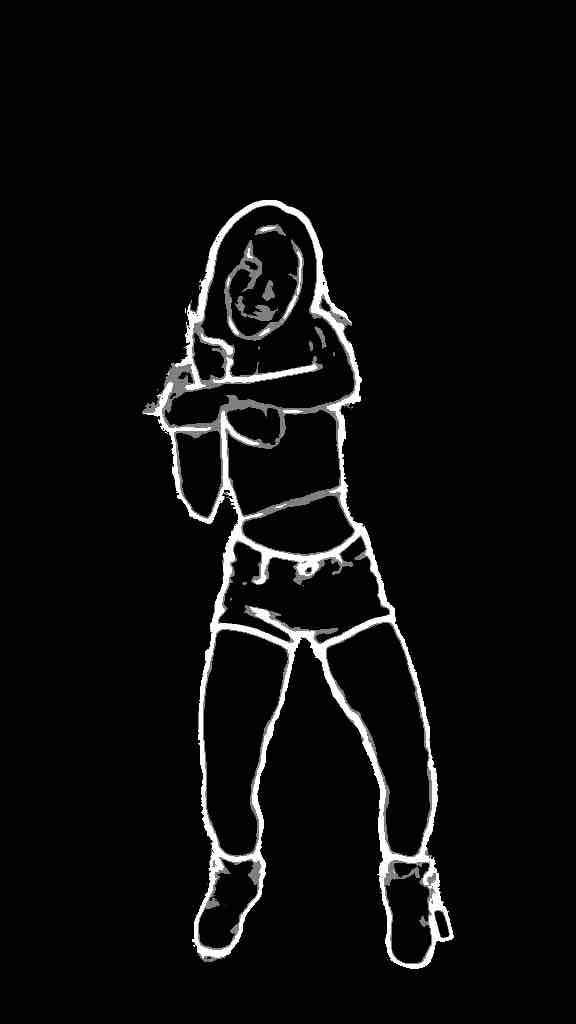} &
            \includegraphics[width=\linewidth]{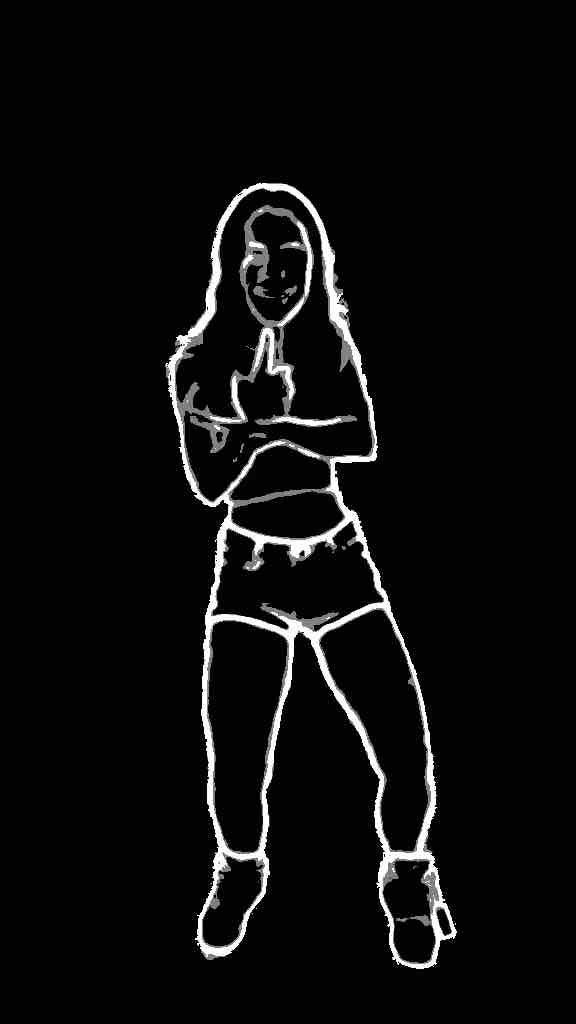} \\
            \includegraphics[width=\linewidth]{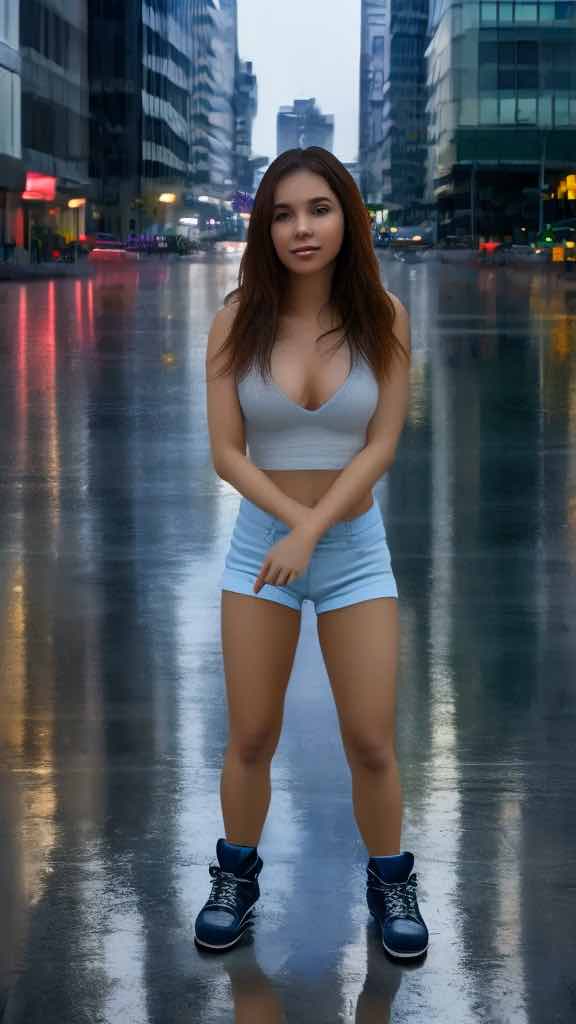} &
            \includegraphics[width=\linewidth]{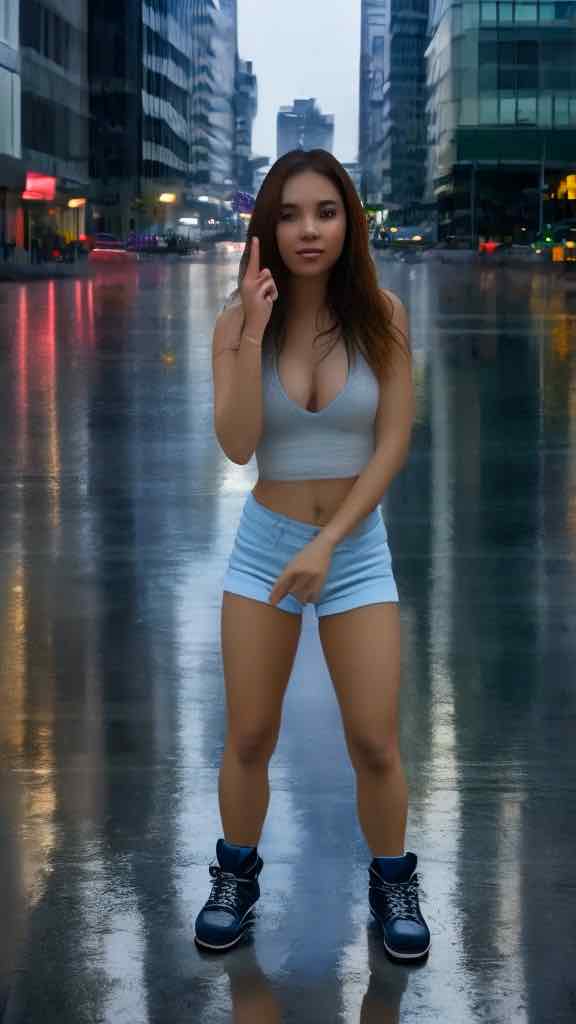} &
            \includegraphics[width=\linewidth]{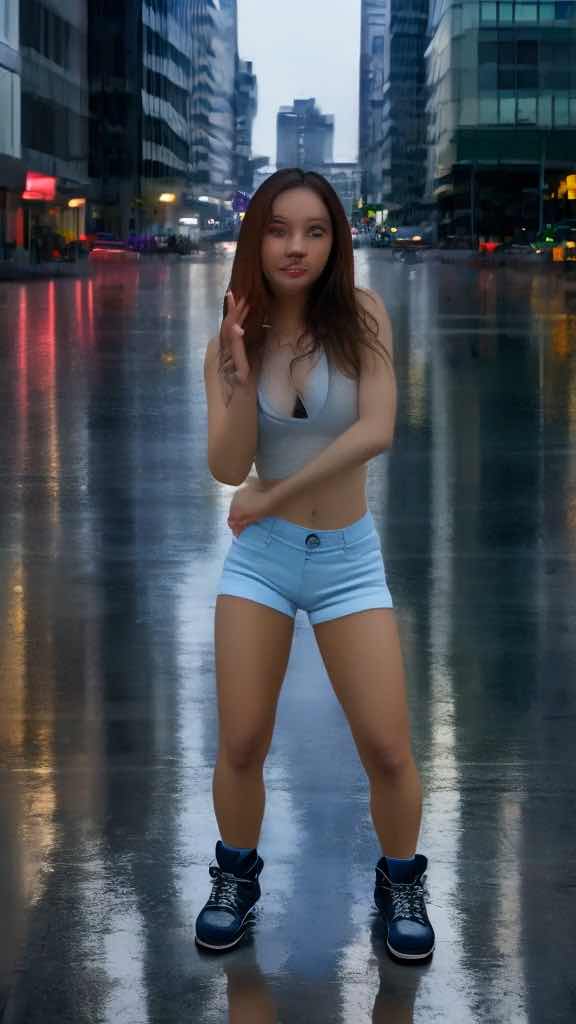} &
            \includegraphics[width=\linewidth]{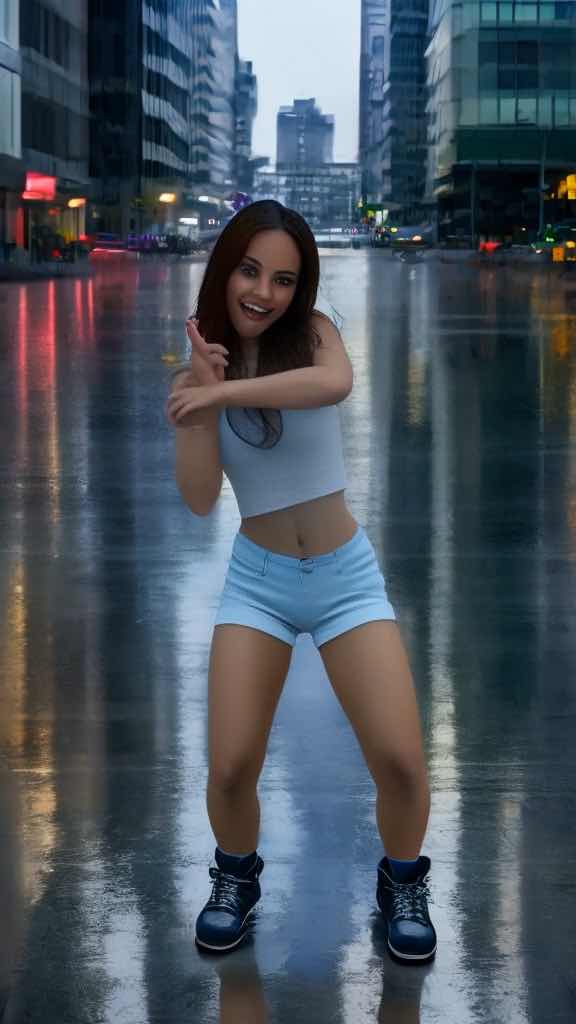} &
            \includegraphics[width=\linewidth]{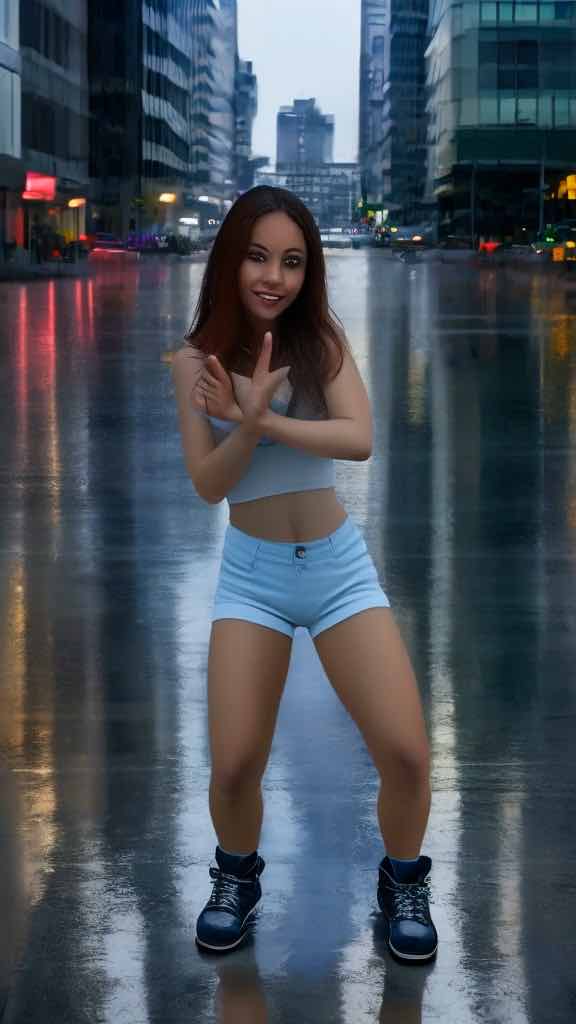}
        \end{tabularx}
        \caption{2 Steps, epiCRealism \cite{epicrealism}, HED \cite{Xie2015HolisticallyNestedED}, RobustVideoMatting \cite{Lin2021RobustHV}}
    \end{subfigure}
    
    \caption{Video-to-video generation with ControlNet \cite{controlnet}. The example videos are generated in $576\times1024$ resolution directly using our model with ControlNet \cite{controlnet}. More sophisticated pipelines, such as using super-resolution, can further enhance the quality.}
    \label{fig:controlnet}
\end{figure*}

\clearpage

\subsection{Quantitative Evaluation}

\begin{table}[h]
    \centering
    \setlength\tabcolsep{3pt}
    \begin{tabularx}{\linewidth}{Xr|rr|rr}
        \toprule
        \multirow{2}{*}{Method} & \multirow{2}{*}{Steps} & \multicolumn{4}{c}{FVD $\downarrow$} \\
        & & RV \cite{realisticvision} & TY \cite{toonyou} & DS \cite{dreamshaper} & DV \cite{dynavision} \\
        \midrule
        \multirow{4}{*}{AnimateLCM} & 1 & 1423.18 & 1825.24 & 1393.10 & 1652.32 \\
         & 2 & 1041.61 & 917.61 & 1034.19 & 1045.49 \\
         & 4 & 1171.54 & 784.81 & 1175.06 & 1097.66 \\
         & 8 & 1300.41 & 804.21 & 1253.43 & 1115.95 \\
        \midrule
        \multirow{4}{*}{Ours} & 1 & 1135.43 & 1037.85 & 974.75 & 1501.34 \\
         & 2 & 1024.13 & 801.04 & 918.74 & 1351.06 \\
         & 4 & 1010.30 & 708.55 & 908.01 & 1175.29 \\
         & 8 & 1058.58 & 690.65 & 865.29 & 979.94 \\
        \bottomrule
    \end{tabularx}
    \caption{FVD computed against original AnimateDiff on different image base models. RV: RealisticVision, TY: ToonYou, DS: DreamShaper, DV: DynaVision.}
    \label{tab:quantitative}
\end{table}

\Cref{tab:quantitative} shows quantatitive comparison. First, we randomly select 100 prompts from the WebVid-10M dataset \cite{Bain2021FrozenIT}. Then, we generate the clips using four different image base models. We select RealisticVision \cite{realisticvision} and ToonYou \cite{toonyou} as seen realistic and anime style models, and select DreamShaper \cite{dreamshaper} and DynaVision \cite{dynavision} as unseen realistic and anime style models. Each prompt uses a random seed but the same seed is used across models on the same prompt. Finally, we compute FVD \cite{Unterthiner2018TowardsAG} against the original AnimateDiff results generated using 32 Euler steps and CFG 7.5 without negative prompts. Both ours and AnimateLCM \cite{wang2024animatelcm} do not use CFG. The metrics show that our models have better FVD compared to AnimateLCM and therefore produce results closer to the original AnimateDiff.

\section{Ablation}

\subsection{Effects of Cross-Model Distillation}
\label{sec:cross}

We conduct a comparison experiment to distill a model only using Stable Diffusion v1.5 \cite{rombach2022highresolution} as the image base model on the WebVid-10M \cite{Bain2021FrozenIT} dataset. This corresponds to the regular single-model distillation paradigm.

\Cref{fig:cross-ablation} shows that single-model distillation can only keep the best quality on the default SD \cite{rombach2022highresolution} base model. The quality degrades after switching to RealisticVision \cite{realisticvision} which has a similar realistic style. The quality significantly degrades after switching to ToonYou \cite{toonyou} which has a drastically different anime style.

\subsection{Effects on Unseen Base Models}
\label{sec:unseen}

We test our model on a wide variety of popular image base models. These base models are unseen during the distillation process. \Cref{fig:unseen} shows that our distilled motion module can generalize well to other unseen base models. Furthermore, our distilled model produces results with sharper details and closer styles to the original model compared to AnimateLCM \cite{wang2024animatelcm}.

\subsection{Compatibility with Motion LoRAs}

\Cref{fig:motion-lora} shows that our model is compatible with Motion LoRAs \cite{guo2023animatediff}. We have tested Motion LoRAs on all our models and have found that they work in all step settings. We apply Motion LoRAs with a strength of 0.8 to avoid watermarks, an issue Motion LoRAs introduce. We find Motion LoRAs enable fine-grained control of the camera motion and they greatly enhance the amount of motion in the generated videos.

\subsection{Support for Different Aspect-Ratios}

\Cref{fig:aspect-ratio,fig:controlnet} show that our model retains the ability to generate videos of different aspect-ratios on both text-to-video and video-to-video scenarios despite the distillation is performed only in square aspect ratio. However, we find that as the aspect ratio deviates more from the square, there is a higher probability of generating bad cases. The distillation training can be done in multiple aspect ratios. We leave this to future improvements.

\subsection{Video-to-Video Generation with ControlNet}

One of AnimateDiff's most popular uses is video-to-video stylization. Given a source video, ControlNet \cite{controlnet} is applied to extract human movement, and then AnimateDiff is used to generate the movement using different styles.

\Cref{fig:controlnet} shows that our model is compatible with ControlNet \cite{controlnet}. Here we only apply the basic setting, but a more sophisticated pipeline, such as using super-resolution and background replacement, can be additionally added. To generate longer videos, the popular approach is context overlapping, which overlaps the 16-frame context window with previously generated clips. We have tested that our models support generating longer videos with context overlapping.

\section{Conclusion}

We have presented AnimateDiff-Lightning, a lightning-fast video generation model. In this paper, we have shown that progressive adversarial diffusion distillation can be applied in the video modality. Our model achieves new state-of-the-art in few-step video generation. Additionally, we have proposed cross-model diffusion distillation to further improve the distillation module's ability to generalize to different stylized base models. We apply the cross-model distillation technique on AnimateDiff because it is most widely used with different image base models. However, this technique can be generalized to create more universal distillation pluggable modules for all modalities. Lastly, we release our distilled AnimateDiff-Lightning models with the hope of facilitating advancements in generative AI.

\newpage

{\small
\bibliographystyle{ieee_fullname}
\bibliography{main}
}

\end{document}